\documentclass[pdflatex,sn-vancouver-ay]{sn-jnl}

\usepackage{amsmath,amssymb,amsfonts}
\usepackage{amsthm}
\usepackage{mathrsfs}
\usepackage{stmaryrd}

\usepackage{graphicx}
\usepackage{svg}
\usepackage{float}
\usepackage{subcaption}
\usepackage{caption}

\usepackage{booktabs}
\usepackage{multirow}
\usepackage{tabularx}
\usepackage{array}
\usepackage{makecell}
\usepackage{colortbl}
\usepackage{threeparttable}
\usepackage{adjustbox}

\usepackage{algorithm}
\usepackage{algorithmicx}
\usepackage{algpseudocode}

\usepackage{lingmacros}
\usepackage[linguistics]{forest}
\usepackage{tikz-qtree}

\usepackage{tikz}
\usepackage{pgfplots}
\usetikzlibrary{arrows,shapes,positioning,shadows,trees}
\usetikzlibrary{mindmap}
\usetikzlibrary{arrows.meta,decorations.markings}

\usepackage[dvipsnames]{xcolor}
\usepackage{tcolorbox}
\tcbuselibrary{skins}

\usepackage{textcomp}
\usepackage{manyfoot}
\usepackage{appendix}
\usepackage{url}
\usepackage{listings}
\usepackage{ulem}
\usepackage{microtype}
\usepackage{balance}
\usepackage{scalerel}
\usepackage{todonotes}
\usepackage{array} 

\makeatletter
\newcommand{\es}{\enumsentence}

\newcommand{\ms}[2][]{%
  \mbox{%
    \delimiterfactor=1000 \delimitershortfall=0pt
    \tabcolsep=0pt
    $\left[%
    \begin{tabular}{>{\upshape\scshape}l@{\hspace{5pt}}>{\normalfont\itshape}l}
      \if\relax\detokenize{#1}\relax\else
      \multicolumn{2}{>{\normalfont\itshape}l}{#1}%
      \\
      \fi
      #2%
    \end{tabular}%
    \right]$%
  }%
  \vspace{1mm}%
}
\def\ibox#1{\mbox{}\setbox2=\hbox{$\scriptstyle #1$}\lower.2ex\vbox{\hrule
    \hbox{\vrule\kern1.25pt
      \vbox{\kern1.25pt\box2\kern1.25pt}\kern1.25pt\vrule}\hrule}}

\newcommand*\FilledCircled[1]{\tikz[baseline=(char.base)]{
    \node[shape=circle,fill=blue!60,text=white,inner sep=1pt] (char) {\small{#1}};}}

\newcommand*{\Scale}[2][4]{\scalebox{#1}{$#2$}}%

\definecolor{myred}{RGB}{204, 0, 0}
\definecolor{mygreen}{RGB}{0, 153, 51}
\definecolor{flowred}{RGB}{240, 130, 141}
\definecolor{flowyellow}{RGB}{244, 224, 138}
\definecolor{flowgreen}{RGB}{150, 225, 191}
\definecolor{flowblue}{RGB}{158, 150, 238}

\theoremstyle{thmstyleone}%
\theoremstyle{thmstyletwo}%

\theoremstyle{thmstylethree}%

\begin{document}

\title[Towards Linguistically-informed Representations for
English as a Second or Foreign Language: Review,
Construction and Application]{Towards Linguistically-informed Representations for
English as a Second or Foreign Language: Review,
Construction and Application}


\author[1,2]{\fnm{Wenxi} \sur{Li}}\email{liwenxi@pku.edu.cn}

\author[3]{\fnm{Xihao} \sur{Wang}}\email{wangxihao@pku.edu.cn}

\author*[4]{\fnm{Weiwei} \sur{Sun}}\email{ws390@cam.ac.uk}

\affil[1]{\orgdiv{School of the Chinese Nation Studies}}

\affil[2]{\orgdiv{School of Liberal Arts}, \orgname{Minzu University of China}, \orgaddress{\street{27 South Zhongguancun Ave}, \city{Beijing}, \postcode{100081},
\country{China}}}

\affil[3]{\orgdiv{Department of Chinese Language and Literature}, \orgname{Peking University}, \orgaddress{\street{5 Yiheyuan Rd}, \city{Beijing}, \postcode{100871}, \state{Beijing}, \country{Chia}}}

\affil*[4]{\orgdiv{Department of Computer Science and Technology}, \orgname{University of Cambridge}, \orgaddress{\street{15 JJ Thomson Ave}, \city{Cambridge}, \postcode{CB3 OFD}, \state{Cambridgeshire}, \country{United Kingdom}}}

\abstract{The widespread use of English as a Second or Foreign Language (ESFL) has sparked a paradigm shift: ESFL is not seen merely as a deviation from standard English but as a distinct linguistic system in its own right. This shift highlights the need for dedicated, knowledge-intensive representations of ESFL.
In response, this paper surveys existing ESFL resources, identifies their limitations, and proposes a novel solution. Grounded in constructivist theories, the paper treats constructions as the fundamental units of analysis, allowing it to model the syntax--semantics interface of both ESFL and standard English. This design captures a wide range of ESFL phenomena by referring to syntactico-semantic mappings of English while preserving ESFL's unique characteristics, resulting a gold-standard syntactico-semantic resource comprising 1643 annotated ESFL sentences.
To demonstrate the sembank's practical utility, we conduct a pilot study testing the Linguistic Niche Hypothesis, highlighting its potential as a valuable tool in Second Language Acquisition research.

}




\maketitle

\section{Introduction}
\label{intro}
In today's globalized world,, English as a Second or Foreign Language (ESFL), produced by non-native speakers who make up over 70\% of English users worldwide, has become a primary tool for cross-lingual communication, particularly in open and informal contexts such as social media. This widespread usage necessitates a critical paradigm shift: ESFL may not be recognized as an erroneous approximation of native norms, but as a legitimate, distinct linguistic system shaped by the complex interplay of a speaker's primary languages and English \citep{Selinker+1972+209+232,nemser1991language,corder1982error}.
As such, ESFL requires dedicated linguistically-informed resources to accurately represent and process its unique structures just like any other natural language.

However, a survey of existing representations for ESFL (see \textsection\ref{related}), ranging from POS tagging \citep{diaz2010towards} to universal syntactic frameworks \citep{berzak-etal-2016-universal} and semantic annotations \citep{zhao-etal-2020-semantic}, reveals a fundamental challenge. Because ESFL lacks a static ``native'' anchor and is characterized by high variability and rapid evolution, developing a consistent framework is difficult. Most existing resources use standard English as a rigid reference point. This creates a persistent tension: strictly adhering to native norms risks erasing the distinctive features and cross-linguistic nuances of the multilingual speaker, while relaxing those norms often compromises the coherence of the representation.

To resolve this, we adopt a multilingual-centric constructivist approach. Constructivist theories posit that language consists of ``constructions'', i.e., form–meaning pairings that serve as the fundamental units of linguistic analysis \citep{goldberg2003constructions, croft2001radical}. We therefore propose deriving sentence-level representations for both standard English and ESFL from a shared inventory of reusable syntactico-semantic constructions. This approach preserves the distinctive surface forms (syntax) generated by multilingual speakers while grounding them in the underlying semantic equivalents shared with English. By modeling the syntax–semantics interface as a flexible mapping, we can capture the idiosyncratic features of the ESFL grammar without sacrificing representational consistency.

We implement this approach using Synchronous Hyperedge Replacement Grammar (SHRG), a formalism that pairs syntactic rules, represented as trees, with corresponding semantic rules over graphs. Together, these synchronized rules form a constructional inventory that treats ESFL sentences as systematic linguistic productions rather than as random or ill-formed errors. Building on this SHRG framework, we develop ESFL SemBank, a gold-standard syntactico-semantic resource (see \textsection\ref{sembank}). Its construction follows a rigorous Review–Revise–Rebuild workflow: we manually review silver-standard semantic graphs \citep{zhao-etal-2020-semantic}, revise rejected analyses to ensure semantic fidelity, and rebuild unparsable sentences by selecting or creating appropriate SHRG rules.

ESFL SemBank provides a powerful lens for language acquisition. By explicitly encoding how ESFL speakers --- whose second or foreign language is English --- map linguistic form to meaning, the resource enables in-depth analyses of the cognitive mechanisms and developmental trajectories underlying ESFL production. In \textsection\ref{compare}, we illustrate its utility through a case study of the Linguistic Niche Hypothesis \citep{lupyan2010language}. Our statistical analysis shows that, while ESFL syntactic patterns largely align with those of English, ESFL often exhibits a more transparent and consistent mapping between form and meaning. This finding suggests that multilingual speakers may optimize the syntax–semantics interface for communicative clarity, offering empirical evidence of how non-native speakers navigate typological constraints to achieve cross-lingual fluency.

\section{Survey: Existing Resources for ESFL}
\label{related}
Efforts to develop knowledge-intensive, task-independent, and linguistically-informed representations for human languages have become an important research area \citep[e.g.,][]{oepen-lonning-2006-discriminant, slacker, amr, abend-rappoport-2013-universal}. Similar efforts have also been extended to ESFL, leading to the development of several corpora specifically designed for it. Based on their strategies for referencing standard English, we group these corpora into three categories:
\begin{itemize}
\item \textbf{Target-language-based}: directly projecting English-based frameworks or models onto ESFL data (\textsection \ref{target}).
\item \textbf{Multiple-criteria-based}: applying various English-derived criteria to interpret ESFL phenomena (\textsection \ref{multiple}).
\item \textbf{Universal-framework-based}: using a language-neutral framework to mediate between English and ESFL (\textsection \ref{universl}).
\end{itemize}

\subsection{Target-language-based Approach}
\label{target}

One representative example using the target-language-based approach is the Konan-JIEM Learner Corpus \citep{nagata-etal-2011-creating,nagata-sakaguchi-2016-phrase}. This resource introduces a phrase-structure-based annotation scheme and a manually constructed treebank. Annotations are grounded in native English syntactic norms, consistently identifying the structure in standard English that most closely reflects the learner’s intended meaning. When learner productions cannot be aligned with any plausible native structure, the scheme introduces specialized node labels --- such as \textsc{-err}, \textsc{in}, \textsc{ce}, \textsc{xp-ord}, \textsc{uk}, and \textsc{up} --- to mark errors related to omission, insertion, substitution, word order, and unknown words or phrases, respectively.

Another example is the SemBank of English as a Second Language \citep{zhao-etal-2020-semantic}, which, so far as we know, is the only effort to construct a comprehensive semantic bank for learner English. This resource adopts a target-language-based strategy by using the ACE parser, trained on English Resource Grammar \citep[ERG;][]{erg}, to parse ESFL sentences. For each ESFL sentence $S$, the parser derives $K$-best corresponding semantic graphs. A reranking procedure then selects the candidate that best aligns with the gold-standard syntactic tree from the TLE corpus \citep{berzak-etal-2016-universal}. In total, 2691 ESFL sentences are successfully parsed.

\subsection{Multiple-criteria-based Approach}
\label{multiple}

Resources under the multiple-criteria-based approach, drawing on multiple English-derived criteria, adopt layered annotation strategies to interpret ESFL data more flexibly.
For instance, \citet{diaz2010towards} propose a tripartite analysis for POS tagging of ESFL in the NOCE Corpus, a Spanish learner corpus of English. Based on an empirical investigation of learner language collected in NOCE, the authors observe that applying single POS tagging scheme is problematic --- the distributional, lexical, and morphological evidence for classifying a token in ESFL data often does not converge on a single POS tag. Therefore, they suggest annotating each word with three parallel POS tags to capture its lexical, morphological, and distributional forms, respectively.

The Syntactically Annotated Learner Language of English (SALLE) project \citep{ragheb_2023_10054513, dickinson-ragheb-2015-grammaticality} also proposes a multi-layer framework to represent the morpho-syntactic information of learner language. Based on the CHILDES annotation framework \citep{sagae-etal-2004-adding, sagae-etal-2007-high}, both POS tags and dependencies are divided into two layers: one for morphological information and the other for distributional information. This corpus also includes subcategorization frames to represent dependencies that are selected for by words but not necessarily realized. 

\subsection{Universal-framework-based Approach}
\label{universl}
The Treebank of Learner English \citep[TLE;][]{berzak-etal-2016-universal}, also known as UD English-ESL, employs a universal-framework-based approach. Each ESFL sentence is manually corrected into standard English, and both versions are annotated using the Penn Treebank POS tagset \citep{santorini1990} and syntactic dependencies from the Universal Dependencies framework \citep[UD;][]{nivre-etal-2016-universal}. These annotation schemes are thought to be language-neutral, thus allowing comprehensive coverage of both learner ESFL sentences and their English counterparts.

Besides TLE, several corpora, including the Treebank of Spoken Second Language English \citep[SL2E Treebank;][]{kyle-etal-2022-dependency}, as well as ESFL resources for other learner populations \citep{lee-etal-2017-l1, lin-etal-2018-semantic, davidson-etal-2020-developing-nlp} also adopts this approach.

\subsection{Summary}
In our opinion, the aforementioned approaches possess distinct strengths and inherent limitations. The target-language-based approach, for instance, benefits significantly from leveraging existing English resources, offering efficiency and convenience in its application. However, a major drawback is its inability to comprehensively capture the distinctive features unique to ESFL.
Conversely, the multiple-criteria-based approach offers more detailed descriptions of ESFL, which is valuable for in-depth analysis. However, its primary challenge lies in scalability, as the complexity and cost of annotating large-scale ESFL data become prohibitive. Meanwhile, the universal-framework-based approach endeavors to connect ESFL with the target language (English) through their shared semantic interpretation. Yet, the finer-grained correspondences between the two languages remain undefined.

We contend that these collective difficulties underscore a fundamental challenge in constructing a systematic representation framework for ESFL --- unlike creoles or pidgins, ESFL cannot readily rely on native speaker intuitions to judge its well-formedness. As a result, explaining and predicting variations of ESFL in a principled way remains especially difficult. Existing approaches typically rely on standard English as a reference point, but this leads to a core dilemma: strictly following standard English rules obscures the distinctive features of ESFL, whereas relaxing those constraints too much undermines systematicity and internal coherence. Therefore, striking an appropriate balance between these two poles becomes the pivotal question to address.

\label{sec3}

\section{Method}
\label{theory}
\subsection{Theoretical Foundation: Construcivism}
Faced with the challenges of representing ESFL data, we draw on insights from constructivism. Though encompassing a broad and heterogeneous body of research (see \citeauthor{hoffmann2013oxford}, \citeyear{hoffmann2013oxford}; \citeauthor{goldberg2013argument}, \citeyear{goldberg2013argument}), constructivist theories are unified by the core assumption that words, idioms, phrases, and even full sentences are form–meaning pairings, or constructions, that are stored in an inventory capable of exhaustively describing human language.
Building on this foundation, we propose using syntactico-semantic constructions as the basic representational units for both ESFL and standard English. By decomposing sentence-level representations into finer-grained constructions and aligning ESFL and English at this micro-level, the constructivist framework offers a principled means of bridging the two languages. 

\begin{figure}[htbp]
\centering
  \subcaptionbox{ESFL sentence\label{fig:1a}}[0.3\linewidth]{%
      \scalebox{0.6}{
  \begin{forest}
        for tree={s sep=1.3em,l sep=0.5em},
        [$\textrm{S}$ 
        [$\textrm{NP}$ [$\textrm{N}$ [\{\textcolor{black}{I}\}]]]
        [$\textrm{VP}$ 
            [$\textrm{V}$ [\{\textcolor{black}{sleep}\}]] 
            [$\textrm{PP}$
                [$\textcolor{black}{\textrm{P}}$ [\{\textcolor{black}{in}\}\\]] 
                [$\textcolor{myred}{\textrm{NP}}$, tikz={\node [draw,inner sep=-1pt,fit to=tree, color=myred,line width=1pt] {};} [$\textcolor{myred}{\textrm{N}}$ [\{\textcolor{myred}{tent}\}]]]]]]
\end{forest}}
  }
  \hfill
  \subcaptionbox{English counterpart\label{fig:1b}}[0.3\linewidth]{%
     \scalebox{0.6}{
  \begin{forest}
        for tree={s sep=1.3em,l sep=0.5em},
        [$\textrm{S}$ 
        [$\textrm{NP}$ [$\textrm{N}$ [\{\textcolor{black}{I}\}]]]
        [$\textrm{VP}$ 
            [$\textrm{V}$ [\{\textcolor{black}{sleep}\}]] 
            [$\textrm{PP}$
                [$\textcolor{black}{\textrm{P}}$ [\{\textcolor{black}{in}\}\\]] 
                [$\textcolor{ForestGreen}{\textrm{NP}}$, tikz={\node [draw,inner sep=-1pt,fit to=tree, color=ForestGreen,line width=1pt] {};} 
                [$\textcolor{ForestGreen}{\textrm{DET}}$ [\{\textcolor{ForestGreen}{a}\}\\]] [$\textcolor{ForestGreen}{\textrm{N}}$ [\{\textcolor{ForestGreen}{tent}\}]]]]]]
\end{forest}}
  }
  \hfill
  \subcaptionbox{SYN-SEM rule\label{fig:1c}}[0.3\linewidth]{%
     \begin{align*}
  \begin{aligned}
  & \Scale[0.6]{\textrm{\textbf{SYNTAX}}}\\
  &\Scale[0.6]{\textrm{\textcolor{myred}{NP} }\textcolor{myred}{\rightarrow} \textrm{ \textcolor{myred}{N} \{tent\} }} \\
  &\Scale[0.6]{\textrm{\textcolor{mygreen}{NP} } 
 \textcolor{mygreen}{\rightarrow} \textrm{ \textcolor{mygreen}{DET} \{a\} } \textcolor{mygreen}{+} \textrm{ \textcolor{mygreen}{N} \{tent\}}} \\[5pt] 
  & \Scale[0.6]{\textrm{\textbf{SEMANTICS}}}\\
  & \Scale[0.6]{\exists x. \textrm{tent}'(x)}\\
\end{aligned} 
 \end{align*}
  }
  \caption{An illustration of our method using the example \textit{I sleep in \textbf{(a)} tent}. The constructions highlighted by rectangles differ in whether the article is omitted, but they share similar semantics and structural encoding within the syntactic tree, allowing them to be aligned.}
  \label{fig:main-intro}
\end{figure}
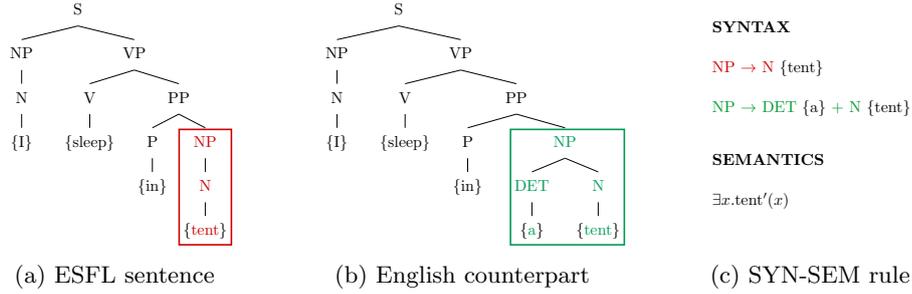

Figure~\ref{fig:main-intro} illustrates how our method addresses the so-called ``errors'' in ESFL, i.e., its deviations from standard English. While ESFL constructions diverge in syntactic form due to token omission, insertion, or unconventional structures (see Examples~\ref{ex1-intro}–\ref{ex3-intro}), their intended semantics remains intelligible. Therefore, these ESFL constructions can be systematically linked to their standard English counterparts by identifying underlying semantic equivalences and shared distributional patterns. 

\es{
Omission: \textit{I had to sleep in \textbf{(a)} tent.} \label{ex1-intro}
}
\es{
Insertion: \textit{We contacted \textbf{\sout{with}} Kim.} \label{ex2-intro}
}
\es{
Transposition: \textit{He \uwave{\textbf{visits often}} Paris.} \label{ex3-intro}
}

We argue that the constructivist approach --- which represents ESFL data through a continually expanding inventory of constructions, in contrast to lexicalist frameworks that rely on systematic rules for combining lexical units --- offers two key advantages.
First, it effectively accounts for the wide range of ``errors'' typically found in ESFL. By recognizing alternative form–meaning pairings that emerge organically from actual usage, rather than imposing rigid grammatical norms, the constructivist framework captures the internal variation of ESFL in a more nuanced and context-sensitive way.
Second, while accommodating variation, this approach also maintains formal coherence across representations. ESFL constructions can be aligned with their standard English counterparts and systematically encoded into structural templates, ensuring their compatibility with subsequent syntactico-semantic composition processes.
Thus, rather than requiring a trade-off between descriptive adequacy and formal systematicity, the constructivist approach achieves both as a flexible, principled, and robust framework.

\subsection{Modeling the Syntax–Semantics Interface}
To put the constructivist approach into practice, we adopt Synchronous Hyperedge Replacement Grammar (SHRG), a formalism that models each linguistic construction as a meaning-bearing unit. In SHRG, every construction consists of two tightly coupled components: a syntactic rule, represented as a tree that encodes form, and a semantic rule, represented as a graph that encodes meaning. These two components are synchronized, forming two sides of the same coin.

\begin{figure}[htbp]
    \centering
    \begin{minipage}[c]{0.45\textwidth}
        \centering
        \subfloat[Syntax Tree\label{fig:2a}]{
        \scalebox{0.75}{
        \begin{forest}
            for tree={s sep=0.5em, l sep=1em, font=\small}
            [VP
                [V 
                    [V [visit]] 
                    [Adv [often]] 
                ]
                [NP [Paris]] 
            ]
        \end{forest}
        }
        }
    \end{minipage}
    \hspace{3pt}
    \begin{minipage}[c]{0.45\textwidth}
        \centering
        \subfloat[Semantic Graph\label{fig:2b}]{
        \scalebox{0.7}{
        \begin{tikzpicture}[>=latex,line join=bevel,]
    \node (x1) at (124.74bp,155.0bp) [draw=black,circle,inner sep=2pt] {};
  \coordinate (x11) at (145.74bp,116.5bp);
  \node (x2) at (91.737bp,116.5bp) [draw=black,inner sep=2pt,circle] {};
  \coordinate (x22) at (91.737bp,78.0bp);
  \node (x3) at (52.737bp,155.0bp) [draw=black,inner sep=2pt,circle] {};
  \coordinate (x33) at (24.737bp,116.5bp);
  \node (x4) at (144.74bp,187.3bp) [draw=black,inner sep=2pt,circle] {};
  \coordinate (x44) at (165.74bp,155.0bp);
  \draw [->] (x1) ..controls (127.91bp,150.02bp) and (136.07bp,139.3bp)  .. (139.74bp,129.0bp) .. controls (141.19bp,124.92bp) and (142.29bp,120.49bp)  .. (x11);
  \definecolor{strokecol}{rgb}{0.0,0.0,0.0};
  \pgfsetstrokecolor{strokecol}
  \draw (169.11bp,130.75bp) node {\_visit\_v\_1};
  \draw [->] 
    (x1) .. controls (115bp,145bp) and (105bp,135bp)  
          .. (100bp,125bp) .. controls (96bp,120bp) and (94bp,119.3bp)  
          .. (x2);
  \draw (122.11bp,130.75bp) node {ARG2};
  \draw [->] (x2) ..controls (91.737bp,79.865bp) and (91.737bp,78.109bp)  .. (x22);
  \draw (130.36bp,92.25bp) node {named(``Pairs'')};

  \draw [->] (x3) .. controls (57.154bp,155bp) and (75bp,145bp) .. (80bp,135bp) 
             .. controls (85bp,125bp) and (88bp,120bp) .. (x2);
  \draw (93.362bp,130.75bp) node {BV};
  \draw [->] (x3) ..controls (47.842bp,150.39bp) and (35.295bp,140.51bp)  .. (29.987bp,129.0bp) .. controls (28.091bp,124.89bp) and (26.808bp,120.32bp)  .. (x33);
  \draw (55.10bp,130.75bp) node {proper\_q};
  
  \draw [->] (x4) .. controls (139.68bp,184.9bp) and (130bp,175bp)  
             .. (126bp,165bp) .. controls (125bp,160bp) and (124.74bp,159bp)  
             .. (x1);
  \draw (145.81bp,174.25bp) node {ARG1};
  \draw [->] (x4) ..controls (149.42bp,185.17bp) and (157.19bp,182.75bp)  .. (160.74bp,177.5bp) .. controls (163.31bp,173.68bp) and (164.92bp,169.21bp)  .. (x44);
  
  \draw (193.78bp,174.25bp) node {\_often\_a\_1};
            \end{tikzpicture}
        }
        }
    \end{minipage}
    \caption{Syntactico-semantic representation of \textit{visit often Paris}}
    \label{fig:shrg_full_representation}
\end{figure}
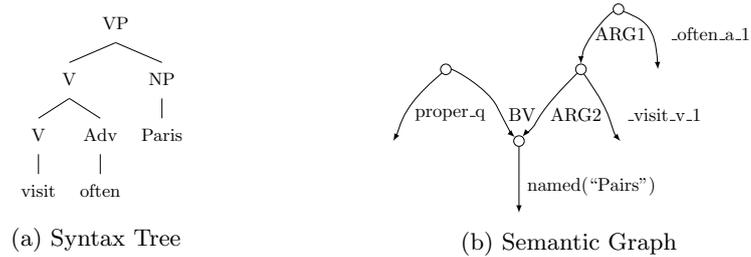

To illustrate this framework, consider the phrase visit often Paris. Figure~\ref{fig:shrg_full_representation} presents its full syntactic and semantic representations, while Table~\ref{tab:shrg_rules} decomposes the phrase into its constituent constructions. Each construction is specified by a paired set of rules: a CFG rule $A \to \beta$ capturing its syntactic structure, and a corresponding HRG rule $A \to G$ encoding its semantic composition.
These paired rules jointly model the syntax--semantics interface. For instance, when the syntactic component combines a verb and an adverb via the rule $V \to V + Adv$, the corresponding semantic rule defines how the adverb modifies the underlying event structure. By explicitly aligning syntactic configurations with their corresponding semantic operations, SHRG not only captures the structures characteristic of ESFL but also renders them systematic, principled, and directly comparable to those of standard English.

\begin{table}[h]
   \caption{Component rules for \textit{visit often Paris}.}
    \label{tab:shrg_rules}
    \begin{tabular}{@{}m{0.8cm} m{2.5cm} m{1.5cm} m{3.0cm}@{}} \toprule
        Index & SYN & \multicolumn{2}{l}{SEM} \\ \midrule
        \FilledCircled{1} & V $\rightarrow$ visit & V $\rightarrow$ & 
        \scalebox{0.6}{
            \begin{tikzpicture}[>=latex,line join=bevel]
                \node (node1) at (1.8bp,33.8bp) [draw,circle, inner sep=1.5pt,fill] {};
                \coordinate (node2) at (1.8bp,0.5bp);
                \draw [->] (node1) ..controls (1.8bp,27.979bp) and (1.8bp,18.64bp) .. (node2);
                \draw (29.3bp,16.5bp) node {\_visit\_v\_1};
            \end{tikzpicture}
        } \\ 
        \FilledCircled{2} & Adv $\rightarrow$ often & Adv $\rightarrow$ & 
        \scalebox{0.6}{
            \begin{tikzpicture}[>=latex,line join=bevel]
                \node (node1) at (1.8bp,33.8bp) [draw,circle, inner sep=1.5pt,fill] {};
                \coordinate (node2) at (1.8bp,0.5bp);
                \draw [->] (node1) ..controls (1.8bp,27.979bp) and (1.8bp,18.64bp) .. (node2);
                \draw (29.3bp,16.5bp) node {\_often\_a\_1};
            \end{tikzpicture}
        } \\ 
        \FilledCircled{3} & V $\rightarrow$ V+Adv & V $\rightarrow$ & 
        \scalebox{0.6}{
            \begin{tikzpicture}[>=latex,line join=bevel]
                \node (node1) at (22.5bp,68.4bp) [draw,circle, inner sep=1.5pt] {};
                \coordinate (node2) at (6.5bp,33.8bp);
                \node (node3) at (44.5bp,33.8bp) [draw,circle,inner sep=1.5pt,fill] {};
                \coordinate (node4) at (44.5bp,0.5bp);
                \draw [->] (node1) ..controls (17.636bp,66.286bp) and (13.5231bp,63.891bp) .. (9.5bp,58.6bp) .. controls (7.625bp,54.819bp) and (6.9113bp,49.765bp) .. (node2);
                \draw (19.5bp,51.1bp) node {Adv};
                \draw [->] (node1) ..controls (27.364bp,66.286bp) and (35.477bp,63.891bp) .. (39.5bp,58.6bp) .. controls (42.26bp,54.97bp) and (43.581bp,50.167bp) .. (node3);
                \draw (61.0bp,51.1bp) node {ARG1};
                \draw [->] (node3) ..controls (44.5bp,27.979bp) and (44.5bp,18.64bp) .. (node4);
                \draw (53.0bp,16.5bp) node {V};
            \end{tikzpicture}
        } \\
        \FilledCircled{4} & NP $\rightarrow$ Paris & NP $\rightarrow$ &
        \scalebox{0.6}{
            \begin{tikzpicture}[>=latex,line join=bevel]
                \node (node1) at (22.5bp,68.4bp) [draw,circle, inner sep=1.5pt] {};
                \coordinate (node2) at (0.5bp,33.8bp);
                \node (node3) at (44.5bp,33.8bp) [draw,circle,inner sep=1.5pt,fill] {};
                \coordinate (node4) at (44.5bp,0.5bp);
                \draw [->] (node1) ..controls (17.636bp,66.286bp) and (9.5231bp,63.891bp) .. (5.5bp,58.6bp) .. controls (2.625bp,54.819bp) and (1.3113bp,49.765bp) .. (node2);
                \draw (22.5bp,51.1bp) node {proper\_q};
                \draw [->] (node1) ..controls (27.364bp,66.286bp) and (35.477bp,63.891bp) .. (39.5bp,58.6bp) .. controls (42.26bp,54.97bp) and (43.581bp,50.167bp) .. (node3);
                \draw (53.0bp,51.1bp) node {BV};
                \draw [->] (node3) ..controls (44.5bp,27.979bp) and (44.5bp,18.64bp) .. (node4);
                \draw (82.0bp,16.5bp) node {named(``Paris'')};
            \end{tikzpicture}
        } \\ 
        \FilledCircled{5} & VP $\rightarrow$ V+NP & VP $\rightarrow$ & 
        \scalebox{0.6}{
            \begin{tikzpicture}[>=latex,line join=bevel]
                \node (node1) at (12.5bp,68.4bp) [draw,circle, inner sep=1.5pt] {};
                \coordinate (node2) at (0.5bp,33.8bp);
                \node (node3) at (25.5bp,33.8bp) [draw,circle,inner sep=1.5pt] {};
                \coordinate (node4) at (25.5bp,0.5bp);
                \draw [->] (node1) ..controls (9.4958bp,65.429bp) and (5.3665bp,62.358bp) .. (3.5bp,58.6bp) .. controls (1.4852bp,54.544bp) and (0.64086bp,49.639bp) .. (node2);
                \draw (8.5bp,51.1bp) node {V};
                \draw [->] (node1) ..controls (13.985bp,63.677bp) and (18.077bp,53.416bp) .. (node3);
                \draw (39.5bp,51.1bp) node {ARG2};
                \draw [->] (node3) ..controls (25.5bp,27.979bp) and (25.5bp,18.64bp) .. (node4);
                \draw (34.5bp,16.5bp) node {NP};
            \end{tikzpicture}
        } \\\bottomrule
    \end{tabular}
\end{table}

These SHRG rules also support the compositional derivation of syntactico-semantic representations, demonstrating the formal systematicity of this method. As exhibited in Figure \ref{fig:shrg_composition}, the syntactic derivation begins with terminal symbols, and production rules of the form $A \to \beta$ are applied recursively to derive non-terminals until the start symbol is reached.
Simultaneously, the semantic representation is built using parallel HRG rules $A \to G$. In this process, a hyperedge $e$ in $G$ is rewritten if it is labeled with a non-terminal symbol $n$ from the $C$. The edge $e$ is then removed, and a copy of the previously derived hypergraph $H$ whose left-hand side label is $n$, is inserted into $G$. The external nodes $X$ of $H$ are fused with the nodes originally connected by $e$, while all other hyperedges in $G$ remain unaffected.
\forestset{math tree/.style={for tree=math content},
           saved tree/.style={math tree,baseline, anchor=center},
           }
\newsavebox{\Ki}
\newsavebox{\Kii}
\newsavebox{\Vcont}
\newsavebox{\NPcont}
\newsavebox{\PPcont}
\savebox{\Ki}{\scalebox{0.6}{
 \begin{tikzpicture}[>=latex, line join=bevel]
      \node[draw, rounded corners=5pt, inner sep=5pt, line width=1.5pt] (background) {
            \begin{tikzpicture}[>=latex,line join=bevel,]
              \node (x1) at (124.74bp,155.0bp) [draw=black,circle,inner sep=2pt] {};
  \coordinate (x11) at (145.74bp,116.5bp);
  \node (x2) at (91.737bp,116.5bp) [draw=black,inner sep=2pt,circle] {};
  \coordinate (x22) at (91.737bp,78.0bp);
  \node (x3) at (52.737bp,155.0bp) [draw=black,inner sep=2pt,circle] {};
  \coordinate (x33) at (24.737bp,116.5bp);
  \node (x4) at (144.74bp,187.3bp) [draw=black,inner sep=2pt,circle] {};
  \coordinate (x44) at (165.74bp,155.0bp);
  \draw [->] (x1) ..controls (127.91bp,150.02bp) and (136.07bp,139.3bp)  .. (139.74bp,129.0bp) .. controls (141.19bp,124.92bp) and (142.29bp,120.49bp)  .. (x11);
  \definecolor{strokecol}{rgb}{0.0,0.0,0.0};
  \pgfsetstrokecolor{strokecol}
  \draw (169.11bp,130.75bp) node {\_visit\_v\_1};
  \draw [->] 
    (x1) .. controls (115bp,145bp) and (105bp,135bp)  
          .. (100bp,125bp) .. controls (96bp,120bp) and (94bp,119.3bp)  
          .. (x2);
  \draw (122.11bp,130.75bp) node {ARG2};
  \draw [->] (x2) ..controls (91.737bp,79.865bp) and (91.737bp,78.109bp)  .. (x22);
  \draw (130.36bp,92.25bp) node {named(``Pairs'')};

  \draw [->] (x3) .. controls (57.154bp,155bp) and (75bp,145bp) .. (80bp,135bp) 
             .. controls (85bp,125bp) and (88bp,120bp) .. (x2);
  \draw (93.362bp,130.75bp) node {BV};
  \draw [->] (x3) ..controls (47.842bp,150.39bp) and (35.295bp,140.51bp)  .. (29.987bp,129.0bp) .. controls (28.091bp,124.89bp) and (26.808bp,120.32bp)  .. (x33);
  \draw (55.10bp,130.75bp) node {proper\_q};
  
  \draw [->] (x4) .. controls (139.68bp,184.9bp) and (130bp,175bp)  
             .. (126bp,165bp) .. controls (125bp,160bp) and (124.74bp,159bp)  
             .. (x1);
  \draw (145.81bp,174.25bp) node {ARG1};
  \draw [->] (x4) ..controls (149.42bp,185.17bp) and (157.19bp,182.75bp)  .. (160.74bp,177.5bp) .. controls (163.31bp,173.68bp) and (164.92bp,169.21bp)  .. (x44);
  
  \draw (193.78bp,174.25bp) node {\_often\_a\_1};
          \node[draw, rounded corners=5pt, fill=white, inner sep=5pt, line width=1.5pt] at (210.6bp,196.5bp) {VP};
            \end{tikzpicture}
            };
        \end{tikzpicture}}}
\sbox{\Kii}{\scalebox{0.6}{
    \begin{tikzpicture}[>=latex, line join=bevel]
      \node[draw, rounded corners=5pt, inner sep=5pt, line width=1.5pt] (background) {
        \begin{tikzpicture}[>=latex,line join=bevel,]
 \node (node1) at (22.5bp,68.4bp) [draw,circle, inner sep=1.5pt] {};
  \coordinate (node2) at (-15.5bp,33.8bp);
  \node (node3) at (44.5bp,33.8bp) [draw,circle,inner sep=1.5pt,fill] {};
  \coordinate (node4) at (44.5bp,0.5bp);
  \draw [->] (node1) ..controls (17.636bp,66.286bp) and (9.5231bp,63.891bp)  .. (4.5bp,62.6bp) .. controls (-12.625bp,57.819bp) and (-9.3113bp,49.765bp)  .. (node2);
  \definecolor{strokecol}{rgb}{0.0,0.0,0.0};
  \pgfsetstrokecolor{strokecol}
  \draw (16.5bp,51.1bp) node {\_often\_a\_1};
  \draw [->] (node1) ..controls (27.364bp,66.286bp) and (35.477bp,63.891bp)  .. (39.5bp,58.6bp) .. controls (42.26bp,54.97bp) and (43.581bp,50.167bp)  .. (node3);
  \draw (61.0bp,51.1bp) node {ARG1};
  \draw [->] (node3) ..controls (44.5bp,27.979bp) and (44.5bp,18.64bp)  .. (node4);
  \draw (68.0bp,16.5bp) node {\_visit\_v\_1};
\node[draw, rounded corners=5pt, fill=white, inner sep=5pt, line width=1.5pt] at (85.0bp,70.5bp) {V};
\end{tikzpicture}
      };
    \end{tikzpicture}}}

\savebox{\Vcont}{\scalebox{0.6}{
 \begin{tikzpicture}[>=latex, line join=bevel]
      \node[draw, rounded corners=5pt, inner sep=5pt, line width=1.5pt] (background) {
            \begin{tikzpicture}[>=latex,line join=bevel,]
\node (node1) at (1.8bp,33.8bp) [draw,circle,fill, inner sep=1.5pt] {};
  \coordinate (node2) at (1.8bp,0.5bp);
  \draw [->] (node1) ..controls (1.8bp,27.979bp) and (1.8bp,18.64bp)  .. (node2);
  \definecolor{strokecol}{rgb}{0.0,0.0,0.0};
  \pgfsetstrokecolor{strokecol}
  \draw (29.3bp,16.5bp) node {\_visit\_v\_1};
  \node[draw, rounded corners=5pt, fill=white, inner sep=5pt, line width=1.5pt] at (52bp,40.5bp) {V};
\end{tikzpicture}
            };
        \end{tikzpicture}}}

\sbox{\NPcont}{\scalebox{0.6}{
    \begin{tikzpicture}[>=latex, line join=bevel]
      \node[draw, rounded corners=5pt, inner sep=5pt, line width=1.5pt] (background) {
        \begin{tikzpicture}[>=latex, line join=bevel]
          \node (node1) at (22.5bp,68.4bp) [draw,circle, inner sep=1.5pt] {};
          \coordinate (node2) at (0.5bp,33.8bp);
          \node (node3) at (44.5bp,33.8bp) [draw,circle,inner sep=1.5pt, fill] {};
          \coordinate (node4) at (44.5bp,0.5bp);
          
          \draw [->] (node1) ..controls (17.636bp,66.286bp) and (9.5231bp,63.891bp)  .. (5.5bp,58.6bp) .. controls (2.625bp,54.819bp) and (1.3113bp,49.765bp)  .. (node2);
          \draw (22.5bp,51.1bp) node {proper\_q};
          \draw [->] (node1) ..controls (27.364bp,66.286bp) and (35.477bp,63.891bp)  .. (39.5bp,58.6bp) .. controls (42.26bp,54.97bp) and (43.581bp,50.167bp)  .. (node3);
          \draw (53.0bp,51.1bp) node {BV};
          \draw [->] (node3) ..controls (44.5bp,27.979bp) and (44.5bp,18.64bp)  .. (node4);
          \draw (82.0bp,16.5bp) node {named(``Paris'')};
          \node[draw, rounded corners=5pt, fill=white, inner sep=5pt, line width=1.5pt] at (108.6bp,70.5bp) {NP};
        \end{tikzpicture}
      };
    \end{tikzpicture}}}

\sbox{\PPcont}{\scalebox{0.6}{
 \begin{tikzpicture}[>=latex, line join=bevel]
      \node[draw, rounded corners=5pt, inner sep=5pt, line width=1.5pt] (background) {
            \begin{tikzpicture}[>=latex,line join=bevel,]
\node (node1) at (1.8bp,33.8bp) [draw,circle,fill, inner sep=1.5pt] {};
  \coordinate (node2) at (1.8bp,0.5bp);
  \draw [->] (node1) ..controls (1.8bp,27.979bp) and (1.8bp,18.64bp)  .. (node2);
  \definecolor{strokecol}{rgb}{0.0,0.0,0.0};
  \pgfsetstrokecolor{strokecol}
  \draw (29.3bp,16.5bp) node {\_often\_a\_1};
  \node[draw, rounded corners=5pt, fill=white, inner sep=5pt, line width=1.5pt] at (52bp,40.5bp) {Adv};
\end{tikzpicture}
            };
        \end{tikzpicture}}}
\begin{figure}[htbp]
    \centering
    \scalebox{1}{
    \begin{forest}math tree
    [\usebox{\Ki}
        [\usebox{\Kii}
            [\usebox{\Vcont} [visit]] 
            [\usebox{\PPcont} [often]]
        ]
        [\usebox{\NPcont} [Paris]]   
    ]
    \node at (-2.9,-3.4)[circle, inner sep=2pt, fill=blue!60,text=white]{\scriptsize{1}};
    \node at (-0.5,-3.4)[circle, inner sep=2pt, fill=blue!60,text=white]{\scriptsize{2}};
    \node at (-1.6,-0.75)[ circle, inner sep=2pt, fill=blue!60,text=white]{\scriptsize{3}};
    \node at (1.6,-0.75)[circle, inner sep=2pt, fill=blue!60,text=white]{\scriptsize{4}};
    \node at (0,2.9)[circle, inner sep=2pt, fill=blue!60,text=white]{\scriptsize{5}};
\end{forest}}
    \caption{Synchronous construction of syntax and semantics for \textit{visit often Paris}.}
    \label{fig:shrg_composition}
\end{figure}
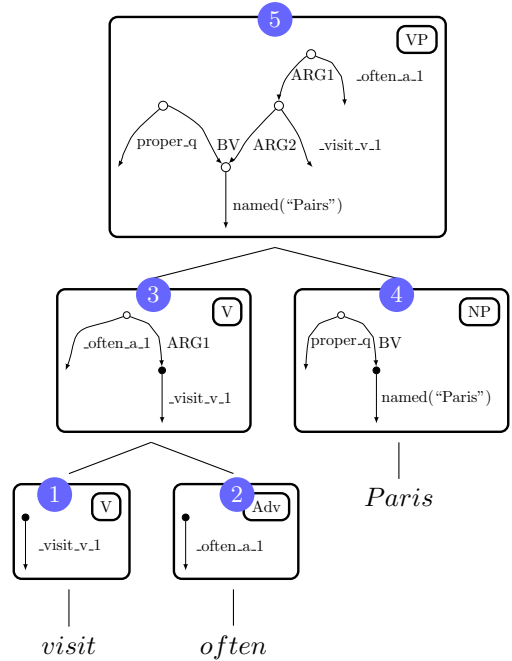

\section{ESFL SemBank}
\label{sembank}
We apply the SHRG-based constructivist approach to ESFL data, resulting a syntactico-semantic ESFL SemBank.
Generally speaking, its development involves three main steps:
\begin{itemize}
    \item \textbf{Review}: silver-standard semantic graphs from \citet{zhao-etal-2020-semantic} are manually reviewed as accepted or rejected. 
    \item \textbf{Revise}: extract SHRG rules and those of rejected graphs are manually revised to produce their accurate semantic representations.
    \item \textbf{Rebuild}: semantics of unparsable ESFL sentences in \citet{zhao-etal-2020-semantic} are manually rebuilt by selecting appropriate SHRG rules.
\end{itemize}
 
\subsection{Review}
{ers}, as well as their parsers.
We start by manually reviewing holistic silver-standard ESFL semantic graphs from \citet{zhao-etal-2020-semantic}, determining if they could be accepted or rejected.
This step establishes a reliable foundation for our sembank.

To be more specific, the review is conducted by a team comprising one Ph.D. student and two undergraduate students majoring in linguistics. To ensure consistency and accuracy, they firstly undergo training in English Resource Semantics \citep[ERS;][]{flickinger-etal-2014-towards}\footnote{ERS is a semantic framework developed in conjunction with English Resource Grammar \citep[ERG;][]{erg}, within Head-driven Phrase Structure Grammar \citep[HPSG;][]{pollard1994head}.}, which is the framework used to represent the previously parsed semantic graphs.
During annotation, DeepBank \citep{flickinger2012deepbank} --- a comprehensive ERS-based semantic resource known for handling a broad range of English linguistic phenomena --- is used as a reference. Annotators evaluate whether each graph from \citet{zhao-etal-2020-semantic} faithfully represents the intended semantics of the corresponding ESFL sentence and then assign one of three labels: accept, reject, or abandon based on its semantic fidelity. 

Following several rounds of training, comparison, and discussion, three annotators reach a high level of inter-annotator agreement (IAA). The annotation process is then streamlined: remaining instances are either cross-validated by two annotators or annotated by a single annotator.
Their annotation quality, as measured by percentage IAA, is reported in Table~\ref{tab:iaa-annotation}. The exceptionally high IAAs across annotators, ranging from 97\% to 99\%, consistently indicate strong consensus on the semantic interpretations of ESFL sentences.

\begin{table}[htbp]
    \centering
    \caption{Inter-annotator agreements (IAAs) in annotation.}
    \label{tab:iaa-annotation}
    \begin{tabular}{@{}ccc@{}}\toprule
         & ESFL & CEFSL \\\midrule
    Anno1-Anno2-Anno3  & 99.29 & 99.48  \\
    Anno1-Anno2  & 98.95 & 98.19  \\
    Anno1-Anno3  & 97.82 & 98.77  \\
    Anno2-Anno2  & 98.28 & 99.12  \\\bottomrule
    \end{tabular}
    
\end{table}
  
In total, 1567 ESFL and 2189 CESFL sentences are annotated. After removing records with inconsistent tags, 1543 ESFL and 2138 CESFL sentences remain. The final distribution of annotation labels is shown in Table~\ref{tab:accept-reject}: 46.92\% of ESFL graphs are accepted and 53.08\% rejected, while for CESFL, 61.04\% are accepted and 38.96\% rejected.

\begin{table}[htbp]
\caption{Numbers (\#num) and percentages (\#per) of valid sentences whose semantic graphs accepted (acc) or rejected (rej) by three, two, or one annotators.}
\label{tab:accept-reject}
\begin{tabular}{@{}lcccc@{}}
\toprule
\multirow{2}[3]{*}{} & \multicolumn{2}{c}{ESL} & \multicolumn{2}{c}{CESL} \\
\cmidrule(lr){2-3} \cmidrule(lr){4-5}
 &\#num & \#per & \#num & \#per \\
\midrule\midrule
Triple-acc & 64        &  45.71\%          &  129          &  66.84\%         \\
Triple-rej   & 76          & 54.28\%      &  64          &  33.16\%          \\
Triple-all   & 140          & 100.00\%      &  193          &  100.00\%          \\\hline
Double-acc  &  586          & 46.03\%          & 1051           & 59.65\%           \\
Double-rej    & 687           & 53.97\%             &  711           &  40.35\%          \\
Double-all    & 1273          & 100.00\%             &  1762           & 100.00\%          \\\hline
Single-acc    & 74         &  56.92\%          & 125           & 68.31\%          \\
Single-rej      & 56          & 43.08\%           & 58            & 31.69\%           \\
Single-all      & 130          & 100.00\%           & 183            & 100.00\%           \\\hline
Overall-acc         & 724         &  \underline{46.92\%}         & 1305            & \underline{61.04\%}  \\
Overall-rej             & 819           & \underline{53.08\%}           & 833            &\underline{38.96\%}   \\
Overall-all            & 1543           & 100.00\%         & 2138            &100.00\%   \\\bottomrule 
\end{tabular}
\end{table}

\subsection{Revise}
\label{modification}
We adopt the method in \citet{li-etal-2025-compositional} to extract SHRG rules from both accepted and rejected semantic graphs. Based on the induced rule inventory, we manually revise the compositional SHRG rules in rejected ESFL graphs to reflect their intended semantics.

\paragraph{Number} 
\label{number}
Number, as a grammatical category, is expressed by contrasts between singular and plural forms in English. However, ESFL may use various forms, particularly the zero form (see example \ref{ex4}), to decode number. This often leads to parsing errors with the ACE parser and thus results in inaccurate semantic interpretations. Since this phenomenon involves determiners and often lacks clear indicators of singularity or plurality, we address it with an SHRG rule that includes an abstract predicate, \texttt{udef\_q}, alongside other relevant SHRG rules (see Table \ref{tab:number}). These rules are then syntactically recombined to produce accurate semantic representations (see Figure \ref{fig:number}).

\es{\textit{I was impressed when I heard that she liked playing \textbf{puzzle} alone.}  \label{ex4}
}
\begin{table}[htbp]
     \caption{Original and modified SHRG rules for Example \ref{ex4}.}
    \label{tab:number}
    \begin{tabular}{ccccc}\toprule
          LHS & $\textrm{N}_{\textrm{ori}}$ & $\textrm{V}_{\textrm{ori}}$ & $\textrm{COMP}_{\textrm{mod}}$ & $\textrm{N}_{\textrm{mod}}$ \\\midrule
          Syn & that & puzzle & that & puzzle\\
          Sem & \scalebox{0.5}{
\begin{tikzpicture}[>=latex',line join=bevel,]
\node (node1) at (66.0bp,135.8bp) [draw,circle,inner sep=1.5pt] {};
  \coordinate (node2) at (27.0bp,100.0bp);
  \node (node3) at (105.0bp,100.0bp) [draw,circle, inner sep=1.5pt,fill] {};
  \coordinate (node4) at (105.0bp,68.0bp);
  \draw [->] (node1) ..controls (58.922bp,135.36bp) and (40.694bp,135.89bp)  .. (32.0bp,126.0bp) .. controls (28.902bp,122.47bp) and (27.084bp,118.05bp)  .. (node2);
  \definecolor{strokecol}{rgb}{0.0,0.0,0.0};
  \pgfsetstrokecolor{strokecol}
  \draw (68.0bp,118.5bp) node {\texttt{\_that\_q\_dem}};
  \draw [->] (node1) ..controls (73.078bp,135.36bp) and (91.306bp,135.89bp)  .. (100.0bp,126.0bp) .. controls (106.87bp,118.18bp) and (107.45bp,105.94bp)  .. (node3);
  \draw (115.5bp,118.5bp) node {BV};
  \draw [->] (node3) ..controls (105.0bp,78.434bp) and (105.0bp,61.674bp)  .. (node4);
  \draw (70.0bp,81.5bp) node {\texttt{generic\_entity}};
\end{tikzpicture}
} & \scalebox{0.6}{
\begin{tikzpicture}[>=latex,line join=bevel,]
\node (node1) at (1.8bp,33.8bp) [circle, inner sep=1.5pt,fill] {};
  \coordinate (node2) at (1.8bp,0.5bp);
  \draw [->] (node1) ..controls (1.8bp,27.979bp) and (1.8bp,18.64bp)  .. (node2);
  \definecolor{strokecol}{rgb}{0.0,0.0,0.0};
  \pgfsetstrokecolor{strokecol}
  \draw (29.3bp,16.5bp) node {\_puzzle\_v\_1};
\end{tikzpicture}
}  & $\varnothing$ & \scalebox{0.6}{
\begin{tikzpicture}[>=latex,line join=bevel,]
\node (node1) at (22.5bp,68.4bp) [draw,circle, inner sep=1.5pt] {};
  \coordinate (node2) at (0.5bp,33.8bp);
  \node (node3) at (44.5bp,33.8bp) [draw,circle,inner sep=1.5pt,fill] {};
  \coordinate (node4) at (44.5bp,0.5bp);
  \draw [->] (node1) ..controls (17.636bp,66.286bp) and (9.5231bp,63.891bp)  .. (5.5bp,58.6bp) .. controls (2.625bp,54.819bp) and (1.3113bp,49.765bp)  .. (node2);
  \definecolor{strokecol}{rgb}{0.0,0.0,0.0};
  \pgfsetstrokecolor{strokecol}
  \draw (22.5bp,51.1bp) node {udef\_q};
  \draw [->] (node1) ..controls (27.364bp,66.286bp) and (35.477bp,63.891bp)  .. (39.5bp,58.6bp) .. controls (42.26bp,54.97bp) and (43.581bp,50.167bp)  .. (node3);
  \draw (53.0bp,51.1bp) node {BV};
  \draw [->] (node3) ..controls (44.5bp,27.979bp) and (44.5bp,18.64bp)  .. (node4);
  \draw (15.0bp,16.5bp) node {\_puzzle\_n\_1};
\end{tikzpicture}
} \\\bottomrule
\end{tabular}
\end{table}
\tikzset{ersnode/.style={right, rounded rectangle, draw, font=\tt}} 
\tikzset{ 
  nodeA/.style={right, fill=blue!40, text=white, rounded rectangle,font=\tt},
  nodeB/.style={right, fill=brown!50, text=white, rounded rectangle,font=\tt},
  edgeA/.style={thick, rounded corners,>=stealth'},
  edgeB/.style={thick, color=brown!50, rounded corners,>=stealth'},
  labelA/.style={anchor=center,fill=white,font=\scriptsize},
  labeltiny/.style={anchor=center,fill=white,font=\tiny},
  alertedge/.style={thick, color=red, rounded corners,>=stealth'}
}
\tikzset{arrow2/.style={decoration={markings,mark=at position 1 with %
    {\arrow[scale=2.3,>=latex]{>}}},postaction={decorate}}}  
    
\begin{figure}[htbp]
\centering
\subcaptionbox{Original Syn}{
\begin{minipage}[t]{0.22\linewidth}
\centering
\scalebox{0.45}{
        \begin{tikzpicture}[level distance=30pt,sibling distance=0pt]
        \Tree 
        [.VP [.NP [.N that ] [.V \edge[roof]; {she liked playing} ] ]
        [.VP [.V puzzle ]
        [.PP \edge[roof]; {alone} ] ] ]  
        \end{tikzpicture}
        }
\end{minipage}%
}%
\subcaptionbox{Modified Syn}{
\begin{minipage}[t]{0.22\linewidth}
\centering
  \scalebox{0.45}{
        \begin{tikzpicture}[level distance=30pt,sibling distance=0pt]
        \Tree 
        [.VP [.COMP that ]
        [.VP [.VP [.V \edge[roof]; {she liked playing} ] [.N puzzle ] ]
        [.PP \edge[roof]; {alone} ] ] ] 
        \end{tikzpicture}}
\end{minipage}%
}
\subcaptionbox{Original Sem}{
\begin{minipage}[t]{0.22\linewidth}
\centering
 \scalebox{0.45}{
    \begin{tikzpicture}
        \node (By) [ersnode, ] {\_puzzle\_v\_1};
        \node (most) [ersnode, right = 1 of By ] {\_that\_q\_dem};
        \node (measure) [ersnode, below right = 1 and -0.5 of By ] {generic\_entity};

        \path[->] (By) edge [edgeA] node [labelA] {ARG1} (measure);
        \path[->] (most) edge [edgeA] node [labelA] {BV} (measure);
       
      \end{tikzpicture}
            } 
\end{minipage}%
}%
\subcaptionbox{Modified Sem}{
\begin{minipage}[t]{0.22\linewidth}
\centering
   \scalebox{0.45}{
    \begin{tikzpicture}
        \node (By) [ersnode, ] {udef\_q};
        \node (measure) [ersnode, below right = 1 and -0.5 of By ] {\_puzzle\_n\_1};
        \path[->] (By) edge [edgeA] node [labelA] {BV} (measure);
      \end{tikzpicture}
            } 
\end{minipage}%
}%
\caption{Relevant original and modified syntactic and  semantic analysis for Example \ref{ex4}.}
\label{fig:number}
\end{figure}
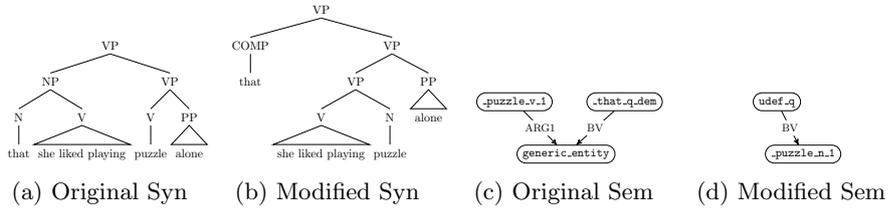

\paragraph{Case}
Case denotes the syntactic function of a nominal constituent within a sentence. In English, case is primarily limited to pronouns and the genitive case. Example \ref{ex5} is a case-related divergence in ESFL, where the genitive case is not explicitly marked by \textit{'s} but is instead implied through zero form and word order. To address this, we apply an SHRG rule (see Table \ref{tab:case}) that specifies the possessive relation. Figure \ref{fig:case} shows the modified result. 
\es{\textit{I think we know a \textbf{writer} life.} \label{ex5}
}
\begin{table}[htbp]
    \centering
    \caption{Original and modified SHRG rules for Example \ref{ex5}.}
    \begin{tabular}{ccc}\toprule
          LHS & $\textrm{NP}_{\textrm{ori}}$   & $\textrm{NP}_{\textrm{mod}}$ \\\midrule
          Syn & N + N   & N + NP  \\
          Sem & \scalebox{0.5}{
\begin{tikzpicture}[>=latex',line join=bevel,]
\node (x1) at (88.737bp,138.8bp) [draw,circle,inner sep=1.5pt] {};
  \coordinate (x11) at (24.737bp,96.5bp);
  \node (x2) at (84.737bp,96.5bp) [draw=black,fill=black,circle,inner sep=1.5pt] {};
  \node (x3) at (131.74bp,96.5bp) [draw,circle,inner sep=1.5pt] {};
  \coordinate (x22) at (82.737bp,58.0bp);
  \coordinate (x33) at (132.74bp,58.0bp);
  \draw [->] (x1) ..controls (78.189bp,137.87bp) and (37.251bp,137.64bp)  .. (28.987bp,129.0bp) .. controls (25.608bp,125.47bp) and (23.715bp,120.91bp)  .. (x11);
  \definecolor{strokecol}{rgb}{0.0,0.0,0.0};
  \pgfsetstrokecolor{strokecol}
  \draw (57.862bp,120.75bp) node {compound};
  \draw [->] (x1) ..controls (88.242bp,131.57bp) and (86.737bp,112.64bp)  .. (x2);
  \draw (105.6bp,120.75bp) node {ARG1};
  \draw [->] (x1) ..controls (96.342bp,138.24bp) and (113.78bp,138.32bp)  .. (122.74bp,129.0bp) .. controls (130.07bp,121.37bp) and (132.01bp,109.26bp)  .. (x3);
  \draw (148.34bp,120.75bp) node {ARG2};
  \draw [->] (x2) ..controls (84.568bp,79.865bp) and (84.064bp,63.109bp)  .. (x22);
  \draw (88.85bp,82.25bp) node {N};
  \draw [->] (x3) ..controls (131.82bp,79.865bp) and (132.07bp,63.109bp)  .. (x33);
  \draw (137.23bp,82.25bp) node {N};
\end{tikzpicture}
} & \scalebox{0.5}{
\begin{tikzpicture}[>=latex',line join=bevel,]
\node (x1) at (54.737bp,138.8bp) [draw,circle,inner sep=1.5pt] {};
  \coordinate (x11) at (24.737bp,96.5bp);
  \node (x2) at (53.737bp,96.5bp) [draw=black,fill=black,circle,inner sep=1.5pt] {};
  \node (x3) at (97.737bp,96.5bp) [draw,circle,inner sep=1.5pt] {};
  \coordinate (x22) at (49.737bp,58.0bp);
  \coordinate (x33) at (99.737bp,58.0bp);
  \draw [->] (x1) ..controls (47.999bp,137.56bp) and (35.611bp,136.32bp)  .. (29.737bp,129.0bp) .. controls (26.807bp,125.35bp) and (25.036bp,120.9bp)  .. (x11);
  \definecolor{strokecol}{rgb}{0.0,0.0,0.0};
  \pgfsetstrokecolor{strokecol}
  \draw (41.737bp,120.75bp) node {poss};
  \draw [->] (x1) ..controls (54.613bp,131.57bp) and (54.237bp,112.64bp)  .. (x2);
  \draw (72.171bp,120.75bp) node {ARG1};
  \draw [->] (x1) ..controls (62.598bp,138.35bp) and (80.611bp,138.65bp)  .. (89.737bp,129.0bp) .. controls (96.972bp,121.35bp) and (98.609bp,109.24bp)  .. (x3);
  \draw (115.02bp,120.75bp) node {ARG2};
  \draw [->] (x2) ..controls (53.398bp,79.865bp) and (52.391bp,63.109bp)  .. (x22);
  \draw (57.088bp,82.25bp) node {N};
  \draw [->] (x3) ..controls (97.907bp,79.865bp) and (98.411bp,63.109bp)  .. (x33);
  \draw (107.6bp,82.25bp) node {NP};
\end{tikzpicture}
}\\\bottomrule
    \end{tabular} 
    \label{tab:case}
\end{table}
\tikzset{ersnode/.style={right, rounded rectangle, draw, font=\tt}} 
\tikzset{ 
  nodeA/.style={right, fill=blue!40, text=white, rounded rectangle,font=\tt},
  nodeB/.style={right, fill=brown!50, text=white, rounded rectangle,font=\tt},
  edgeA/.style={thick, rounded corners,>=stealth'},
  edgeB/.style={thick, color=brown!50, rounded corners,>=stealth'},
  labelA/.style={anchor=center,fill=white,font=\scriptsize},
  labeltiny/.style={anchor=center,fill=white,font=\tiny},
  alertedge/.style={thick, color=red, rounded corners,>=stealth'}
}
\tikzset{arrow2/.style={decoration={markings,mark=at position 1 with %
    {\arrow[scale=2.3,>=latex]{>}}},postaction={decorate}}}  
    
\begin{figure}[H]
\centering
\subcaptionbox{Original Syn}{
\begin{minipage}[t]{0.18\textwidth}
\centering
\scalebox{0.45}{
        \begin{tikzpicture}[level distance=30pt,sibling distance=0pt]
        \Tree 
        [.NP 
            [.DET a ]
            [.NP [.N writer ]
            [.N life ] ] 
        ]  
        \end{tikzpicture}
        }
\end{minipage}%
}%
\subcaptionbox{Modified Syn}{
\begin{minipage}[t]{0.18\textwidth}
\centering
  \scalebox{0.45}{
        \begin{tikzpicture}[level distance=30pt,sibling distance=5pt]
        \Tree 
        [.NP 
            [.NP [.DET a ] [.N writer ] ]
            [.N life ] 
        ] 
        \end{tikzpicture}}
\end{minipage}%
}%
\subcaptionbox{Original Sem}{
\begin{minipage}[t]{0.24\textwidth}
\centering
 \scalebox{0.45}{
    \begin{tikzpicture}
        \node (By) [ersnode, ] {compound};
        \node (most) [ersnode, right = 1 of By ] {\_a\_q};
        \node (alone) [ersnode, left = 1 of By ] {udef\_q};
        \node (measure) [ersnode, below right = 1 and -0.4 of By ] {\_life\_n\_of};
        \node (growing) [ersnode, below left = 1 and -0.4 of By ] {\_writer\_n\_of};

        \path[->] (By) edge [edgeA] node [labelA] {ARG1} (measure);
        \path[->] (By) edge [edgeA] node [labelA] {ARG2} (growing);
        \path[->] (most) edge [edgeA] node [labelA] {BV} (measure);
        \path[->] (alone) edge [edgeA] node [labelA] {BV} (growing);
       
      \end{tikzpicture}
            } 
\end{minipage}%
}%
\subcaptionbox{Modified Sem}{
\begin{minipage}[t]{0.24\textwidth}
\centering
   \scalebox{0.45}{
    \begin{tikzpicture}
        \node (By) [ersnode, ] {poss};
        \node (most) [ersnode, right = 1 of By ] {def\_implicit\_q};
        \node (alone) [ersnode, left = 1 of By ] {\_a\_q};
        \node (measure) [ersnode, below right = 1 and 0.2 of By ] {\_life\_n\_of};
        \node (growing) [ersnode, below left = 1 and -0.3 of By ] {\_writer\_n\_of};

        \path[->] (By) edge [edgeA] node [labelA] {ARG1} (measure);
        \path[->] (By) edge [edgeA] node [labelA] {ARG2} (growing);
        \path[->] (most) edge [edgeA] node [labelA] {BV} (measure);
        \path[->] (alone) edge [edgeA] node [labelA] {BV} (growing);
       
      \end{tikzpicture}
            } 
\end{minipage}%
}%
\caption{Relevant original and modified syntactic and  semantic analysis for Example \ref{ex5}.}
\label{fig:case}
\end{figure}
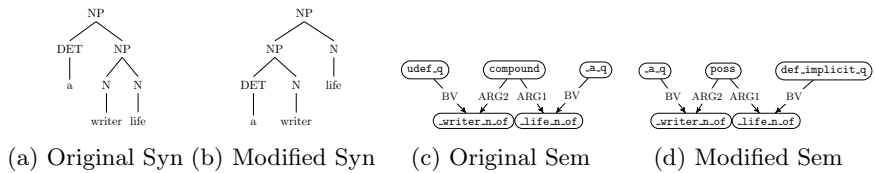

\paragraph{Tense \& Aspect}
Tense describes the time when an action or event occurs while aspect deals with how an action unfolds over time. These two grammatical categories are closely related as both of them encode temporal information.
As shown in Example \ref{ex6}, ESFL may employ grammatical means distinct from English to convey tense and aspect.
Figure \ref{fig:tense-aspect} further illustrates how the zero-form verb \textit{give} introduces ambiguity for the ACE parser, preventing it from correctly identifying the perfect aspect.
To address this diversity, we apply SHRG rules, as detailed in Table \ref{tab:tense-aspect}, to make targeted adjustments.
\es{\textit{I hope I \textbf{have give to you the information} you needed.} \label{ex6}}
\begin{table}[htbp]
    \caption{Original and modified SHRG rules for Example \ref{ex6}.}
    \label{tab:tense-aspect}
    \begin{tabular}{ccccc}\toprule
          LHS & $\textrm{V}_{\textrm{ori}}$ & $\textrm{VP}_{\textrm{ori}}$ & $\textrm{V}_{\textrm{mod}}$ & $\textrm{VP}_{\textrm{mod}}$\\\midrule
          Syn & have & V + NP & have & V + NP \\
          Sem & \scalebox{0.6}{
\begin{tikzpicture}[>=latex,line join=bevel,]
\node (node1) at (1.8bp,33.8bp) [circle, inner sep=1.5pt,fill] {};
  \coordinate (node2) at (1.8bp,0.5bp);
  \draw [->] (node1) ..controls (1.8bp,27.979bp) and (1.8bp,18.64bp)  .. (node2);
  \definecolor{strokecol}{rgb}{0.0,0.0,0.0};
  \pgfsetstrokecolor{strokecol}
  \draw (33.3bp,16.5bp) node {\_have\_v\_cause};
\end{tikzpicture}
} & \scalebox{0.6}{

\begin{tikzpicture}[>=latex,line join=bevel,]
\node (node1) at (12.5bp,68.4bp) [draw,circle, inner sep=1.5pt,fill] {};
  \coordinate (node2) at (0.5bp,33.8bp);
  \node (node3) at (25.5bp,33.8bp) [draw,circle,inner sep=1.5pt] {};
  \coordinate (node4) at (25.5bp,0.5bp);
  \draw [->] (node1) ..controls (9.4958bp,65.429bp) and (5.3665bp,62.358bp)  .. (3.5bp,58.6bp) .. controls (1.4852bp,54.544bp) and (0.64086bp,49.639bp)  .. (node2);
  \definecolor{strokecol}{rgb}{0.0,0.0,0.0};
  \pgfsetstrokecolor{strokecol}
  \draw (8.5bp,51.1bp) node {V};
  \draw [->] (node1) ..controls (13.985bp,63.677bp) and (18.077bp,53.416bp)  .. (node3);
  \draw (39.5bp,51.1bp) node {ARG1};
  \draw [->] (node3) ..controls (25.5bp,27.979bp) and (25.5bp,18.64bp)  .. (node4);
  \draw (34.5bp,16.5bp) node {NP};
\end{tikzpicture}
} & $\varnothing$ & \scalebox{0.6}{
\begin{tikzpicture}[>=latex,line join=bevel,]
\node (node1) at (12.5bp,68.4bp) [draw,circle, inner sep=1.5pt,fill] {};
  \coordinate (node2) at (0.5bp,33.8bp);
  \node (node3) at (25.5bp,33.8bp) [draw,circle,inner sep=1.5pt] {};
  \coordinate (node4) at (25.5bp,0.5bp);
  \draw [->] (node1) ..controls (9.4958bp,65.429bp) and (5.3665bp,62.358bp)  .. (3.5bp,58.6bp) .. controls (1.4852bp,54.544bp) and (0.64086bp,49.639bp)  .. (node2);
  \definecolor{strokecol}{rgb}{0.0,0.0,0.0};
  \pgfsetstrokecolor{strokecol}
  \draw (8.5bp,51.1bp) node {V};
  \draw [->] (node1) ..controls (13.985bp,63.677bp) and (18.077bp,53.416bp)  .. (node3);
  \draw (39.5bp,51.1bp) node {ARG2};
  \draw [->] (node3) ..controls (25.5bp,27.979bp) and (25.5bp,18.64bp)  .. (node4);
  \draw (34.5bp,16.5bp) node {NP};
\end{tikzpicture}
}  \\\bottomrule
    \end{tabular} 
\end{table}
\tikzset{ersnode/.style={right, rounded rectangle, draw, font=\tt}} 
\tikzset{ 
  nodeA/.style={right, fill=blue!40, text=white, rounded rectangle,font=\tt},
  nodeB/.style={right, fill=brown!50, text=white, rounded rectangle,font=\tt},
  edgeA/.style={thick, rounded corners,>=stealth'},
  edgeB/.style={thick, color=brown!50, rounded corners,>=stealth'},
  labelA/.style={anchor=center,fill=white,font=\scriptsize},
  labeltiny/.style={anchor=center,fill=white,font=\tiny},
  alertedge/.style={thick, color=red, rounded corners,>=stealth'}
}
\tikzset{arrow2/.style={decoration={markings,mark=at position 1 with %
    {\arrow[scale=2.3,>=latex]{>}}},postaction={decorate}}}  
    
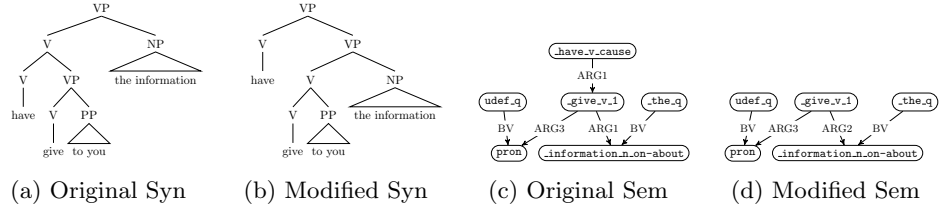
\begin{figure}[htbp]
\centering
\subcaptionbox{Original Syn}{
\begin{minipage}[t]{0.23\textwidth}
\centering
\scalebox{0.45}{
        \begin{tikzpicture}[level distance=30pt,sibling distance=0pt]
        \Tree 
        [.VP 
        [.V 
            [.V have ]
            [.VP [.V give ]
            [.PP \edge[roof]; {to you} ] ] 
        ]
        [.NP \edge[roof]; {the information} ]
        ]  
        \end{tikzpicture}
        }
\end{minipage}%
}%
\subcaptionbox{Modified Syn}{
\begin{minipage}[t]{0.23\textwidth}
\centering
  \scalebox{0.45}{
        \begin{tikzpicture}[level distance=30pt,sibling distance=0pt]
        \Tree 
        [.VP [.V have ]
        [.VP 
            [.V [.V give ]
            [.PP \edge[roof]; {to you} ] ] 
            [.NP \edge[roof]; {the information} ]
        ]
        ] 
        \end{tikzpicture}}
\end{minipage}%
}%
\subcaptionbox{Original Sem}{
\begin{minipage}[t]{0.24\textwidth}
\centering
 \scalebox{0.45}{
    \begin{tikzpicture}
        \node (By) [ersnode, ] {\_give\_v\_1};
        \node (most) [ersnode, right = 0.5 of By ] {\_the\_q};
         \node (have) [ersnode, above = 1 of By ] {\_have\_v\_cause};
        \node (alone) [ersnode, left = 1 of By ] {udef\_q};
        \node (measure) [ersnode, below right = 1 and -2.0 of By ] {\_information\_n\_on-about};
        \node (growing) [ersnode, below left = 1 and 1.5 of By ] {pron};

        \path[->] (By) edge [edgeA] node [labelA] {ARG1} (measure);
        \path[->] (have) edge [edgeA] node [labelA] {ARG1} (By);
        \path[->] (By) edge [edgeA] node [labelA] {ARG3} (growing);
        \path[->] (most) edge [edgeA] node [labelA] {BV} (measure);
        \path[->] (alone) edge [edgeA] node [labelA] {BV} (growing);
       
      \end{tikzpicture}
            } 
\end{minipage}%
}%
\subcaptionbox{Modified Sem}{
\begin{minipage}[t]{0.24\textwidth}
\centering
   \scalebox{0.45}{
    \begin{tikzpicture}
 eds
        \node (By) [ersnode, ] {\_give\_v\_1};
        \node (most) [ersnode, right = 1 of By ] {\_the\_q};
        \node (alone) [ersnode, left = 0.5 of By ] {udef\_q};
        \node (measure) [ersnode, below right = 1 and -2.0 of By ] {\_information\_n\_on-about};
        \node (growing) [ersnode, below left = 1 and 1.5 of By ] {pron};

        \path[->] (By) edge [edgeA] node [labelA] {ARG2} (measure);
        \path[->] (By) edge [edgeA] node [labelA] {ARG3} (growing);
        \path[->] (most) edge [edgeA] node [labelA] {BV} (measure);
        \path[->] (alone) edge [edgeA] node [labelA] {BV} (growing);       
      \end{tikzpicture}
            } 
\end{minipage}%
}%
\caption{Relevant original and modified syntactic and  semantic analysis for Example \ref{ex6}.}
\label{fig:tense-aspect}
\end{figure}

\paragraph{Voice}
Voice refers to the relationship between a verb and its participants, with active and passive being the most commonly used. In English, this distinction is typically conveyed through word order and inflectional markers. However, as illustrated in Example \ref{ex7}, ESFL exhibits notable variations in how voice is expressed, such as the zero-form verb \textit{mind}. Figure \ref{fig:voice} demonstrates how this creates challenges for the ACE parser in correctly identifying the passive voice. To address this issue, we apply SHRG rules outlined in Table \ref{tab:voice} to make the necessary adjustments.

\es{\textit{\textbf{It wouldn't be mind} if the restaurant had pressed in order to eat or drink something.} \label{ex7}}

\begin{table}[htbp]
    \caption{Original and modified SHRG rules for Example \ref{ex7}.}\label{tab:voice}
    \begin{tabular}{lllll}
\toprule
LHS & $\textrm{V}_{\textrm{ori}}$ & $\textrm{N}_{\textrm{ori}}$ & $\textrm{VP}_{\textrm{ori}}$ & $\textrm{S}_{\textrm{ori}}$ \\ \midrule
Syn & be & mind & V + N & NP + VP \\ \addlinespace
Sem & 
\scalebox{0.6}{
\begin{tikzpicture}[>=latex,line join=bevel]
  \node (node1) at (1.8bp,33.8bp) [circle, inner sep=1.5pt,fill] {};
  \coordinate (node2) at (1.8bp,0.5bp);
  \draw [->] (node1) ..controls (1.8bp,27.979bp) and (1.8bp,18.64bp) .. (node2);
  \draw (25.3bp,16.5bp) node {\_be\_v\_id};
\end{tikzpicture}
} & 
\scalebox{0.56}{
\begin{tikzpicture}[>=latex,line join=bevel,]
\node (node1) at (22.5bp,68.4bp) [draw,circle, inner sep=1.5pt] {};
  \coordinate (node2) at (0.5bp,33.8bp);
  \node (node3) at (44.5bp,33.8bp) [draw,circle,inner sep=1.5pt,fill] {};
  \coordinate (node4) at (44.5bp,0.5bp);
  \draw [->] (node1) ..controls (17.636bp,66.286bp) and (9.5231bp,63.891bp)  .. (5.5bp,58.6bp) .. controls (2.625bp,54.819bp) and (1.3113bp,49.765bp)  .. (node2);
  \definecolor{strokecol}{rgb}{0.0,0.0,0.0};
  \pgfsetstrokecolor{strokecol}
  \draw (22.5bp,51.1bp) node {udef\_q};
  \draw [->] (node1) ..controls (27.364bp,66.286bp) and (35.477bp,63.891bp)  .. (39.5bp,58.6bp) .. controls (42.26bp,54.97bp) and (43.581bp,50.167bp)  .. (node3);
  \draw (53.0bp,51.1bp) node {BV};
  \draw [->] (node3) ..controls (44.5bp,27.979bp) and (44.5bp,18.64bp)  .. (node4);
  \draw (75.0bp,16.5bp) node {\_mind\_n\_1};
\end{tikzpicture}
} & 
\scalebox{0.56}{
\begin{tikzpicture}[>=latex,line join=bevel]
  \node (node1) at (12.5bp,68.4bp) [draw,circle, inner sep=1.5pt, fill] {};
  \coordinate (node2) at (0.5bp,33.8bp);
  \node (node3) at (25.5bp,33.8bp) [draw,circle,inner sep=1.5pt] {};
  \coordinate (node4) at (25.5bp,0.5bp);
  \draw [->] (node1) ..controls (9.4958bp,65.429bp) and (5.3665bp,62.358bp) .. (3.5bp,58.6bp) .. controls (1.4852bp,54.544bp) and (0.64086bp,49.639bp) .. (node2);
  \draw (8.5bp,51.1bp) node {V};
  \draw [->] (node1) ..controls (13.985bp,63.677bp) and (18.077bp,53.416bp) .. (node3);
  \draw (39.5bp,51.1bp) node {ARG2};
  \draw [->] (node3) ..controls (25.5bp,27.979bp) and (25.5bp,18.64bp) .. (node4);
  \draw (32.5bp,16.5bp) node {N};
\end{tikzpicture}
} & 
\scalebox{0.56}{
\begin{tikzpicture}[>=latex,line join=bevel]
  \node (node1) at (12.5bp,68.4bp) [draw,circle, inner sep=1.5pt] {};
  \coordinate (node2) at (0.5bp,33.8bp);
  \node (node3) at (25.5bp,33.8bp) [draw,circle,inner sep=1.5pt] {};
  \coordinate (node4) at (25.5bp,0.5bp);
  \draw [->] (node1) ..controls (9.4958bp,65.429bp) and (5.3665bp,62.358bp) .. (3.5bp,58.6bp) .. controls (1.4852bp,54.544bp) and (0.64086bp,49.639bp) .. (node2);
  \draw (8.5bp,51.1bp) node {VP};
  \draw [->] (node1) ..controls (13.985bp,63.677bp) and (18.077bp,53.416bp) .. (node3);
  \draw (39.5bp,51.1bp) node {ARG1};
  \draw [->] (node3) ..controls (25.5bp,27.979bp) and (25.5bp,18.64bp) .. (node4);
  \draw (32.5bp,16.5bp) node {NP};
\end{tikzpicture}
} \\\midrule
LHS & $\textrm{V}_{\textrm{mod}1}$ & $\textrm{V}_{\textrm{mod}2}$ & $\textrm{VP/NP}_{\textrm{mod}}$ & $\textrm{S}_{\textrm{mod}}$ \\ \midrule
Syn & be & mind & V + V & NP + VP/NP \\ \addlinespace
Sem & $\varnothing$ & 
\scalebox{0.56}{
\begin{tikzpicture}[>=latex,line join=bevel]
  \node (node1) at (1.8bp,33.8bp) [circle, inner sep=1.5pt,fill] {};
  \coordinate (node2) at (1.8bp,0.5bp);
  \draw [->] (node1) ..controls (1.8bp,27.979bp) and (1.8bp,18.64bp) .. (node2);
  \draw (25.3bp,16.5bp) node {\_mind\_v\_1};
\end{tikzpicture}
} & 
\scalebox{0.56}{
\begin{tikzpicture}[>=latex',line join=bevel]
  \node (x1) at (39.737bp,138.8bp) [draw=black,fill=black,circle,inner sep=1.5pt] {};
  \coordinate (x11) at (24.737bp,96.5bp);
  \node (x2) at (54.737bp,96.5bp) [draw=black,fill=black,circle,inner sep=1.5pt] {};
  \coordinate (x22) at (54.737bp,62.0bp);
  \draw [->] (x1) ..controls (36.396bp,135.91bp) and (31.806bp,132.93bp) .. (29.737bp,129.0bp) .. controls (27.63bp,125.0bp) and (26.253bp,120.48bp) .. (x11);
  \draw (41.737bp,120.75bp) node {parg};
  \draw [->] (x1) ..controls (43.078bp,135.91bp) and (47.669bp,132.93bp) .. (49.737bp,129.0bp) .. controls (54.47bp,120.01bp) and (55.521bp,108.44bp) .. (x2);
  \draw (72.258bp,120.75bp) node {ARG1};
  \draw [->] (x2) ..controls (54.737bp,79.865bp) and (54.737bp,63.109bp) .. (x22);
  \draw (59.612bp,82.25bp) node {V};
\end{tikzpicture}
} & 
\scalebox{0.56}{
\begin{tikzpicture}[>=latex',line join=bevel]
  \node (x1) at (24.737bp,151.6bp) [draw,circle,inner sep=1.5pt] {};
  \node (x2) at (24.737bp,100.3bp) [draw,inner sep=1.5pt,circle] {};
  \node (x3) at (65.737bp,122.85bp) [draw,inner sep=1.5pt,circle] {};
  \coordinate (x22) at (24.737bp,62.0bp);
  \draw [->] (x1) ..controls (24.079bp,145.67bp) and (21.606bp,140.73bp) .. (22.487bp,134.6bp) .. controls (22.718bp,127.759bp) and (23.215bp,117.141bp) .. (x2);
  \draw (40.112bp,120.85bp) node {ARG2};
  \draw [->] (x1) ..controls (25.639bp,146.98bp) and (27.704bp,145.36bp) .. (30.737bp,143.6bp) .. controls (33.948bp,140.5bp) and (44.627bp,133.31bp) .. (x3);
  \draw (66.273bp,137.35bp) node {VP/NP};
  \draw [->] (x2) ..controls (24.737bp,64.993bp) and (24.737bp,62.679bp) .. (x22);
  \draw (33.362bp,77.25bp) node {NP};
  \draw [->] (x3) ..controls (57.736bp,109.4bp) and (55.598bp,107.79bp) .. (53.737bp,106.6bp) .. controls (50.915bp,104.564bp) and (48.055bp,102.554bp) .. (x2);
  \draw (54.27bp,94.35bp) node {ARG2};
\end{tikzpicture}
} \\ \bottomrule
\end{tabular}
\end{table}
\tikzset{ersnode/.style={right, rounded rectangle, draw, font=\tt}} 
\tikzset{ 
  nodeA/.style={right, fill=blue!40, text=white, rounded rectangle,font=\tt},
  nodeB/.style={right, fill=brown!50, text=white, rounded rectangle,font=\tt},
  edgeA/.style={thick, rounded corners,>=stealth'},
  edgeB/.style={thick, color=brown!50, rounded corners,>=stealth'},
  labelA/.style={anchor=center,fill=white,font=\scriptsize},
  labeltiny/.style={anchor=center,fill=white,font=\tiny},
  alertedge/.style={thick, color=red, rounded corners,>=stealth'}
}
\tikzset{arrow2/.style={decoration={markings,mark=at position 1 with %
    {\arrow[scale=2.3,>=latex]{>}}},postaction={decorate}}}  
    
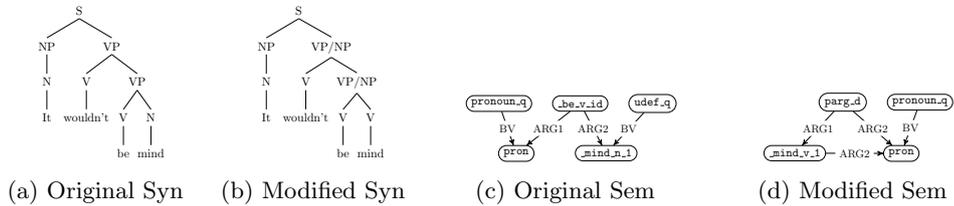
\begin{figure}[htbp]
\centering
\subcaptionbox{Original Syn}{
\begin{minipage}[t]{0.21\textwidth}
\centering
\scalebox{0.45}{
        \begin{tikzpicture}[level distance=30pt,sibling distance=0pt]
        \Tree 
        [.S [.NP [.N It ] ]
        [.VP [.V wouldn't ]
        [.VP [.V be ] [.N mind ] ] ] 
        ] 
        \end{tikzpicture}
        }
\end{minipage}%
}%
\subcaptionbox{Modified Syn}{
\begin{minipage}[t]{0.21\textwidth}
\centering
  \scalebox{0.45}{
        \begin{tikzpicture}[level distance=30pt,sibling distance=0pt]
        \Tree 
        [.S [.NP [.N It ] ]
        [.VP/NP [.V wouldn't ]
        [.VP/NP [.V be ] [.V mind ] ] ] 
        ] 
        \end{tikzpicture}}
\end{minipage}%
}%
\subcaptionbox{Original Sem}{
\begin{minipage}[t]{0.28\textwidth}
\centering
 \scalebox{0.45}{
    \begin{tikzpicture}
        \node (By) [ersnode, ] {\_be\_v\_id};
        \node (most) [ersnode, right = 0.6 of By ] {udef\_q};
        \node (quan) [ersnode, left = 0.6 of By] {pronoun\_q};
        \node (measure) [ersnode, below right = 1 and -0.5 of By ] {\_mind\_n\_1};
        \node (growing) [ersnode, below left = 1 and 1 of By ] {pron};

        \path[->] (By) edge [edgeA] node [labelA] {ARG2} (measure);
        \path[->] (most) edge [edgeA] node [labelA] {BV} (measure);
        \path[->] (By) edge [edgeA] node [labelA] {ARG1} (growing);
        \path[->] (quan) edge [edgeA] node [labelA] {BV} (growing);
       
      \end{tikzpicture}
            } 
\end{minipage}%
}%
\subcaptionbox{Modified Sem}{
\begin{minipage}[t]{0.28\textwidth}
\centering
   \scalebox{0.45}{
    \begin{tikzpicture}
        \node (By) [ersnode, ] {parg\_d};
        \node (most) [ersnode, right = 0.6 of By ] {pronoun\_q};
        \node (measure) [ersnode, below right = 1 and 1 of By ] {pron};
        \node (growing) [ersnode, below left = 1 and 0.3 of By ] {\_mind\_v\_1};

        \path[->] (By) edge [edgeA] node [labelA] {ARG2} (measure);
        \path[->] (most) edge [edgeA] node [labelA] {BV} (measure);
        \path[->] (By) edge [edgeA] node [labelA] {ARG1} (growing);
        \path[->] (growing) edge [edgeA] node [labelA] {ARG2} (measure);
      \end{tikzpicture}
            } 
\end{minipage}%
}%
\caption{Relevant original and modified syntactic and semantic analysis for Example \ref{ex7}.}
\label{fig:voice}
\end{figure}

\paragraph{Person}
Person is a grammatical category that distinguishes forms of different participants in an event. It could affect verb inflections: in English, the form of a verb typically depends on the person and number of its subject. However, in ESFL, this agreement may sometimes be conveyed through a zero form, as shown in Example \ref{ex8}. The absence of overt marking can lead to misidentification by the ACE parser. For instance, as illustrated in Figure \ref{fig:person}, the phrase \textit{have good meaning} is incorrectly analyzed as an imperative sentence. We address this issue by applying the SHRG rules in Table \ref{tab:person}.
\es{\textit{It looks nice and \textbf{have} good meaning.} \label{ex8}}
\begin{table}[htbp]
    \caption{Original and modified SHRG rules for Example \ref{ex8}.}
    \label{tab:person}
    \begin{tabular}{ccccc}\toprule
          LHS & $\textrm{S}_{\textrm{ori}}$ & $\textrm{S}_{\textrm{ori}}$ & $\textrm{VP}_{\textrm{ori}}$ & $\textrm{VP}_{\textrm{ori}}$ \\\midrule
          Syn & Conj + S & S + S & Conj + VP & VP + VP\\
          Sem & \scalebox{0.6}{
\begin{tikzpicture}[>=latex,line join=bevel,]
\node (node1) at (22.5bp,68.4bp) [draw,circle, inner sep=1.5pt, fill] {};
  \coordinate (node2) at (2.5bp,33.8bp);
  \node (node3) at (44.5bp,33.8bp) [draw,circle,inner sep=1.5pt] {};
  \coordinate (node4) at (44.5bp,0.5bp);
  \draw [->] (node1) ..controls (17.636bp,66.286bp) and (9.5231bp,63.891bp)  .. (5.5bp,58.6bp) .. controls (2.625bp,54.819bp) and (2.5113bp,49.765bp)  .. (node2);
  \definecolor{strokecol}{rgb}{0.0,0.0,0.0};
  \pgfsetstrokecolor{strokecol}
  \draw (19.5bp,51.1bp) node {Conj};
  \draw [->] (node1) ..controls (27.364bp,66.286bp) and (35.477bp,63.891bp)  .. (39.5bp,58.6bp) .. controls (42.26bp,54.97bp) and (43.581bp,50.167bp)  .. (node3);
  \draw (62.0bp,55.1bp) node {R-IND};
  \draw (64.0bp,45.1bp) node {R-HND};
  \draw [->] (node3) ..controls (44.5bp,27.979bp) and (44.5bp,18.64bp)  .. (node4);
  \draw (50.0bp,16.5bp) node {S};
\end{tikzpicture}
} & \scalebox{0.6}{
\begin{tikzpicture}[>=latex,line join=bevel,]
\node (node1) at (12.5bp,68.4bp) [draw,circle, inner sep=1.5pt, fill] {};
  \coordinate (node2) at (0.5bp,33.8bp);
  \node (node3) at (25.5bp,33.8bp) [draw,circle,inner sep=1.5pt] {};
  \coordinate (node4) at (25.5bp,0.5bp);
  \draw [->] (node1) ..controls (9.4958bp,65.429bp) and (5.3665bp,62.358bp)  .. (3.5bp,58.6bp) .. controls (1.4852bp,54.544bp) and (0.64086bp,49.639bp)  .. (node2);
  \definecolor{strokecol}{rgb}{0.0,0.0,0.0};
  \pgfsetstrokecolor{strokecol}
  \draw (8.5bp,51.1bp) node {S};
  \draw [->] (node1) ..controls (13.985bp,63.677bp) and (18.077bp,53.416bp)  .. (node3);
  \draw (39.5bp,55.1bp) node {L-IND};
  \draw (41.5bp,45.1bp) node {L-HND};
  \draw [->] (node3) ..controls (25.5bp,27.979bp) and (25.5bp,18.64bp)  .. (node4);
  \draw (32.5bp,16.5bp) node {S};
\end{tikzpicture}
}  & \scalebox{0.6}{
\begin{tikzpicture}[>=latex,line join=bevel,]
\node (node1) at (22.5bp,68.4bp) [draw,circle, inner sep=1.5pt, fill] {};
  \coordinate (node2) at (0.5bp,33.8bp);
  \node (node3) at (44.5bp,33.8bp) [draw,circle,inner sep=1.5pt,fill] {};
  \coordinate (node4) at (44.5bp,0.5bp);
  \draw [->] (node1) ..controls (17.636bp,66.286bp) and (9.5231bp,63.891bp)  .. (5.5bp,58.6bp) .. controls (2.625bp,54.819bp) and (1.3113bp,49.765bp)  .. (node2);
  \definecolor{strokecol}{rgb}{0.0,0.0,0.0};
  \pgfsetstrokecolor{strokecol}
  \draw (19.5bp,51.1bp) node {Conj};
  \draw [->] (node1) ..controls (27.364bp,66.286bp) and (35.477bp,63.891bp)  .. (39.5bp,58.6bp) .. controls (42.26bp,54.97bp) and (43.581bp,50.167bp)  .. (node3);
  \draw (62.0bp,55.1bp) node {R-IND};
  \draw (64.0bp,45.1bp) node {R-HND};
  \draw [->] (node3) ..controls (44.5bp,27.979bp) and (44.5bp,18.64bp)  .. (node4);
  \draw (52.0bp,16.5bp) node {VP};
\end{tikzpicture}
} & \scalebox{0.6}{
\begin{tikzpicture}[>=latex,line join=bevel,]
\node (node1) at (12.5bp,68.4bp) [draw,circle, inner sep=1.5pt] {};
  \coordinate (node2) at (0.5bp,33.8bp);
  \node (node22) at (0.5bp,33.0bp) [draw,circle,inner sep=1.5pt,fill] {};
  \node (node3) at (25.5bp,33.8bp) [draw,circle,inner sep=1.5pt,fill] {};
  \coordinate (node4) at (25.5bp,0.5bp);
  \coordinate (node44) at (0.5bp,0.5bp);
  \draw [->] (node1) ..controls (9.4958bp,65.429bp) and (5.3665bp,62.358bp)  .. (3.5bp,58.6bp) .. controls (1.4852bp,54.544bp) and (0.64086bp,49.639bp)  .. (node2);
  \definecolor{strokecol}{rgb}{0.0,0.0,0.0};
  \pgfsetstrokecolor{strokecol}
  \draw (8.5bp,51.1bp) node {VP};
  \draw [->] (node1) ..controls (13.985bp,63.677bp) and (18.077bp,53.416bp)  .. (node3);
  \draw (39.5bp,55.1bp) node {L-IND};
  \draw (41.5bp,45.1bp) node {L-HND};
  \draw [->] (node3) ..controls (25.5bp,27.979bp) and (25.5bp,18.64bp)  .. (node4);
  \draw (32.5bp,16.5bp) node {VP};
  \draw [->] (node22) ..controls (0.5bp,23.677bp) and (0.5bp,14.416bp)  .. (node44);
\end{tikzpicture}
}\\\bottomrule
    \end{tabular}
\end{table}
\tikzset{ersnode/.style={right, rounded rectangle, draw, font=\tt}} 
\tikzset{ 
  nodeA/.style={right, fill=blue!40, text=white, rounded rectangle,font=\tt},
  nodeB/.style={right, fill=brown!50, text=white, rounded rectangle,font=\tt},
  edgeA/.style={thick, rounded corners,>=stealth'},
  edgeB/.style={thick, color=brown!50, rounded corners,>=stealth'},
  labelA/.style={anchor=center,fill=white,font=\scriptsize},
  labeltiny/.style={anchor=center,fill=white,font=\tiny},
  alertedge/.style={thick, color=red, rounded corners,>=stealth'}
}
\tikzset{arrow2/.style={decoration={markings,mark=at position 1 with %
    {\arrow[scale=2.3,>=latex]{>}}},postaction={decorate}}}  
    
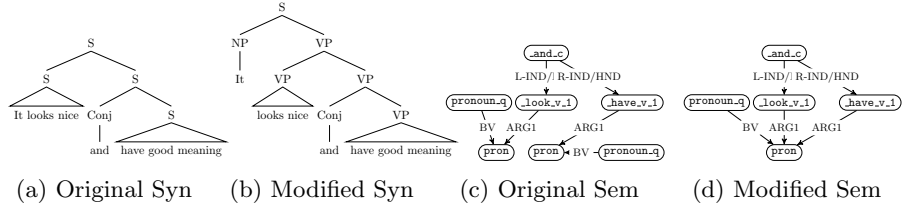
\begin{figure}[htbp]
\centering
\subcaptionbox{Original Syn}{
\begin{minipage}[t]{0.21\linewidth}
\centering
\scalebox{0.45}{
        \begin{tikzpicture}[level distance=30pt,sibling distance=0pt]
        \Tree 
        [.S [.S \edge[roof]; {It looks nice} ]
        [.S [.Conj and ]
        [.S \edge[roof]; {have good meaning} ] ] 
        ] 
        \end{tikzpicture}
        }
\end{minipage}%
}%
\subcaptionbox{Modified Syn}{
\begin{minipage}[t]{0.21\linewidth}
\centering
  \scalebox{0.45}{
        \begin{tikzpicture}[level distance=30pt,sibling distance=0pt]
        \Tree 
        [.S [.NP It ]
        [.VP [.VP \edge[roof]; {looks nice} ]
        [.VP [.Conj and ]
        [.VP \edge[roof]; {have good meaning} ] ] 
        ] ]
        \end{tikzpicture}}
\end{minipage}%
}%
\subcaptionbox{Original Sem}{
\begin{minipage}[t]{0.23\linewidth}
\centering
 \scalebox{0.45}{
    \begin{tikzpicture}
        \node (By) [ersnode, ] {\_look\_v\_1};
        \node (and) [ersnode, above = 1 of By ] {\_and\_c};
        \node (most) [ersnode, right = 0.7 of By ] {\_have\_v\_1};
        \node (quan) [ersnode, left = 0.1 of By] {pronoun\_q};
        \node (measure) [ersnode, below = 1 of By ] {pron};
        \node (good) [ersnode, right = 1 of measure] {pronoun\_q};
        \node (growing) [ersnode, below left = 1 and 0.5 of By ] {pron};

        \path[->] (most) edge [edgeA] node [labelA] {ARG1} (measure);
        \path[->] (By) edge [edgeA] node [labelA] {ARG1} (growing);
        \path[->] (and) edge [edgeA] node [labelA] {L-IND/HND} (By);
        \path[->] (and) edge [edgeA] node [labelA] {R-IND/HND} (most);
        \path[->] (quan) edge [edgeA] node [labelA] {BV} (growing);
        \path[->] (good) edge [edgeA] node [labelA] {BV} (measure);
       
      \end{tikzpicture}
            } 
\end{minipage}%
}%
\subcaptionbox{Modified Sem}{
\begin{minipage}[t]{0.23\linewidth}
\centering
   \scalebox{0.45}{
    \begin{tikzpicture}
        \node (By) [ersnode, ] {\_look\_v\_1};
        \node (and) [ersnode, above = 1 of By ] {\_and\_c};
        \node (most) [ersnode, right = 0.7 of By ] {\_have\_v\_1};
        \node (quan) [ersnode, left = 0.1 of By] {pronoun\_q};
        \node (growing) [ersnode, below = 1 of By ] {pron};

        \path[->] (most) edge [edgeA] node [labelA] {ARG1} (growing);
        \path[->] (By) edge [edgeA] node [labelA] {ARG1} (growing);
        \path[->] (and) edge [edgeA] node [labelA] {L-IND/HND} (By);
        \path[->] (and) edge [edgeA] node [labelA] {R-IND/HND} (most);
        \path[->] (quan) edge [edgeA] node [labelA] {BV} (growing);
      \end{tikzpicture}
            } 
\end{minipage}%
}%
\caption{Relevant original and modified syntactic and semantic analysis for Example \ref{ex8}.}
\label{fig:person}
\end{figure}

\paragraph{Binding}
Binding theory examines the constraints governing the use of anaphoric expressions, such as pronouns and reflexives, to determine how their meanings are derived from other elements in the context. According to the principles of binding theory \citep{chomsky1981gb1,Chomsky1986-CHOKOL}, a reflexive anaphora in English must be bound within its governing category or local domain.  However, this principle is less strictly adhered to in ESFL, as demonstrated in Example \ref{ex9}, where \textit{herself} is interpreted as referring to \textit{Pat}, rather than the principle-sanctioned antecedent \textit{my friends} (see Figure \ref{fig:binding}). We utilize the SHRG rules detailed in Table \ref{tab:binding} to resolve this discrepancy.

\es{\textit{At first, Pat denied all the things my friends told me about \textbf{herself} but she finally agreed with that.} \label{ex9}}
\begin{table}[htbp]
    \caption{Original and modified SHRG rules for Example \ref{ex9}.}
    \label{tab:binding}
    \begin{tabular}{ccccc}\toprule
          LHS & $\textrm{PP/N}_{\textrm{ori}}$ & $\textrm{NP}_{\textrm{ori}}$ & $\textrm{P}_{\textrm{mod}}$ & $\textrm{PP}_{\textrm{mod}}$ \\\midrule
          Syn & about & NP + NP & about & P + NP\\
          Sem & $\varnothing$ & \scalebox{0.5}{
\begin{tikzpicture}[>=latex',line join=bevel,]
\node (x1) at (54.737bp,138.8bp) [draw,circle,inner sep=1.5pt] {};
  \coordinate (x11) at (24.737bp,96.5bp);
  \node (x2) at (53.737bp,96.5bp) [draw=black,fill=black,circle,inner sep=1.5pt] {};
  \node (x3) at (97.737bp,96.5bp) [draw,circle,inner sep=1.5pt] {};
  \coordinate (x22) at (49.737bp,58.0bp);
  \coordinate (x33) at (99.737bp,58.0bp);
  \draw [->] (x1) ..controls (47.999bp,137.56bp) and (35.611bp,136.32bp)  .. (29.737bp,129.0bp) .. controls (26.807bp,125.35bp) and (25.036bp,120.9bp)  .. (x11);
  \definecolor{strokecol}{rgb}{0.0,0.0,0.0};
  \pgfsetstrokecolor{strokecol}
  \draw (41.737bp,120.75bp) node {appos};
  \draw [->] (x1) ..controls (54.613bp,131.57bp) and (54.237bp,112.64bp)  .. (x2);
  \draw (72.171bp,120.75bp) node {ARG1};
  \draw [->] (x1) ..controls (62.598bp,138.35bp) and (80.611bp,138.65bp)  .. (89.737bp,129.0bp) .. controls (96.972bp,121.35bp) and (98.609bp,109.24bp)  .. (x3);
  \draw (115.02bp,120.75bp) node {ARG2};
  \draw [->] (x2) ..controls (53.398bp,79.865bp) and (52.391bp,63.109bp)  .. (x22);
  \draw (63.088bp,82.25bp) node {NP};
  \draw [->] (x3) ..controls (97.907bp,79.865bp) and (98.411bp,63.109bp)  .. (x33);
  \draw (107.6bp,82.25bp) node {NP};
\end{tikzpicture}
} & \scalebox{0.6}{
\begin{tikzpicture}[>=latex,line join=bevel,]
\node (node1) at (1.8bp,33.8bp) [circle, inner sep=1.5pt,fill] {};
  \coordinate (node2) at (1.8bp,0.5bp);
  \draw [->] (node1) ..controls (1.8bp,27.979bp) and (1.8bp,18.64bp)  .. (node2);
  \definecolor{strokecol}{rgb}{0.0,0.0,0.0};
  \pgfsetstrokecolor{strokecol}
  \draw (25.3bp,16.5bp) node {\_about\_p};
\end{tikzpicture}
}  & \scalebox{0.6}{
\begin{tikzpicture}[>=latex,line join=bevel,]
\node (node1) at (12.5bp,68.4bp) [draw,circle, inner sep=1.5pt, fill] {};
  \coordinate (node2) at (0.5bp,33.8bp);
  \node (node3) at (25.5bp,33.8bp) [draw,circle,inner sep=1.5pt] {};
  \coordinate (node4) at (25.5bp,0.5bp);
  \draw [->] (node1) ..controls (9.4958bp,65.429bp) and (5.3665bp,62.358bp)  .. (3.5bp,58.6bp) .. controls (1.4852bp,54.544bp) and (0.64086bp,49.639bp)  .. (node2);
  \definecolor{strokecol}{rgb}{0.0,0.0,0.0};
  \pgfsetstrokecolor{strokecol}
  \draw (8.5bp,51.1bp) node {P};
  \draw [->] (node1) ..controls (13.985bp,63.677bp) and (18.077bp,53.416bp)  .. (node3);
  \draw (39.5bp,51.1bp) node {ARG2};
  \draw [->] (node3) ..controls (25.5bp,27.979bp) and (25.5bp,18.64bp)  .. (node4);
  \draw (32.5bp,16.5bp) node {NP};
\end{tikzpicture}
} \\\bottomrule
    \end{tabular}
\end{table}
\tikzset{ersnode/.style={right, rounded rectangle, draw, font=\tt}} 
\tikzset{ 
  nodeA/.style={right, fill=blue!40, text=white, rounded rectangle,font=\tt},
  nodeB/.style={right, fill=brown!50, text=white, rounded rectangle,font=\tt},
  edgeA/.style={thick, rounded corners,>=stealth'},
  edgeB/.style={thick, color=brown!50, rounded corners,>=stealth'},
  labelA/.style={anchor=center,fill=white,font=\scriptsize},
  labeltiny/.style={anchor=center,fill=white,font=\tiny},
  alertedge/.style={thick, color=red, rounded corners,>=stealth'}
}
\tikzset{arrow2/.style={decoration={markings,mark=at position 1 with %
    {\arrow[scale=2.3,>=latex]{>}}},postaction={decorate}}}  
    
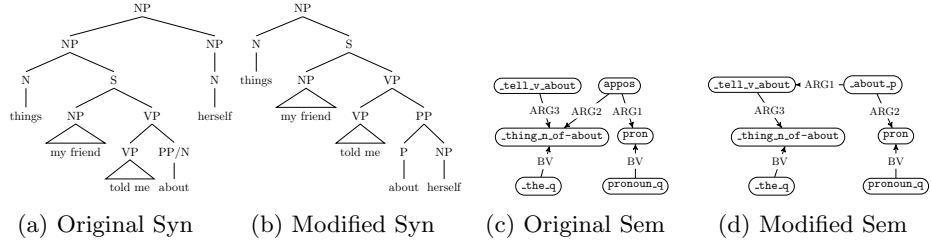
\begin{figure}[htbp]
\centering
\subcaptionbox{Original Syn}{
\begin{minipage}[t]{0.22\textwidth}
\centering
\scalebox{0.45}{
        \begin{tikzpicture}[level distance=30pt,sibling distance=0pt]
        \Tree 
        [.NP 
        [.NP [.N things ] [.S [.NP \edge[roof]; {my friend} ] [.VP [.VP \edge[roof]; {told me} ] [.PP/N about ] ] ] ]
        [.NP [.N herself ] ]
        ]
        \end{tikzpicture}
        }
\end{minipage}%
}%
\subcaptionbox{Modified Syn}{
\begin{minipage}[t]{0.24\textwidth}
\centering
  \scalebox{0.45}{
        \begin{tikzpicture}[level distance=30pt,sibling distance=0pt]
        \Tree 
        [.NP [.N things ] [.S [.NP \edge[roof]; {my friend} ] [.VP [.VP \edge[roof]; {told me} ] [.PP [.P about ] [.NP herself ] ] ] ] ]
        \end{tikzpicture}}
\end{minipage}%
}%
\subcaptionbox{Original Sem}{
\begin{minipage}[t]{0.21\textwidth}
\centering
 \scalebox{0.45}{
    \begin{tikzpicture}
        \node (By) [ersnode, ] {appos};
        \node (alone) [ersnode, left = 0.5 of By ] {\_tell\_v\_about};
        \node (growing) [ersnode, below left = 1 and 0.2 of By ] {\_thing\_n\_of-about};
        \node (measure) [ersnode, below right = 1 and -0.2 of By ] {pron};
        \node (most) [ersnode, below = 1 of measure ] {pronoun\_q};
        \node (thingq) [ersnode, left = 1.2 of most ] {\_the\_q};
       
        \path[->] (By) edge [edgeA] node [labelA] {ARG1} (measure);
        \path[->] (By) edge [edgeA] node [labelA] {ARG2} (growing);
        \path[->] (most) edge [edgeA] node [labelA] {BV} (measure);
        \path[->] (alone) edge [edgeA] node [labelA] {ARG3} (growing);
        \path[->] (thingq) edge [edgeA] node [labelA] {BV} (growing);
      \end{tikzpicture}
            } 
\end{minipage}%
}%
\subcaptionbox{Modified Sem}{
\begin{minipage}[t]{0.25\textwidth}
\centering
   \scalebox{0.45}{
    \begin{tikzpicture}
         \node (By) [ersnode, ] {\_about\_p};
        \node (alone) [ersnode, left = 1.4 of By ] {\_tell\_v\_about};
        \node (growing) [ersnode, below left = 1 and 0.5 of By ] {\_thing\_n\_of-about};
        \node (measure) [ersnode, below right = 1 and -0.2 of By ] {pron};
        \node (most) [ersnode, below = 1 of measure ] {pronoun\_q};
        \node (thingq) [ersnode, left = 2 of most ] {\_the\_q};
       
        \path[->] (By) edge [edgeA] node [labelA] {ARG2} (measure);
        \path[->] (By) edge [edgeA] node [labelA] {ARG1} (alone);
        \path[->] (most) edge [edgeA] node [labelA] {BV} (measure);
        \path[->] (alone) edge [edgeA] node [labelA] {ARG3} (growing);
        \path[->] (thingq) edge [edgeA] node [labelA] {BV} (growing);
      \end{tikzpicture}
            } 
\end{minipage}%
}%
\caption{Relevant original and modified syntactic and semantic analysis for Example \ref{ex9}.}
\label{fig:binding}
\end{figure}

\paragraph{Ellipsis}
Ellipsis refers to the omission of a previously mentioned string in subsequent structures. In ESFL, ellipsis demonstrates notable flexibility. As shown in Example \ref{ex10}, constituents that are typically not elided in standard English may be omitted. This variability can result in inaccuracies in the ACE parser's analysis, as illustrated in Figure \ref{fig:ellipsis}. To mitigate this issue, we apply the SHRG rules presented in Table \ref{tab:ellipsis}.

\es{\textit{His voice is also soulful and sometimes even gave us \textbf{blissful}.} \label{ex10}}
\begin{table}[htbp]
    \caption{Original and modified SHRG rules for Example \ref{ex10}.}
    \begin{tabular}{ccccccc}\toprule
          LHS & $\textrm{VP}_{\textrm{ori}}$ & $\textrm{PP}_{\textrm{ori}}$ & $\textrm{VP}_{\textrm{ori}}$ & $\textrm{VP}_{\textrm{mod}}$ & $\textrm{NP}_{\textrm{mod}}$ & $\textrm{VP}_{\textrm{mod}}$ \\\midrule
          Syn & V+NP & ADJ & VP+PP & V+NP & ADJ & VP+NP \\
          Sem & \scalebox{0.56}{
\begin{tikzpicture}[>=latex,line join=bevel,]
\node (node1) at (12.5bp,68.4bp) [draw,circle, inner sep=1.5pt, fill] {};
  \coordinate (node2) at (0.5bp,33.8bp);
  \node (node3) at (25.5bp,33.8bp) [draw,circle,inner sep=1.5pt] {};
  \coordinate (node4) at (25.5bp,0.5bp);
  \draw [->] (node1) ..controls (9.4958bp,65.429bp) and (5.3665bp,62.358bp)  .. (3.5bp,58.6bp) .. controls (1.4852bp,54.544bp) and (0.64086bp,49.639bp)  .. (node2);
  \definecolor{strokecol}{rgb}{0.0,0.0,0.0};
  \pgfsetstrokecolor{strokecol}
  \draw (8.5bp,51.1bp) node {V};
  \draw [->] (node1) ..controls (13.985bp,63.677bp) and (18.077bp,53.416bp)  .. (node3);
  \draw (39.5bp,51.1bp) node {ARG2};
  \draw [->] (node3) ..controls (25.5bp,27.979bp) and (25.5bp,18.64bp)  .. (node4);
  \draw (35.5bp,16.5bp) node {NP};
\end{tikzpicture}
}  & \scalebox{0.6}{
\begin{tikzpicture}[>=latex,line join=bevel,]
\node (node1) at (1.8bp,33.8bp) [circle, inner sep=1.5pt,fill] {};
  \coordinate (node2) at (1.8bp,0.5bp);
  \draw [->] (node1) ..controls (1.8bp,27.979bp) and (1.8bp,18.64bp)  .. (node2);
  \definecolor{strokecol}{rgb}{0.0,0.0,0.0};
  \pgfsetstrokecolor{strokecol}
  \draw (15.3bp,16.5bp) node {ADJ};
\end{tikzpicture}
}  & \scalebox{0.56}{
\begin{tikzpicture}[>=latex',line join=bevel,]
\node (x1) at (54.737bp,138.8bp) [draw,circle,inner sep=1.5pt] {};
  \coordinate (x11) at (18.737bp,96.5bp);
  \node (x2) at (53.737bp,96.5bp) [draw=black,fill=black,circle,inner sep=1.5pt] {};
  \node (x3) at (97.737bp,96.5bp) [draw,circle,inner sep=1.5pt] {};
  \coordinate (x22) at (49.737bp,58.0bp);
  \coordinate (x33) at (99.737bp,58.0bp);
  \draw [->] (x1) ..controls (37.999bp,137.56bp) and (32.611bp,136.32bp)  .. (26.737bp,129.0bp) .. controls (24.807bp,125.35bp) and (19.036bp,120.9bp)  .. (x11);
  \definecolor{strokecol}{rgb}{0.0,0.0,0.0};
  \pgfsetstrokecolor{strokecol}
  \draw (37.737bp,120.75bp) node {subord};
  \draw [->] (x1) ..controls (54.613bp,131.57bp) and (54.237bp,112.64bp)  .. (x2);
  \draw (72.171bp,120.75bp) node {ARG1};
  \draw [->] (x1) ..controls (62.598bp,138.35bp) and (80.611bp,138.65bp)  .. (89.737bp,129.0bp) .. controls (96.972bp,121.35bp) and (98.609bp,109.24bp)  .. (x3);
  \draw (115.02bp,120.75bp) node {ARG2};
  \draw [->] (x2) ..controls (53.398bp,79.865bp) and (52.391bp,63.109bp)  .. (x22);
  \draw (63.088bp,82.25bp) node {VP};
  \draw [->] (x3) ..controls (97.907bp,79.865bp) and (98.411bp,63.109bp)  .. (x33);
  \draw (107.6bp,82.25bp) node {PP};
\end{tikzpicture}
}  & \scalebox{0.56}{
\begin{tikzpicture}[>=latex,line join=bevel,]
\node (node1) at (12.5bp,68.4bp) [draw,circle, inner sep=1.5pt, fill] {};
  \coordinate (node2) at (0.5bp,33.8bp);
  \node (node3) at (25.5bp,33.8bp) [draw,circle,inner sep=1.5pt] {};
  \coordinate (node4) at (25.5bp,0.5bp);
  \draw [->] (node1) ..controls (9.4958bp,65.429bp) and (5.3665bp,62.358bp)  .. (3.5bp,58.6bp) .. controls (1.4852bp,54.544bp) and (0.64086bp,49.639bp)  .. (node2);
  \definecolor{strokecol}{rgb}{0.0,0.0,0.0};
  \pgfsetstrokecolor{strokecol}
  \draw (8.5bp,51.1bp) node {V};
  \draw [->] (node1) ..controls (13.985bp,63.677bp) and (18.077bp,53.416bp)  .. (node3);
  \draw (39.5bp,51.1bp) node {ARG3};
  \draw [->] (node3) ..controls (25.5bp,27.979bp) and (25.5bp,18.64bp)  .. (node4);
  \draw (35.5bp,16.5bp) node {NP};
\end{tikzpicture}
} & \scalebox{0.56}{
\begin{tikzpicture}[>=latex',line join=bevel,]
\node (x1) at (46.737bp,138.8bp) [draw,circle,inner sep=1.5pt] {};
  \coordinate (x11) at (24.737bp,96.5bp);
  \node (x2) at (80.737bp,96.5bp) [draw,fill,inner sep=1.5pt,circle] {};
  \coordinate (x22) at (80.737bp,58.0bp);
  \node (x3) at (113.74bp,138.8bp) [draw,inner sep=1.5pt,circle] {};
  \coordinate (x33) at (134.74bp,96.5bp);
  \draw [->] (x1) ..controls (41.927bp,136.71bp) and (33.937bp,134.34bp)  .. (30.237bp,129.0bp) .. controls (27.61bp,125.21bp) and (25.947bp,120.75bp)  .. (x11);
  \definecolor{strokecol}{rgb}{0.0,0.0,0.0};
  \pgfsetstrokecolor{strokecol}
  \draw (48.987bp,120.75bp) node {udef\_q};
  \draw [->] (x1) ..controls (51.427bp,136.54bp) and (59.317bp,133.96bp)  .. (63.737bp,129.0bp) .. controls (68.953bp,123.14bp) and (67.049bp,119.77bp)  .. (69.987bp,112.5bp) .. controls (71.84bp,107.92bp) and (73.973bp,102.9bp)  .. (x2);
  \draw (79.362bp,120.75bp) node {BV};
  \draw [->] (x2) ..controls (80.737bp,79.865bp) and (80.737bp,63.109bp)  .. (x22);
  \draw (119.36bp,74.25bp) node {generic\_entity};
  \draw [->] (x3) ..controls (108.08bp,136.87bp) and (98.597bp,134.72bp)  .. (93.487bp,129.0bp) .. controls (92.463bp,127.85bp) and (87.567bp,111.32bp)  .. (x2);
  \draw (111.11bp,120.75bp) node {ARG1};
  \draw [->] (x3) ..controls (118.68bp,136.36bp) and (125.5bp,133.81bp)  .. (128.74bp,129.0bp) .. controls (131.31bp,125.17bp) and (132.98bp,120.69bp)  .. (x33);
  \draw (146.99bp,120.75bp) node {ADJ};
\end{tikzpicture}
} & 
\scalebox{0.56}{
\begin{tikzpicture}[>=latex,line join=bevel,]
\node (node1) at (12.5bp,68.4bp) [draw,circle, inner sep=1.5pt, fill] {};
  \coordinate (node2) at (0.5bp,33.8bp);
  \node (node3) at (25.5bp,33.8bp) [draw,circle,inner sep=1.5pt] {};
  \coordinate (node4) at (25.5bp,0.5bp);
  \draw [->] (node1) ..controls (9.4958bp,65.429bp) and (5.3665bp,62.358bp)  .. (3.5bp,58.6bp) .. controls (1.4852bp,54.544bp) and (0.64086bp,49.639bp)  .. (node2);
  \definecolor{strokecol}{rgb}{0.0,0.0,0.0};
  \pgfsetstrokecolor{strokecol}
  \draw (8.5bp,51.1bp) node {VP};
  \draw [->] (node1) ..controls (13.985bp,63.677bp) and (18.077bp,53.416bp)  .. (node3);
  \draw (39.5bp,51.1bp) node {ARG2};
  \draw [->] (node3) ..controls (25.5bp,27.979bp) and (25.5bp,18.64bp)  .. (node4);
  \draw (35.5bp,16.5bp) node {NP};
\end{tikzpicture}
}\\\bottomrule
    \end{tabular}
    
    \label{tab:ellipsis}
\end{table}
\tikzset{ersnode/.style={right, rounded rectangle, draw, font=\tt}} 
\tikzset{ 
  nodeA/.style={right, fill=blue!40, text=white, rounded rectangle,font=\tt},
  nodeB/.style={right, fill=brown!50, text=white, rounded rectangle,font=\tt},
  edgeA/.style={thick, rounded corners,>=stealth'},
  edgeB/.style={thick, color=brown!50, rounded corners,>=stealth'},
  labelA/.style={anchor=center,fill=white,font=\scriptsize},
  labeltiny/.style={anchor=center,fill=white,font=\tiny},
  alertedge/.style={thick, color=red, rounded corners,>=stealth'}
}
\tikzset{arrow2/.style={decoration={markings,mark=at position 1 with %
    {\arrow[scale=2.3,>=latex]{>}}},postaction={decorate}}}  
    
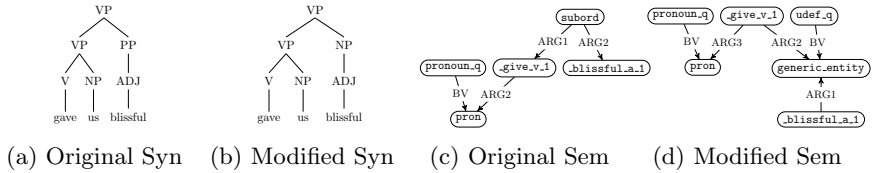
\begin{figure}[htbp]
\centering
\subcaptionbox{Original Syn}{
\begin{minipage}[t]{0.2\textwidth}
\centering
\scalebox{0.45}{
        \begin{tikzpicture}[level distance=30pt,sibling distance=0pt]
        \Tree 
        [.VP 
            [.VP [.V gave ] [.NP us ] ]
            [.PP [.ADJ blissful ] ] 
        ]  
        \end{tikzpicture}
        }
\end{minipage}%
}%
\subcaptionbox{Modified Syn}{
\begin{minipage}[t]{0.2\linewidth}
\centering
  \scalebox{0.45}{
        \begin{tikzpicture}[level distance=30pt,sibling distance=5pt]
        \Tree 
        [.VP 
            [.VP [.V gave ] [.NP us ] ]
            [.NP [.ADJ blissful ] ] 
        ]   
        \end{tikzpicture}}
\end{minipage}%
}%
\subcaptionbox{Original Sem}{
\begin{minipage}[t]{0.22\linewidth}
\centering
 \scalebox{0.45}{
    \begin{tikzpicture}
        \node (By) [ersnode, ] {\_give\_v\_1};
        \node (most) [ersnode, above right = 1 and 0.5 of By ] {subord};
        \node (alone) [ersnode, left = 0.2 of By ] {pronoun\_q};
        
        \node (growing) [ersnode, below left = 1 and 0.7 of By ] {pron};
        \node (measure) [ersnode, right = 0.2 of By ] {\_blissful\_a\_1};
       
        \path[->] (most) edge [edgeA] node [labelA] {ARG1} (By);
        \path[->] (By) edge [edgeA] node [labelA] {ARG2} (growing);
        \path[->] (most) edge [edgeA] node [labelA] {ARG2} (measure);
        \path[->] (alone) edge [edgeA] node [labelA] {BV} (growing);
       
      \end{tikzpicture}
            } 
\end{minipage}%
}%
\subcaptionbox{Modified Sem}{
\begin{minipage}[t]{0.22\textwidth}
\centering
   \scalebox{0.45}{
    \begin{tikzpicture}
         \node (By) [ersnode, ] {\_give\_v\_1};
        \node (most) [ersnode, right = 0.2 of By ] {udef\_q};
        \node (alone) [ersnode, left = 0.2 of By ] {pronoun\_q};
        \node (measure) [ersnode, below right = 1 and 0.2 of By ] {generic\_entity};
        \node (blissful) [ersnode, below = 1 of measure ] {\_blissful\_a\_1};
        \node (growing) [ersnode, below left = 1 and 0.5 of By ] {pron};
       
        \path[->] (By) edge [edgeA] node [labelA] {ARG2} (measure);
        \path[->] (By) edge [edgeA] node [labelA] {ARG3} (growing);
        \path[->] (most) edge [edgeA] node [labelA] {BV} (measure);
        \path[->] (alone) edge [edgeA] node [labelA] {BV} (growing);
        \path[->] (blissful) edge [edgeA] node [labelA] {ARG1} (measure);

      \end{tikzpicture}
            } 
\end{minipage}%
}%
\caption{Relevant original and modified syntactic and semantic analysis for Example \ref{ex10}.}
\label{fig:ellipsis}
\end{figure}

\paragraph{Filler-Gap}
Filler-Gap constructions are considered as the result of phrasal movement in generative grammar --- the moved phrase serves as a filler for its canonical position, which is left unexpressed as a gap. 
Establishing a connection between the filler and its original in situ position is essential to ensure the utterance is interpretable.
In ESFL, however, the presence and forms of fillers and gaps can differ significantly from those in English (see Example \ref{ex11}). As shown in Figure \ref{fig:filler}, these variations pose challenges for the ACE parser. We employ modified SHRG rules in Table \ref{tab:filler} to address it.
\es{\textit{You can find any information from it \textbf{what} you want.} \label{ex11}}
\begin{table}[htbp]
    \caption{Original and modified SHRG rules for Example \ref{ex11}.}
    \label{tab:filler}
    \begin{tabular}{ccccc}\toprule
          LHS & $\textrm{N}_{\textrm{ori}}$ & $\textrm{NP}_{\textrm{ori}}$ & $\textrm{N}_{\textrm{mod}}$ & $\textrm{NP}_{\textrm{mod}}$ \\\midrule
          Syn & what & N + S & what & N + S \\
          Sem & \scalebox{0.6}{
\begin{tikzpicture}[>=latex,line join=bevel,]
\node (node1) at (22.5bp,68.4bp) [draw,circle, inner sep=1.5pt] {};
  \coordinate (node2) at (0.5bp,33.8bp);
  \node (node3) at (44.5bp,33.8bp) [draw,circle,inner sep=1.5pt,fill] {};
  \coordinate (node4) at (44.5bp,0.5bp);
  \draw [->] (node1) ..controls (17.636bp,66.286bp) and (9.5231bp,63.891bp)  .. (5.5bp,58.6bp) .. controls (2.625bp,54.819bp) and (1.3113bp,49.765bp)  .. (node2);
  \definecolor{strokecol}{rgb}{0.0,0.0,0.0};
  \pgfsetstrokecolor{strokecol}
  \draw (22.5bp,51.1bp) node {which\_q};
  \draw [->] (node1) ..controls (27.364bp,66.286bp) and (35.477bp,63.891bp)  .. (39.5bp,58.6bp) .. controls (42.26bp,54.97bp) and (43.581bp,50.167bp)  .. (node3);
  \draw (53.0bp,51.1bp) node {BV};
  \draw [->] (node3) ..controls (44.5bp,27.979bp) and (44.5bp,18.64bp)  .. (node4);
  \draw (58.0bp,16.5bp) node {thing};
\end{tikzpicture}
}   & \scalebox{0.6}{
\begin{tikzpicture}[>=latex,line join=bevel,]
\node (node1) at (12.5bp,68.4bp) [draw,circle, inner sep=1.5pt, fill] {};
  \coordinate (node2) at (0.5bp,33.8bp);
  \node (node3) at (25.5bp,33.8bp) [draw,circle,inner sep=1.5pt] {};
  \coordinate (node4) at (25.5bp,0.5bp);
  \draw [->] (node1) ..controls (9.4958bp,65.429bp) and (5.3665bp,62.358bp)  .. (3.5bp,58.6bp) .. controls (1.4852bp,54.544bp) and (0.64086bp,49.639bp)  .. (node2);
  \definecolor{strokecol}{rgb}{0.0,0.0,0.0};
  \pgfsetstrokecolor{strokecol}
  \draw (8.5bp,51.1bp) node {N};
  \draw [->] (node1) ..controls (13.985bp,63.677bp) and (18.077bp,53.416bp)  .. (node3);
  \draw (39.5bp,51.1bp) node {ARG1};
  \draw [->] (node3) ..controls (25.5bp,27.979bp) and (25.5bp,18.64bp)  .. (node4);
  \draw (32.5bp,16.5bp) node {S};
\end{tikzpicture}
}  & $\varnothing$ & \scalebox{0.6}{
\begin{tikzpicture}[>=latex,line join=bevel,]
\node (node1) at (12.5bp,68.4bp) [draw,circle, inner sep=1.5pt] {};
  \coordinate (node2) at (0.5bp,33.8bp);
  \node (node3) at (25.5bp,33.8bp) [draw,circle,inner sep=1.5pt,fill] {};
  \coordinate (node4) at (25.5bp,0.5bp);
  \draw [->] (node1) ..controls (9.4958bp,65.429bp) and (5.3665bp,62.358bp)  .. (3.5bp,58.6bp) .. controls (1.4852bp,54.544bp) and (0.64086bp,49.639bp)  .. (node2);
  \definecolor{strokecol}{rgb}{0.0,0.0,0.0};
  \pgfsetstrokecolor{strokecol}
  \draw (8.5bp,51.1bp) node {S};
  \draw [->] (node1) ..controls (13.985bp,63.677bp) and (18.077bp,53.416bp)  .. (node3);
  \draw (39.5bp,51.1bp) node {ARG2};
  \draw [->] (node3) ..controls (25.5bp,27.979bp) and (25.5bp,18.64bp)  .. (node4);
  \draw (32.5bp,16.5bp) node {N};
\end{tikzpicture}
} \\\bottomrule
    \end{tabular}  
\end{table}
\tikzset{ersnode/.style={right, rounded rectangle, draw, font=\tt}} 
\tikzset{ 
  nodeA/.style={right, fill=blue!40, text=white, rounded rectangle,font=\tt},
  nodeB/.style={right, fill=brown!50, text=white, rounded rectangle,font=\tt},
  edgeA/.style={thick, rounded corners,>=stealth'},
  edgeB/.style={thick, color=brown!50, rounded corners,>=stealth'},
  labelA/.style={anchor=center,fill=white,font=\scriptsize},
  labeltiny/.style={anchor=center,fill=white,font=\tiny},
  alertedge/.style={thick, color=red, rounded corners,>=stealth'}
}
\tikzset{arrow2/.style={decoration={markings,mark=at position 1 with %
    {\arrow[scale=2.3,>=latex]{>}}},postaction={decorate}}}  
    
\begin{figure}[htbp]
\centering
\subcaptionbox{Original Syn}{
\begin{minipage}[t]{0.22\textwidth}
\centering
\scalebox{0.45}{
        \begin{tikzpicture}[level distance=30pt,sibling distance=0pt]
        \Tree 
        [.NP 
            [.N [.N information ] [.PP \edge[roof]; {from it} ] ] 
            [.S [.N what ]
            [.S/N \edge[roof]; {you want} ] ] 
        ]  
        \end{tikzpicture}
        }
\end{minipage}%
}%
\subcaptionbox{Modified Syn}{
\begin{minipage}[t]{0.22\textwidth}
\centering
  \scalebox{0.45}{
        \begin{tikzpicture}[level distance=30pt,sibling distance=5pt]
        \Tree 
        [.NP 
            [.N [.N information ] [.PP \edge[roof]; {from it} ] ] 
            [.S [.N what ]
            [.S/N \edge[roof]; {you want} ] ] 
        ]  
        \end{tikzpicture}}
\end{minipage}%
}%
\subcaptionbox{Original Sem}{
\begin{minipage}[t]{0.24\linewidth}
\centering
 \scalebox{0.45}{
    \begin{tikzpicture}
        \node (By) [ersnode, ] {\_want\_v\_1};
        \node (information) [ersnode, above = 1 of By ] {\_information\_n\_on-about};
        \node (most) [ersnode, right = 0.5 of By ] {which\_q};
        \node (alone) [ersnode, left = 0.5 of By ] {pron\_q};
        \node (measure) [ersnode, below right = 1 and -0.2 of By ] {thing};
        \node (growing) [ersnode, below left = 1 and -0.2 of By ] {pron};
       
        \path[->] (information) edge [edgeA] node [labelA] {ARG1} (By);
        \path[->] (By) edge [edgeA] node [labelA] {ARG2} (measure);
        \path[->] (By) edge [edgeA] node [labelA] {ARG1} (growing);
        \path[->] (most) edge [edgeA] node [labelA] {BV} (measure);
        \path[->] (alone) edge [edgeA] node [labelA] {BV} (growing);
       
      \end{tikzpicture}
            } 
\end{minipage}%
}%
\subcaptionbox{Modified Sem}{
\begin{minipage}[t]{0.24\linewidth}
\centering
   \scalebox{0.45}{
    \begin{tikzpicture}
        \node (By) [ersnode, ] {\_want\_v\_1};
        \node (most) [ersnode, right = 0.5 of By ] {\_any\_q};
        \node (alone) [ersnode, left = 0.5 of By ] {pron\_q};
        \node (measure) [ersnode, below right = 1 and -1.5 of By ] {\_information\_n\_on-about};
        \node (growing) [ersnode, below left = 1 and 1 of By ] {pron};
        \path[->] (By) edge [edgeA] node [labelA] {ARG2} (measure);
        \path[->] (By) edge [edgeA] node [labelA] {ARG1} (growing);
        \path[->] (most) edge [edgeA] node [labelA] {BV} (measure);
        \path[->] (alone) edge [edgeA] node [labelA] {BV} (growing);

      \end{tikzpicture}
            } 
\end{minipage}%
}%
\caption{Relevant original and modified syntactic and semantic analysis for Example \ref{ex11}.}
\label{fig:filler}
\end{figure}
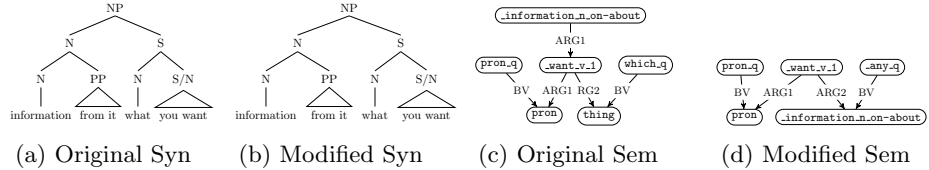

\paragraph{Argument Structure}
Argument structure refers to the syntactic configuration projected by a lexical item, particularly by a verb, within a sentence. It defines the specific requirements regarding both the number and realization of arguments. However, in ESFL, the argument structure of the same verb can differ significantly from its English counterpart. These variations are evident not only in the number of arguments (see Example \ref{ex12}), but also in how they are realized (see Example \ref{ex13}).
Figure \ref{fig:arguement} and Figure \ref{fig:arguement2}, along with Table \ref{tab:argument} and Table \ref{tab:argument2},
illustrate the challenges posed by these diverse argument structures for the ACE parser, as well as how we address them.

\es{\textit{I hope you can \textbf{enjoy}.}\label{ex12}}

\begin{table}[htbp]
\caption{Original and modified SHRG rules for Example \ref{ex12}.}
\label{tab:argument}
\begin{tabular}{lllll}
\toprule
LHS & $\textrm{N}_{\textrm{ori}}$ & $\textrm{NP}_{\textrm{ori}}$ & $\textrm{V/N}_{\textrm{ori}}$ & $\textrm{S}_{\textrm{ori}}$ \\ \midrule
Syn & hope & N + N & enjoy & NP + S/N \\ \addlinespace
Sem & 
\scalebox{0.6}{
\begin{tikzpicture}[>=latex,line join=bevel]
  \node (node1) at (1.8bp,33.8bp) [circle, inner sep=1.5pt,fill] {};
  \coordinate (node2) at (1.8bp,0.5bp);
  \draw [->] (node1) ..controls (1.8bp,27.979bp) and (1.8bp,18.64bp) .. (node2);
  \draw (26.3bp,16.5bp) node {\_hope\_n\_1};
\end{tikzpicture}
} & 
\scalebox{0.5}{
\begin{tikzpicture}[>=latex',line join=bevel]
  \node (x1) at (88.737bp,138.8bp) [draw,circle,inner sep=1.5pt] {};
  \coordinate (x11) at (24.737bp,96.5bp);
  \node (x2) at (84.737bp,96.5bp) [draw=black,fill=black,circle,inner sep=1.5pt] {};
  \node (x3) at (131.74bp,96.5bp) [draw,circle,inner sep=1.5pt] {};
  \coordinate (x22) at (82.737bp,58.0bp);
  \coordinate (x33) at (132.74bp,58.0bp);
  \draw [->] (x1) ..controls (78.189bp,137.87bp) and (37.251bp,137.64bp) .. (28.987bp,129.0bp) .. controls (25.608bp,125.47bp) and (23.715bp,120.91bp) .. (x11);
  \draw (57.862bp,120.75bp) node {compound};
  \draw [->] (x1) ..controls (88.242bp,131.57bp) and (86.737bp,112.64bp) .. (x2);
  \draw (105.6bp,120.75bp) node {ARG1};
  \draw [->] (x1) ..controls (96.342bp,138.24bp) and (113.78bp,138.32bp) .. (122.74bp,129.0bp) .. controls (130.07bp,121.37bp) and (132.01bp,109.26bp) .. (x3);
  \draw (148.34bp,120.75bp) node {ARG2};
  \draw [->] (x2) ..controls (84.568bp,79.865bp) and (84.064bp,63.109bp) .. (x22);
  \draw (88.85bp,82.25bp) node {N};
  \draw [->] (x3) ..controls (131.82bp,79.865bp) and (132.07bp,63.109bp) .. (x33);
  \draw (137.23bp,82.25bp) node {N};
\end{tikzpicture}
} & 
\scalebox{0.6}{
\begin{tikzpicture}[>=latex,line join=bevel]
  \node (node1) at (1.8bp,33.8bp) [circle, inner sep=1.5pt,fill] {};
  \coordinate (node2) at (1.8bp,0.5bp);
  \draw [->] (node1) ..controls (1.8bp,27.979bp) and (1.8bp,18.64bp) .. (node2);
  \draw (30.3bp,16.5bp) node {\_enjoy\_v\_1};
\end{tikzpicture}
} & 
\scalebox{0.6}{
\begin{tikzpicture}[>=latex,line join=bevel]
  \node (node1) at (12.5bp,68.4bp) [draw,circle, inner sep=1.5pt] {};
  \coordinate (node2) at (0.5bp,33.8bp);
  \node (node3) at (25.5bp,33.8bp) [draw,circle,inner sep=1.5pt,fill] {};
  \coordinate (node4) at (25.5bp,0.5bp);
  \draw [->] (node1) ..controls (9.4958bp,65.429bp) and (5.3665bp,62.358bp) .. (3.5bp,58.6bp) .. controls (1.4852bp,54.544bp) and (0.64086bp,49.639bp) .. (node2);
  \draw (8.5bp,51.1bp) node {S/N};
  \draw [->] (node1) ..controls (13.985bp,63.677bp) and (18.077bp,53.416bp) .. (node3);
  \draw (39.5bp,51.1bp) node {ARG2};
  \draw [->] (node3) ..controls (25.5bp,27.979bp) and (25.5bp,18.64bp) .. (node4);
  \draw (32.5bp,16.5bp) node {NP};
\end{tikzpicture}
} \\ \midrule
LHS & $\textrm{V}_{\textrm{mod}}$ & $\textrm{N}_{\textrm{mod}}$ & $\textrm{V/N}_{\textrm{mod}}$ & $\textrm{VP}_{\textrm{mod}}$ \\ \midrule
Syn & hope & I & enjoy & V + S \\ \addlinespace
Sem & 
\scalebox{0.6}{
\begin{tikzpicture}[>=latex,line join=bevel]
  \node (node1) at (1.8bp,33.8bp) [circle, inner sep=1.5pt,fill] {};
  \coordinate (node2) at (1.8bp,0.5bp);
  \draw [->] (node1) ..controls (1.8bp,27.979bp) and (1.8bp,18.64bp) .. (node2);
  \draw (26.3bp,16.5bp) node {\_hope\_v\_1};
\end{tikzpicture}
} & 
\scalebox{0.56}{
\begin{tikzpicture}[>=latex,line join=bevel]
  \node (node1) at (22.5bp,68.4bp) [draw,circle, inner sep=1.5pt] {};
  \coordinate (node2) at (-15.5bp,33.8bp);
  \node (node3) at (44.5bp,33.8bp) [draw,circle,inner sep=1.5pt,fill] {};
  \coordinate (node4) at (44.5bp,0.5bp);
  \draw [->] (node1) ..controls (17.636bp,66.286bp) and (9.5231bp,63.891bp) .. (4.5bp,62.6bp) .. controls (-12.625bp,57.819bp) and (-9.3113bp,49.765bp) .. (node2);
  \draw (16.5bp,51.1bp) node {pronoun\_q};
  \draw [->] (node1) ..controls (27.364bp,66.286bp) and (35.477bp,63.891bp) .. (39.5bp,58.6bp) .. controls (42.26bp,54.97bp) and (43.581bp,50.167bp) .. (node3);
  \draw (53.0bp,51.1bp) node {BV};
  \draw [->] (node3) ..controls (44.5bp,27.979bp) and (44.5bp,18.64bp) .. (node4);
  \draw (55.0bp,16.5bp) node {pron};
\end{tikzpicture}
} & 
\scalebox{0.6}{
\begin{tikzpicture}[>=latex,line join=bevel]
  \node (node1) at (1.8bp,33.8bp) [circle, inner sep=1.5pt,fill] {};
  \coordinate (node2) at (1.8bp,0.5bp);
  \draw [->] (node1) ..controls (1.8bp,27.979bp) and (1.8bp,18.64bp) .. (node2);
  \draw (28.3bp,16.5bp) node {ellipsis\_ref};
\end{tikzpicture}
} & 
\scalebox{0.6}{
\begin{tikzpicture}[>=latex,line join=bevel]
  \node (node1) at (12.5bp,68.4bp) [draw,circle, inner sep=1.5pt] {};
  \coordinate (node2) at (0.5bp,33.8bp);
  \node (node3) at (25.5bp,33.8bp) [draw,circle,inner sep=1.5pt,fill] {};
  \coordinate (node4) at (25.5bp,0.5bp);
  \draw [->] (node1) ..controls (9.4958bp,65.429bp) and (5.3665bp,62.358bp) .. (3.5bp,58.6bp) .. controls (1.4852bp,54.544bp) and (0.64086bp,49.639bp) .. (node2);
  \draw (8.5bp,51.1bp) node {V};
  \draw [->] (node1) ..controls (13.985bp,63.677bp) and (18.077bp,53.416bp) .. (node3);
  \draw (39.5bp,51.1bp) node {ARG2};
  \draw [->] (node3) ..controls (25.5bp,27.979bp) and (25.5bp,18.64bp) .. (node4);
  \draw (32.5bp,16.5bp) node {S};
\end{tikzpicture}
} \\ \bottomrule
\end{tabular}
\end{table}
\tikzset{ersnode/.style={right, rounded rectangle, draw, font=\tt}} 
\tikzset{ 
  nodeA/.style={right, fill=blue!40, text=white, rounded rectangle,font=\tt},
  nodeB/.style={right, fill=brown!50, text=white, rounded rectangle,font=\tt},
  edgeA/.style={thick, rounded corners,>=stealth'},
  edgeB/.style={thick, color=brown!50, rounded corners,>=stealth'},
  labelA/.style={anchor=center,fill=white,font=\scriptsize},
  labeltiny/.style={anchor=center,fill=white,font=\tiny},
  alertedge/.style={thick, color=red, rounded corners,>=stealth'}
}
\tikzset{arrow2/.style={decoration={markings,mark=at position 1 with %
    {\arrow[scale=2.3,>=latex]{>}}},postaction={decorate}}}  
    
\begin{figure}[htbp]
\centering
\subcaptionbox{Original Syn}{
\begin{minipage}[t]{0.16\linewidth}
\centering
\scalebox{0.45}{
        \begin{tikzpicture}[level distance=30pt,sibling distance=0pt]
        \Tree 
        [.S 
            [.NP [.N I ] [.N hope ] ] 
            [.S/N [.N you ]
            [.VP/N [.V can ] [.V/N enjoy ] ] ] 
        ]  
        \end{tikzpicture}
        }
\end{minipage}%
}%
\subcaptionbox{Modified Syn}{
\begin{minipage}[t]{0.2\linewidth}
\centering
  \scalebox{0.45}{
        \begin{tikzpicture}[level distance=30pt,sibling distance=5pt]
        \Tree 
        [.S 
            [.NP [.N I ] ]
            [.VP [.V hope ] 
            [.S [.N you ]
            [.VP [.V can ] [.V/N enjoy ] ] ] 
            ]
        ]  
        \end{tikzpicture}}
\end{minipage}%
}%
\subcaptionbox{Original Sem}{
\begin{minipage}[t]{0.26\linewidth}
\centering
 \scalebox{0.45}{
    \begin{tikzpicture}
        \node (By) [ersnode, ] {compound};
        
        \node (most) [ersnode, right = 0.2 of By ] {proper\_q};
        \node (alone) [ersnode, left = 0.2 of By ] {udef\_q};
        \node (measure) [ersnode, below right = 1 and 0.8 of By ] {named(I)};
        \node (growing) [ersnode, below = 1 of By ] {\_hope\_n\_1};
        \node (you) [ersnode, left = 0.1 of growing ] {pronoun\_q};
        \node (information) [ersnode, below = 1 of growing ] {\_enjoy\_v\_1};
         \node (younew) [ersnode, left = 1.2 of information ] {pron};
         \node (can) [ersnode, right = 1.2 of information ] {\_can\_v\_modal};
       
        \path[->] (information) edge [edgeA] node [labelA] {ARG2} (growing);
        \path[->] (information) edge [edgeA] node [labelA] {ARG1} (younew);
        \path[->] (By) edge [edgeA] node [labelA] {ARG2} (measure);
        \path[->] (By) edge [edgeA] node [labelA] {ARG1} (growing);
        \path[->] (most) edge [edgeA] node [labelA] {BV} (measure);
        \path[->] (alone) edge [edgeA] node [labelA] {BV} (growing);
        \path[->] (you) edge [edgeA] node [labelA] {BV} (younew);
       \path[->] (can) edge [edgeA] node [labelA] {ARG1} (information);
      \end{tikzpicture}
            } 
\end{minipage}%
}%
\subcaptionbox{Modified Sem}{
\begin{minipage}[t]{0.24\linewidth}
\centering
   \scalebox{0.45}{
    \begin{tikzpicture}
       \node (By) [ersnode, ] {\_hope\_v\_1};
        
        \node (alone) [ersnode, left = 0.5 of By ] {pronoun\_q};
        \node (measure) [ersnode, below right = 1 and -0.2 of By ] {\_can\_v\_modal};
        \node (growing) [ersnode, below left = 1 and -0.5 of By ] {pron};
         \node (new) [ersnode, below left = 1 and 0.99 of By ] {pronoun\_q};
        \node (information) [ersnode, below = 1 of measure ] {ellipsis\_ref};
        \node (most) [ersnode, below right = 1.15 and 0.2 of new ] {pron};
       
        \path[->] (By) edge [edgeA] node [labelA] {ARG2} (measure);
        \path[->] (By) edge [edgeA] node [labelA] {ARG1} (growing);
        \path[->] (information) edge [edgeA] node [labelA] {ARG1} (most);
         \path[->] (new) edge [edgeA] node [labelA] {BV} (most);
        \path[->] (alone) edge [edgeA] node [labelA] {BV} (growing);
        \path[->] (measure) edge [edgeA] node [labelA] {ARG1} (information);

      \end{tikzpicture}
            } 
\end{minipage}%
}%
\caption{Relevant original and modified syntactic and semantic analysis for Example \ref{ex12}.}
\label{fig:arguement}
\end{figure}
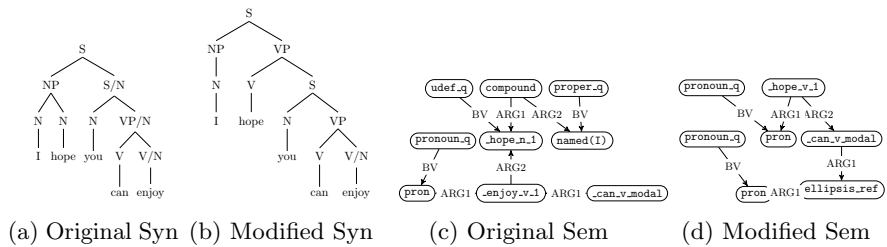

\es{\textit{A telephone is used when we need to contact \textbf{with} someone.}\label{ex13}}

\begin{table}[htbp]
\centering
\caption{Original and modified SHRG rules for Example \ref{ex13}.}
\label{tab:argument2}
\begin{tabular}{lllll}
\toprule
LHS & $\textrm{P}_{\textrm{ori}}$ & $\textrm{N}_{\textrm{ori}}$ & $\textrm{P}_{\textrm{ori}}$ & $\textrm{NP}_{\textrm{ori}}$ \\ \midrule
Syn & to & contact & with & N + PP \\ \addlinespace
Sem & 
\scalebox{0.6}{
\begin{tikzpicture}[>=latex,line join=bevel]
  \node (node1) at (1.8bp,33.8bp) [circle, inner sep=1.5pt,fill] {};
  \coordinate (node2) at (1.8bp,0.5bp);
  \draw [->] (node1) ..controls (1.8bp,27.979bp) and (1.8bp,18.64bp) .. (node2);
  \draw (28.3bp,16.5bp) node {\_used\_a\_to};
\end{tikzpicture}
} & 
\scalebox{0.6}{
\begin{tikzpicture}[>=latex,line join=bevel]
  \node (node1) at (1.8bp,33.8bp) [circle, inner sep=1.5pt,fill] {};
  \coordinate (node2) at (1.8bp,0.5bp);
  \draw [->] (node1) ..controls (1.8bp,27.979bp) and (1.8bp,18.64bp) .. (node2);
  \draw (30.3bp,16.5bp) node {\_contact\_n\_1};
\end{tikzpicture}
} & 
\scalebox{0.6}{
\begin{tikzpicture}[>=latex,line join=bevel]
  \node (node1) at (1.8bp,33.8bp) [circle, inner sep=1.5pt,fill] {};
  \coordinate (node2) at (1.8bp,0.5bp);
  \draw [->] (node1) ..controls (1.8bp,27.979bp) and (1.8bp,18.64bp) .. (node2);
  \draw (26.3bp,16.5bp) node {\_with\_p};
\end{tikzpicture}
} & 
\scalebox{0.6}{
\begin{tikzpicture}[>=latex,line join=bevel]
  \node (node1) at (12.5bp,68.4bp) [draw,circle, inner sep=1.5pt] {};
  \coordinate (node2) at (0.5bp,33.8bp);
  \node (node3) at (25.5bp,33.8bp) [draw,circle,inner sep=1.5pt,fill] {};
  \coordinate (node4) at (25.5bp,0.5bp);
  \draw [->] (node1) ..controls (9.4958bp,65.429bp) and (5.3665bp,62.358bp) .. (3.5bp,58.6bp) .. controls (1.4852bp,54.544bp) and (0.64086bp,49.639bp) .. (node2);
  \draw (8.5bp,51.1bp) node {PP};
  \draw [->] (node1) ..controls (13.985bp,63.677bp) and (18.077bp,53.416bp) .. (node3);
  \draw (39.5bp,51.1bp) node {ARG1};
  \draw [->] (node3) ..controls (25.5bp,27.979bp) and (25.5bp,18.64bp) .. (node4);
  \draw (32.5bp,16.5bp) node {N};
\end{tikzpicture}
} \\ \bottomrule
LHS & $\textrm{COMP}_{\textrm{mod}}$ & $\textrm{V}_{\textrm{mod}}$ & $\textrm{P}_{\textrm{mod}}$ & $\textrm{VP}_{\textrm{mod}}$ \\ \midrule
Syn & to & contact & with & V + NP \\ \addlinespace
Sem & $\varnothing$ & 
\scalebox{0.6}{
\begin{tikzpicture}[>=latex,line join=bevel]
  \node (node1) at (1.8bp,33.8bp) [circle, inner sep=1.5pt,fill] {};
  \coordinate (node2) at (1.8bp,0.5bp);
  \draw [->] (node1) ..controls (1.8bp,27.979bp) and (1.8bp,18.64bp) .. (node2);
  \draw (30.3bp,16.5bp) node {\_contact\_v\_1};
\end{tikzpicture}
} & $\varnothing$ & 
\scalebox{0.6}{
\begin{tikzpicture}[>=latex,line join=bevel]
  \node (node1) at (12.5bp,68.4bp) [draw,circle, inner sep=1.5pt,fill] {};
  \coordinate (node2) at (0.5bp,33.8bp);
  \node (node3) at (25.5bp,33.8bp) [draw,circle,inner sep=1.5pt] {};
  \coordinate (node4) at (25.5bp,0.5bp);
  \draw [->] (node1) ..controls (9.4958bp,65.429bp) and (5.3665bp,62.358bp) .. (3.5bp,58.6bp) .. controls (1.4852bp,54.544bp) and (0.64086bp,49.639bp) .. (node2);
  \draw (8.5bp,51.1bp) node {V};
  \draw [->] (node1) ..controls (13.985bp,63.677bp) and (18.077bp,53.416bp) .. (node3);
  \draw (39.5bp,51.1bp) node {ARG2};
  \draw [->] (node3) ..controls (25.5bp,27.979bp) and (25.5bp,18.64bp) .. (node4);
  \draw (32.5bp,16.5bp) node {NP};
\end{tikzpicture}
} \\ \bottomrule
\end{tabular}
\end{table}

\tikzset{ersnode/.style={right, rounded rectangle, draw, font=\tt}} 
\tikzset{ 
  nodeA/.style={right, fill=blue!40, text=white, rounded rectangle,font=\tt},
  nodeB/.style={right, fill=brown!50, text=white, rounded rectangle,font=\tt},
  edgeA/.style={thick, rounded corners,>=stealth'},
  edgeB/.style={thick, color=brown!50, rounded corners,>=stealth'},
  labelA/.style={anchor=center,fill=white,font=\scriptsize},
  labeltiny/.style={anchor=center,fill=white,font=\tiny},
  alertedge/.style={thick, color=red, rounded corners,>=stealth'}
}
\tikzset{arrow2/.style={decoration={markings,mark=at position 1 with %
    {\arrow[scale=2.3,>=latex]{>}}},postaction={decorate}}}  
    
\begin{figure}[htbp]
\centering
\subcaptionbox{Original Syn}{
\begin{minipage}[t]{0.22\linewidth}
\centering
\scalebox{0.45}{
        \begin{tikzpicture}[level distance=30pt,sibling distance=0pt]
        \Tree 
        [.PP 
            [.P to ] 
            [.NP [.N contact ]
            [.PP [.P with ] [.N someone ] ] ] 
        ]  
        \end{tikzpicture}
        }
\end{minipage}%
}%
\subcaptionbox{Modified Syn}{
\begin{minipage}[t]{0.22\linewidth}
\centering
  \scalebox{0.45}{
        \begin{tikzpicture}[level distance=30pt,sibling distance=5pt]
        \Tree 
        [.VP 
            [.COMP to ] 
            [.VP [.V contact ]
            [.NP [.P with ] [.N someone ] ] ] 
        ]  
        \end{tikzpicture}}
\end{minipage}%
}%
\subcaptionbox{Original Sem}{
\begin{minipage}[t]{0.22\linewidth}
\centering
 \scalebox{0.45}{
    \begin{tikzpicture}
        \node (By) [ersnode, ] {\_with\_p};
        
        \node (most) [ersnode, right = 0.5 of By ] {\_some\_q};
        \node (alone) [ersnode, left = 0.5 of By ] {udef\_q};
        \node (measure) [ersnode, below right = 1 and -0.2 of By ] {person};
        \node (growing) [ersnode, below left = 1 and -0.2 of By ] {\_contact\_n\_1};
        \node (information) [ersnode, below = 1 of growing ] {\_used\_a\_to};
       
        \path[->] (information) edge [edgeA] node [labelA] {ARG2} (growing);
        \path[->] (By) edge [edgeA] node [labelA] {ARG2} (measure);
        \path[->] (By) edge [edgeA] node [labelA] {ARG1} (growing);
        \path[->] (most) edge [edgeA] node [labelA] {BV} (measure);
        \path[->] (alone) edge [edgeA] node [labelA] {BV} (growing);
       
      \end{tikzpicture}
            } 
\end{minipage}%
}%
\subcaptionbox{Modified Sem}{
\begin{minipage}[t]{0.22\linewidth}
\centering
   \scalebox{0.45}{
    \begin{tikzpicture}
        \node (By) [ersnode, ] {\_contact\_v\_1};
        \node (most) [ersnode, right = 0.5 of By ] {\_some\_q};
        \node (measure) [ersnode, below right = 1 and -0.5 of By ] {person};

        \path[->] (By) edge [edgeA] node [labelA] {ARG2} (measure);
        \path[->] (most) edge [edgeA] node [labelA] {BV} (measure);
      \end{tikzpicture}
            } 
\end{minipage}%
}%
\caption{Relevant original and modified syntactic and semantic analysis for Example \ref{ex13}.}
\label{fig:arguement2}
\end{figure}
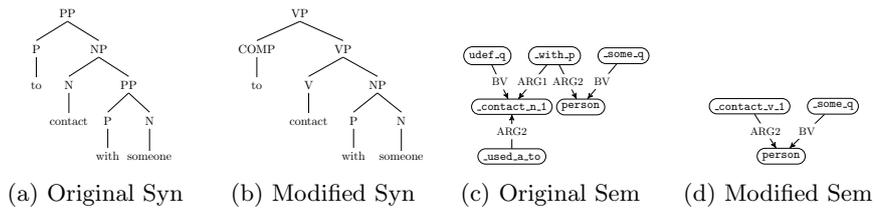

Broadly, we argue that the wide range of diverse form-meaning mappings in ESFL shown above, can be classified into three categories, each with a preferred handling strategy.  
\begin{itemize}
    \item \textbf{Diversity of Grammatical Forms}: ESFL demonstrates diversity in grammatical forms it employs to express grammatical meanings. Besides morphological markers, which are typically used by English, formal means like zero form, word order, and functional words, are also utilized by ESFL. This diversity is particularly evident in phenomena such as Number, Case, Tense, Aspect, Voice, and Agreement. To address this, analogous SHRG rules are adapted to account for these forms.
    \item \textbf{Diversity of Syntactic Derivations}: ESFL may impose unique constraints on syntactic derivations, exemplified by phenomena such as Binding, Ellipsis, and Filler-Gap. These phenomena are typically analyzed as outcomes of binding, deletion, or movement within a deep structure in mainstream generative grammar. In our framework, they are accommodated by establishing diverse mappings between syntactic configurations and their corresponding semantic interpretations.
    \item \textbf{Diversity of Lexical Items}: ESFL also exhibits significant lexical diversity, particularly in verbs, which vary in both the number of semantic arguments and the ways these arguments are realized. This variation, exemplified by the phenomenon of Argument Structure, can be addressed by adapting or extending existing SHRG rules.
\end{itemize}

It is also noteworthy that ESFL is grounded in English, an inflectional language with multifunctional grammatical markers, so constructions in above examples may overlap across categories.
For instance, Example \ref{ex8} reflects both Number and Person, while Example \ref{ex12} can be analyzed in terms of either Argument Structure or Ellipsis.
Despite such overlaps, we retain this set of phenomena because they are central to the acquisition of morphology, syntax, semantics, and their interfaces ---well-established topics in language acquisiton research. Therefore, covering them highlights not only the effectiveness of our SHRG-based constructivist approach but also the value of the resulting resorce for advancing research in second language acquisition.

\subsection{Rebuild}
For unparsable sentences\footnote{According to \citet{zhao-etal-2020-semantic}, 47.50\% of ESFL sentences, amounting to 2494 in total, are unparsable due to limitations of the ACE parser and inconsistent tokenization.}, we manually select composition rules from the SHRG inventory to generate syntactico-semantic representations.

As an initial investigation, we randomly select 100 of these unparsable sentences\footnote{Though limited in size, the sample is methodologically sound, as this rebuilding process mirrors that of most rejected graphs --- 81.2\% (665 cases) involved syntactic tree reconstruction, according to statistical analysis.}.
The annotation process is conducted in two phases. In the first phase, one annotator constructs semantic graphs by selecting appropriate SHRG rules, while the second annotator reviews them for acceptability. In the second phase, both annotators independently annotate the remaining 50 sentences to assess inter-annotator consistency.

Annotation results from both phases are presented in Table~\ref{consistency}. For the first 50 sentences, we report the proportion of accepted graphs to evaluate agreement between annotators. In the second phase, we compute the average S-match score across the 50 independently annotated graphs as an element-wise measure of consistency.

\begin{table}[htbp]
    \caption{Consistency scores in selecting rules.}
    \label{consistency}
    \begin{tabular}{@{}ccc@{}}\toprule
         &  First Phase & Second Phase\\\midrule
Consistency Score &94\%& 90.93\\\bottomrule
    \end{tabular}
\end{table}

\subsection{Summary}
As Table \ref{tab:summary} presents, following the three steps above, we develop an ESFL SemBank, which contains manually annotated syntactico-semantic representations of 1643 ESFL sentences. The resource is now can be accessed through \url{https://github.com/MandarinMeaningBank/ESFL_SemBank}.
\begin{table}[htbp]
    \caption{Development of ESFL SemBank at each step.}
    \label{tab:summary}
    \begin{tabular}{@{}ccc@{}}\toprule
        Step & Manner & Number \\\midrule
         1& Manually Accepted & 724\\
         2&Manually Modified & 819 \\
         3&Manually Composed & 100 \\\midrule
         && Overall: 1643 \\
         \bottomrule
    \end{tabular}
    
\end{table}

\section{A Study on the Linguistic Niche Hypothesis}

\label{compare}
To demonstrate the practical utility of our gold-standard syntactico-semantic resource for ESFL, along with its corresponding SHRG rules, we conduct an empirical study towards the Linguistic Niche Hypothesis  \citep[LNH;][]{lupyan2010language}. The LNH suggests that languages used in broad, communicative contexts, particularly those spoken by large populations of non-native speakers, tend to be more regular than those maintained within smaller, esoteric communities. Our resource, covering both ESFL and standard English, then provides a controlled testbed for evaluating this hypothesis.

\subsection{Syntactic Complexity}
We begin our empirical evaluation by comparing the syntactic structures of ESFL and native English, aiming to determine whether they exhibit statistically significant differences. 

Specifically, we analyze the distributions of CFG rules derived from SHRG-based derivations in our ESFL and English data, which involves a total of 620 unique rules.
To ensure statistical robustness and avoid distortions caused by low-frequency items, we filter out CFG rules with insufficient counts that may introduce noise or violate assumptions of normality. In particular, we retain only those rules with an expected frequency greater than 4, resulting in a focused subset of 77 CFG rules for analysis\footnote{According to standard guidelines for $\alpha = 0.05$, a minimum frequency of 4 corresponds to the safe sample size (Cohen's $d=0.5$).}.

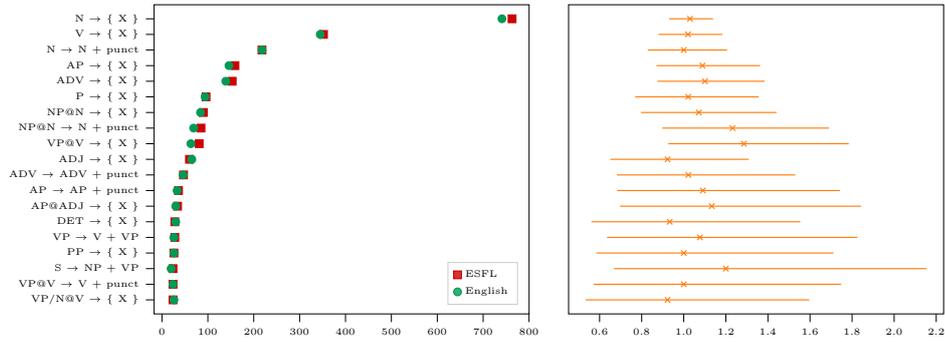
\begin{figure}[htbp]
\centering
\begin{minipage}[t]{0.525\textwidth}
\centering
\begin{tikzpicture}[scale=0.72, transform shape]

\definecolor{color0}{rgb}{0.12156862745098,0.466666666666667,0.705882352941177}
\definecolor{color1}{rgb}{1,0.498039215686275,0.0549019607843137}

\begin{axis}[
legend cell align={left},
legend style={
  fill opacity=0.8,
  draw opacity=1,
  text opacity=1,
  at={(0.97,0.03)},
  font=\tiny,
  anchor=south east,
  draw=white!80!black
},
tick align=outside,
tick pos=left,
x grid style={white!69.0196078431373!black},
xmin=-17.15, xmax=800.15,
xtick style={color=black},
xtick={-100,0,100,200,300,400,500,600,700,800,900},
xticklabels={\ensuremath{-}100,0,100,200,300,400,500,600,700,800,900},
xticklabel style = {font=\tiny},
y dir=reverse,
y grid style={white!69.0196078431373!black},
ymin=-0.9, ymax=18.9,
ytick style={color=black},
ytick={0,1,2,3,4,5,6,7,8,9,10,11,12,13,14,15,16,17,18},
yticklabel style = {font=\tiny},
yticklabels={
  N $\rightarrow$ \{ X \},
  V $\rightarrow$ \{ X \},
  N $\rightarrow$ N $+$ punct,
  AP $\rightarrow$ \{ X \},
  ADV $\rightarrow$ \{ X \},
  P $\rightarrow$ \{ X \},
  NP@N $\rightarrow$ \{ X \},
  NP@N $\rightarrow$ N $+$ punct,
  VP@V $\rightarrow$ \{ X \},
  ADJ $\rightarrow$ \{ X \},
  ADV $\rightarrow$ ADV $+$ punct,
  AP $\rightarrow$ AP $+$ punct,
  AP@ADJ $\rightarrow$ \{ X \},
  DET $\rightarrow$ \{ X \},
  VP $\rightarrow$ V $+$ VP,
  PP $\rightarrow$ \{ X \},
  S $\rightarrow$ NP $+$ VP,
  VP@V $\rightarrow$ V $+$ punct,
  VP/N@V $\rightarrow$ \{ X \}
}
]
\addplot [semithick, myred, mark=square*, mark size=2, mark options={solid}, only marks]
table {%
763 0
352 1
218 2
159 3
153 4
96 5
90 6
85 7
81 8
60 9
47 10
36 11
34 12
28 13
28 14
26 15
24 16
24 17
24 18
};
\addlegendentry{ESFL}
\addplot [semithick, ForestGreen, mark=*, mark size=2, mark options={solid}, only marks]
table {%
741 0
345 1
218 2
146 3
139 4
94 5
84 6
69 7
63 8
65 9
46 10
33 11
30 12
30 13
26 14
26 15
20 16
24 17
26 18
};
\addlegendentry{English}
\end{axis}

\end{tikzpicture}
\end{minipage}
\begin{minipage}[t]{0.435\textwidth}
\centering
\begin{tikzpicture}[scale=0.72, transform shape]

\definecolor{color0}{rgb}{1,0.498039215686275,0.0549019607843137}

\begin{axis}[
tick align=outside,
tick pos=left,
x grid style={white!69.0196078431373!black},
xmin=0.45264939466226, xmax=2.23509617773228,
xtick style={color=black},
xtick={0.4,0.6,0.8,1,1.2,1.4,1.6,1.8,2,2.2,2.4},
xticklabels={0.4,0.6,0.8,1.0,1.2,1.4,1.6,1.8,2.0,2.2,2.4},
xticklabel style = {font=\tiny},
y dir=reverse,
y grid style={white!69.0196078431373!black},
ymin=-0.9, ymax=18.9,
ytick style={draw=none},
ytick={0,1,2,3,4,5,6,7,8,9,10,11,12,13,14,15,16,17,18},
yticklabel style = {font=\tiny},
yticklabels={},
]
\path [draw=color0, semithick]
(axis cs:0.930726609845461,0)
--(axis cs:1.13917521957497,0);

\path [draw=color0, semithick]
(axis cs:0.879620038087161,1)
--(axis cs:1.18345574599181,1);

\path [draw=color0, semithick]
(axis cs:0.829062817491368,2)
--(axis cs:1.20618121920588,2);

\path [draw=color0, semithick]
(axis cs:0.870319130296634,3)
--(axis cs:1.36273059760729,3);

\path [draw=color0, semithick]
(axis cs:0.875290883851103,4)
--(axis cs:1.38420640924967,4);

\path [draw=color0, semithick]
(axis cs:0.769203924385642,5)
--(axis cs:1.35595497104215,5);

\path [draw=color0, semithick]
(axis cs:0.796702703082222,6)
--(axis cs:1.44088777310825,6);

\path [draw=color0, semithick]
(axis cs:0.897863952418024,7)
--(axis cs:1.69016511710519,7);

\path [draw=color0, semithick]
(axis cs:0.926385061658549,8)
--(axis cs:1.7844212875478,8);

\path [draw=color0, semithick]
(axis cs:0.651053624952617,9)
--(axis cs:1.30875702593496,9);

\path [draw=color0, semithick]
(axis cs:0.682321597124867,10)
--(axis cs:1.5299982516464,10);

\path [draw=color0, semithick]
(axis cs:0.683134770754098,11)
--(axis cs:1.74209057359852,11);

\path [draw=color0, semithick]
(axis cs:0.696982552911811,12)
--(axis cs:1.8428645581993,12);

\path [draw=color0, semithick]
(axis cs:0.560743540511762,13)
--(axis cs:1.55349290393268,13);

\path [draw=color0, semithick]
(axis cs:0.635362424183716,14)
--(axis cs:1.8253571024435,14);

\path [draw=color0, semithick]
(axis cs:0.58442544161441,15)
--(axis cs:1.71108225069328,15);

\path [draw=color0, semithick]
(axis cs:0.668500130589089,16)
--(axis cs:2.15407586941091,16);

\path [draw=color0, semithick]
(axis cs:0.572057592662607,17)
--(axis cs:1.74807574067073,17);

\path [draw=color0, semithick]
(axis cs:0.533669702983624,18)
--(axis cs:1.59662615500454,18);

\addplot [semithick, color0, mark=x, mark size=2, mark options={solid}, only marks]
table {%
1.02968960863698 0
1.02028985507246 1
1 2
1.08904109589041 3
1.10071942446043 4
1.02127659574468 5
1.07142857142857 6
1.23188405797101 7
1.28571428571429 8
0.923076923076923 9
1.02173913043478 10
1.09090909090909 11
1.13333333333333 12
0.933333333333333 13
1.07692307692308 14
1 15
1.2 16
1 17
0.923076923076923 18
};
\end{axis}

\end{tikzpicture}
\end{minipage}
\caption{Frequency Distribution of top 20 CFG Rules and their ratios with 95\% confidence intervals in ESFL and English Data}
\label{fig:com-frequency-safe-size}
\end{figure}

Figure~\ref{fig:com-frequency-safe-size} displays the frequencies and relative ratios of the 20 most frequently used CFG rules, allowing for an initial visual comparison between ESFL and English syntactic usage.

\begin{figure}[htbp]
\centering
\begin{minipage}[t]{0.56\textwidth}
\centering
\begin{tikzpicture}[scale=0.72, transform shape]
\definecolor{color0}{rgb}{0.12156862745098,0.466666666666667,0.705882352941177}
\definecolor{color1}{rgb}{1,0.498039215686275,0.0549019607843137}

\begin{axis}[
legend cell align={left},
legend style={
  fill opacity=0.8,
  draw opacity=1,
  text opacity=1,
  at={(0.97,0.03)},
  font=\tiny,
  anchor=south east,
  draw=white!80!black
},
tick align=outside,
tick pos=left,
x grid style={white!69.0196078431373!black},
xmin=4.9, xmax=29.1,
xtick style={color=black},
xticklabel style = {font=\tiny},
y dir=reverse,
y grid style={white!69.0196078431373!black},
ymin=-0.95, ymax=19.95,
ytick style={color=black},
yticklabel style = {font=\tiny},
ytick={0,1,2,3,4,5,6,7,8,9,10,11,12,13,14,15,16,17,18,19},
yticklabels={
  VP $\rightarrow$ V $+$ VP,
  S $\rightarrow$ NP $+$ VP,
  S/PP $\rightarrow$ NP $+$ VP/PP,
  VP $\rightarrow$ V $+$ PP,
  S $\rightarrow$ N $+$ VP,
  VP $\rightarrow$ VP $+$ VP-C,
  VP/PP $\rightarrow$ V $+$ VP/PP@VP,
  S $\rightarrow$ NP@N $+$ VP,
  N $\rightarrow$ N $+$ N,
  VP $\rightarrow$ VP $+$ PP,
  VP $\rightarrow$ ADV $+$ VP,
  VP/PP $\rightarrow$ V $+$ VP/PP,
  ROOT $\rightarrow$ S $+$ S,
  VP $\rightarrow$ VP@V $+$ PP,
  VP $\rightarrow$ V $+$ AP,
  S/N $\rightarrow$ NP $+$ VP/N,
  V $\rightarrow$ V $+$ NP,
  VP $\rightarrow$ V $+$ VP@V,
  N $\rightarrow$ DET $+$ N,
  VP/N $\rightarrow$ V $+$ VP/N
}
]
\addplot [semithick, myred, mark=square*, mark size=2, mark options={solid}, only marks]
table {%
28 0
24 1
12 2
13 3
13 4
11 5
9 6
11 7
11 8
10 9
10 10
10 11
8 12
8 13
8 14
7 15
7 16
7 17
7 18
6 19
};
\addlegendentry{ESFL}
\addplot [semithick, ForestGreen, mark=*, mark size=2, mark options={solid}, only marks]
table {%
26 0
20 1
13 2
11 3
9 4
10 5
11 6
9 7
9 8
9 9
9 10
8 11
10 12
9 13
8 14
8 15
7 16
7 17
7 18
7 19
};
\addlegendentry{English}
\end{axis}

\end{tikzpicture}
\end{minipage}
\begin{minipage}[t]{0.41\textwidth}
\centering
\begin{tikzpicture}[scale=0.72, transform shape]
\definecolor{color0}{rgb}{1,0.498039215686275,0.0549019607843137}

\begin{axis}[
tick align=outside,
tick pos=left,
x grid style={white!69.0196078431373!black},
xmin=0.152287594406379, xmax=3.44994032362268,
xtick style={color=black},
xticklabel style = {font=\tiny},
y dir=reverse,
y grid style={white!69.0196078431373!black},
ymin=-0.95, ymax=19.95,
ytick style={draw=none},
yticklabel style = {font=\tiny},
ytick={0,1,2,3,4,5,6,7,8,9,10,11,12,13,14,15,16,17,18,19},
yticklabels={
}
]

\path [draw=color0, semithick]
(axis cs:0.622471012678539,0)
--(axis cs:1.78832087137943,0);

\path [draw=color0, semithick]
(axis cs:0.654936359823514,1)
--(axis cs:2.11036998220547,1);

\path [draw=color0, semithick]
(axis cs:0.61941135616611,2)
--(axis cs:3.23308954202677,2);

\path [draw=color0, semithick]
(axis cs:0.529266995073962,3)
--(axis cs:2.53292151133399,3);

\path [draw=color0, semithick]
(axis cs:0.420495946257237,4)
--(axis cs:1.94495332910508,4);

\path [draw=color0, semithick]
(axis cs:0.509396639350199,5)
--(axis cs:2.81474778720196,5);

\path [draw=color0, semithick]
(axis cs:0.468905580797452,6)
--(axis cs:2.47682418152139,6);

\path [draw=color0, semithick]
(axis cs:0.509396639350199,7)
--(axis cs:2.81474778720196,7);

\path [draw=color0, semithick]
(axis cs:0.454868899690605,8)
--(axis cs:2.60509729104119,8);

\path [draw=color0, semithick]
(axis cs:0.454868899690605,9)
--(axis cs:2.60509729104119,9);

\path [draw=color0, semithick]
(axis cs:0.498340623149741,10)
--(axis cs:3.00946256525606,10);

\path [draw=color0, semithick]
(axis cs:0.341001056093935,11)
--(axis cs:1.88425265093685,11);

\path [draw=color0, semithick]
(axis cs:0.380983981669638,12)
--(axis cs:2.51934993137384,12);

\path [draw=color0, semithick]
(axis cs:0.347177519490359,13)
--(axis cs:2.18443219960071,13);

\path [draw=color0, semithick]
(axis cs:0.318937998815834,14)
--(axis cs:1.92605604176388,14);

\path [draw=color0, semithick]
(axis cs:0.358670993563339,15)
--(axis cs:2.6760791513642,15);

\path [draw=color0, semithick]
(axis cs:0.358670993563339,16)
--(axis cs:2.6760791513642,16);

\path [draw=color0, semithick]
(axis cs:0.323255741927525,17)
--(axis cs:2.27334353343479,17);

\path [draw=color0, semithick]
(axis cs:0.402956527410212,18)
--(axis cs:3.24213126647063,18);

\path [draw=color0, semithick]
(axis cs:0.402956527410212,19)
--(axis cs:3.24213126647063,19);
\addplot [semithick, color0, mark=*, mark size=1.5, mark options={solid}, only marks]
table {%
1.05507246376812 0
1.17565217391304 1
1.41513687600644 2
1.15783926218709 3
0.904347826086957 4
1.19742351046699 5
1.07768115942029 6
1.19742351046699 7
1.08856682769726 8
1.08856682769726 9
1.22463768115942 10
0.801581027667984 11
0.979710144927536 12
0.87085346215781 13
0.783768115942029 14
0.979710144927536 15
0.979710144927536 16
0.857246376811594 17
1.14299516908213 18
1.14299516908213 19
};
\end{axis}

\end{tikzpicture}

\end{minipage}
\caption{Frequency Distribution of top 20 non-lexical CFG Rules and their ratios with 95\% confidence intervals in ESFL and English Data}
\label{fig:com-frequency}
\end{figure}
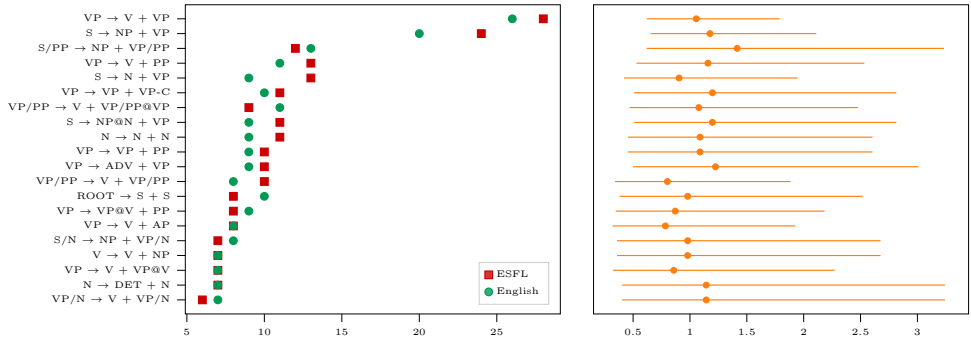
Furthermore, since certain CFG rules, those containing only a lexical node or a lexical node combined with a punctuation mark, primarily affect surface realizations and offer limited insight into underlying syntactic structures, we exclude them from our analysis and focus instead on the 43 retained non-lexical CFG rules.
Figure~\ref{fig:com-frequency} presents the frequencies and relative ratios of the top 20 non-lexical CFG rules across both the ESFL and English data.

A Chi-Square Test is then performed to determine whether the distribution of remaining CFG rules differs significantly between ESFL and English.
The results of this test, as summarized in Table~\ref{tab:chi-square}, reveal that the syntactic profiles of the two corpora does not diverge in a statistically meaningful way.

\begin{table}[htbp]
\centering
\caption{Chi-Square Test of Independence on non-lexical CFG rule distributions between ESFL and English.}
\label{tab:chi-square}
\begin{tabular}{ccc}
\toprule
Degrees of Freedom  &  $\chi^2$ & $p$-value \\\midrule
42 & 4.244  & 0.999 \\\bottomrule
\end{tabular}
\end{table}


\subsection{Semantic Transparency}
We then turn to the concept of semantic transparency --- the extent to which syntactic structures reliably map onto semantic representations --- in both our ESFL and English data. This investigation addresses whether ESFL exhibits a more systematic syntax–semantics mapping compared to English.

To be more specific, we design a targeted experiment based on semantic derivation stability --- we examine the degree to which meaning is preserved when the original SHRG rules are systematically replaced with the most frequent alternatives sharing the same CFG rules (i.e., identical syntactic structures). Semantic graphs are then regenerated from these substituted rules, and their fidelity to the gold-standard annotations is evaluated using S-match scores. Higher post-substitution S-match scores indicate greater semantic transparency, reflecting a stronger and more consistent coupling between surface-level syntactic choices and underlying semantic interpretations.

\begin{figure}[htbp]
\centering
\begin{minipage}[t]{0.48\textwidth}
\centering

\begin{tikzpicture}[scale=0.75,transform shape]

\definecolor{color0}{rgb}{0.194607843137255,0.453431372549019,0.632843137254902}
\definecolor{color1}{rgb}{0.881862745098039,0.505392156862745,0.173039215686275}

\begin{axis}[
tick align=outside,
tick pos=left,
title={Violin Plot},
x grid style={white!69.0196078431373!black},
xmin=-0.5, xmax=1.5,
xtick style={color=black},
xtick={0,1},
xticklabels={ESFL, English},
y grid style={white!69.0196078431373!black},
ylabel={S-match score},
ymin=0.490458066464575, ymax=1.06338808753542,
ytick style={color=black},
ytick={0.4,0.5,0.6,0.7,0.8,0.9,1,1.1},
yticklabels={0.4,0.5,0.6,0.7,0.8,0.9,1.0,1.1}
]
\path [draw=white!24.7058823529412!black, fill=color0, semithick]
(axis cs:0.00033849120538979,0.631360624740933)
--(axis cs:-0.00033849120538979,0.631360624740933)
--(axis cs:-0.00052719910894137,0.63544088148354)
--(axis cs:-0.000780699656302879,0.639521138226148)
--(axis cs:-0.00110045149524591,0.643601394968755)
--(axis cs:-0.00147881748275446,0.647681651711362)
--(axis cs:-0.00189862678210742,0.65176190845397)
--(axis cs:-0.00233557517441842,0.655842165196577)
--(axis cs:-0.00276339168982192,0.659922421939185)
--(axis cs:-0.00316063130498714,0.664002678681792)
--(axis cs:-0.00351725839197487,0.6680829354244)
--(axis cs:-0.00383926387118105,0.672163192167007)
--(axis cs:-0.00415037169662564,0.676243448909614)
--(axis cs:-0.00449092438671418,0.680323705652222)
--(axis cs:-0.00491463428145291,0.684403962394829)
--(axis cs:-0.00548374247338566,0.688484219137437)
--(axis cs:-0.00626257524173723,0.692564475880044)
--(axis cs:-0.00730928805400709,0.696644732622651)
--(axis cs:-0.00866630143882193,0.700724989365259)
--(axis cs:-0.0103513495396245,0.704805246107866)
--(axis cs:-0.0123521715416049,0.708885502850474)
--(axis cs:-0.0146275282091623,0.712965759593081)
--(axis cs:-0.0171150576422174,0.717046016335689)
--(axis cs:-0.0197434238230379,0.721126273078296)
--(axis cs:-0.0224440836512737,0.725206529820903)
--(axis cs:-0.0251583791747109,0.729286786563511)
--(axis cs:-0.0278387032217952,0.733367043306118)
--(axis cs:-0.030446552372837,0.737447300048726)
--(axis cs:-0.0329527743008839,0.741527556791333)
--(axis cs:-0.0353442970258837,0.745607813533941)
--(axis cs:-0.0376372242859223,0.749688070276548)
--(axis cs:-0.0398908086673832,0.753768327019155)
--(axis cs:-0.042213889201322,0.757848583761763)
--(axis cs:-0.0447572151565897,0.76192884050437)
--(axis cs:-0.0476912709325855,0.766009097246978)
--(axis cs:-0.0511766544540978,0.770089353989585)
--(axis cs:-0.0553386351572565,0.774169610732192)
--(axis cs:-0.0602565072494369,0.7782498674748)
--(axis cs:-0.0659719689411109,0.782330124217407)
--(axis cs:-0.0725119402448459,0.786410380960015)
--(axis cs:-0.0799140122863992,0.790490637702622)
--(axis cs:-0.0882402836823656,0.79457089444523)
--(axis cs:-0.0975688449345605,0.798651151187837)
--(axis cs:-0.107960615707306,0.802731407930444)
--(axis cs:-0.119409936748577,0.806811664673052)
--(axis cs:-0.131796580212863,0.810891921415659)
--(axis cs:-0.1448608502034,0.814972178158267)
--(axis cs:-0.158219504593451,0.819052434900874)
--(axis cs:-0.171428208392859,0.823132691643481)
--(axis cs:-0.184079392524311,0.827212948386089)
--(axis cs:-0.19590876680326,0.831293205128696)
--(axis cs:-0.206875891337293,0.835373461871304)
--(axis cs:-0.217188365516332,0.839453718613911)
--(axis cs:-0.227255097589343,0.843533975356519)
--(axis cs:-0.237576791044872,0.847614232099126)
--(axis cs:-0.248603273782669,0.851694488841733)
--(axis cs:-0.260599701804598,0.855774745584341)
--(axis cs:-0.27356222188346,0.859855002326948)
--(axis cs:-0.287208654049007,0.863935259069556)
--(axis cs:-0.301046405730645,0.868015515812163)
--(axis cs:-0.314496267252499,0.87209577255477)
--(axis cs:-0.327034679916329,0.876176029297378)
--(axis cs:-0.33831298797167,0.880256286039985)
--(axis cs:-0.348220230242027,0.884336542782593)
--(axis cs:-0.356872920747131,0.8884167995252)
--(axis cs:-0.364536177365161,0.892497056267808)
--(axis cs:-0.371500488144828,0.896577313010415)
--(axis cs:-0.377952574794679,0.900657569753022)
--(axis cs:-0.383882783493399,0.90473782649563)
--(axis cs:-0.389062188919346,0.908818083238237)
--(axis cs:-0.393100714808558,0.912898339980845)
--(axis cs:-0.395568958902757,0.916978596723452)
--(axis cs:-0.396141499054499,0.92105885346606)
--(axis cs:-0.394709171400245,0.925139110208667)
--(axis cs:-0.391418015650036,0.929219366951274)
--(axis cs:-0.386620366704492,0.933299623693882)
--(axis cs:-0.380757892886326,0.937379880436489)
--(axis cs:-0.374222962451662,0.941460137179097)
--(axis cs:-0.367252538852061,0.945540393921704)
--(axis cs:-0.359894776850124,0.949620650664311)
--(axis cs:-0.352058332260731,0.953700907406919)
--(axis cs:-0.34361986934791,0.957781164149526)
--(axis cs:-0.334539545179192,0.961861420892134)
--(axis cs:-0.324927499494702,0.965941677634741)
--(axis cs:-0.315020495263352,0.970021934377349)
--(axis cs:-0.305063079318282,0.974102191119956)
--(axis cs:-0.295130797835948,0.978182447862563)
--(axis cs:-0.284967819934121,0.982262704605171)
--(axis cs:-0.273921201519724,0.986342961347778)
--(axis cs:-0.261029121851835,0.990423218090386)
--(axis cs:-0.245264572359588,0.994503474832993)
--(axis cs:-0.225868997814173,0.9985837315756)
--(axis cs:-0.202662009883512,1.00266398831821)
--(axis cs:-0.176209203412633,1.00674424506082)
--(axis cs:-0.14777700810763,1.01082450180342)
--(axis cs:-0.119082855778409,1.01490475854603)
--(axis cs:-0.0919250139421363,1.01898501528864)
--(axis cs:-0.0678150950448714,1.02306527203124)
--(axis cs:-0.0477230460422167,1.02714552877385)
--(axis cs:-0.0319908141542945,1.03122578551646)
--(axis cs:-0.0204055135661317,1.03530604225907)
--(axis cs:0.0204055135661317,1.03530604225907)
--(axis cs:0.0204055135661317,1.03530604225907)
--(axis cs:0.0319908141542945,1.03122578551646)
--(axis cs:0.0477230460422167,1.02714552877385)
--(axis cs:0.0678150950448714,1.02306527203124)
--(axis cs:0.0919250139421363,1.01898501528864)
--(axis cs:0.119082855778409,1.01490475854603)
--(axis cs:0.14777700810763,1.01082450180342)
--(axis cs:0.176209203412633,1.00674424506082)
--(axis cs:0.202662009883512,1.00266398831821)
--(axis cs:0.225868997814173,0.9985837315756)
--(axis cs:0.245264572359588,0.994503474832993)
--(axis cs:0.261029121851835,0.990423218090386)
--(axis cs:0.273921201519724,0.986342961347778)
--(axis cs:0.284967819934121,0.982262704605171)
--(axis cs:0.295130797835948,0.978182447862563)
--(axis cs:0.305063079318282,0.974102191119956)
--(axis cs:0.315020495263352,0.970021934377349)
--(axis cs:0.324927499494702,0.965941677634741)
--(axis cs:0.334539545179192,0.961861420892134)
--(axis cs:0.34361986934791,0.957781164149526)
--(axis cs:0.352058332260731,0.953700907406919)
--(axis cs:0.359894776850124,0.949620650664311)
--(axis cs:0.367252538852061,0.945540393921704)
--(axis cs:0.374222962451662,0.941460137179097)
--(axis cs:0.380757892886326,0.937379880436489)
--(axis cs:0.386620366704492,0.933299623693882)
--(axis cs:0.391418015650036,0.929219366951274)
--(axis cs:0.394709171400245,0.925139110208667)
--(axis cs:0.396141499054499,0.92105885346606)
--(axis cs:0.395568958902757,0.916978596723452)
--(axis cs:0.393100714808558,0.912898339980845)
--(axis cs:0.389062188919346,0.908818083238237)
--(axis cs:0.383882783493399,0.90473782649563)
--(axis cs:0.377952574794679,0.900657569753022)
--(axis cs:0.371500488144828,0.896577313010415)
--(axis cs:0.364536177365161,0.892497056267808)
--(axis cs:0.356872920747131,0.8884167995252)
--(axis cs:0.348220230242027,0.884336542782593)
--(axis cs:0.33831298797167,0.880256286039985)
--(axis cs:0.327034679916329,0.876176029297378)
--(axis cs:0.314496267252499,0.87209577255477)
--(axis cs:0.301046405730645,0.868015515812163)
--(axis cs:0.287208654049007,0.863935259069556)
--(axis cs:0.27356222188346,0.859855002326948)
--(axis cs:0.260599701804598,0.855774745584341)
--(axis cs:0.248603273782669,0.851694488841733)
--(axis cs:0.237576791044872,0.847614232099126)
--(axis cs:0.227255097589343,0.843533975356519)
--(axis cs:0.217188365516332,0.839453718613911)
--(axis cs:0.206875891337293,0.835373461871304)
--(axis cs:0.19590876680326,0.831293205128696)
--(axis cs:0.184079392524311,0.827212948386089)
--(axis cs:0.171428208392859,0.823132691643481)
--(axis cs:0.158219504593451,0.819052434900874)
--(axis cs:0.1448608502034,0.814972178158267)
--(axis cs:0.131796580212863,0.810891921415659)
--(axis cs:0.119409936748577,0.806811664673052)
--(axis cs:0.107960615707306,0.802731407930444)
--(axis cs:0.0975688449345605,0.798651151187837)
--(axis cs:0.0882402836823656,0.79457089444523)
--(axis cs:0.0799140122863992,0.790490637702622)
--(axis cs:0.0725119402448459,0.786410380960015)
--(axis cs:0.0659719689411109,0.782330124217407)
--(axis cs:0.0602565072494369,0.7782498674748)
--(axis cs:0.0553386351572565,0.774169610732192)
--(axis cs:0.0511766544540978,0.770089353989585)
--(axis cs:0.0476912709325855,0.766009097246978)
--(axis cs:0.0447572151565897,0.76192884050437)
--(axis cs:0.042213889201322,0.757848583761763)
--(axis cs:0.0398908086673832,0.753768327019155)
--(axis cs:0.0376372242859223,0.749688070276548)
--(axis cs:0.0353442970258837,0.745607813533941)
--(axis cs:0.0329527743008839,0.741527556791333)
--(axis cs:0.030446552372837,0.737447300048726)
--(axis cs:0.0278387032217952,0.733367043306118)
--(axis cs:0.0251583791747109,0.729286786563511)
--(axis cs:0.0224440836512737,0.725206529820903)
--(axis cs:0.0197434238230379,0.721126273078296)
--(axis cs:0.0171150576422174,0.717046016335689)
--(axis cs:0.0146275282091623,0.712965759593081)
--(axis cs:0.0123521715416049,0.708885502850474)
--(axis cs:0.0103513495396245,0.704805246107866)
--(axis cs:0.00866630143882193,0.700724989365259)
--(axis cs:0.00730928805400709,0.696644732622651)
--(axis cs:0.00626257524173723,0.692564475880044)
--(axis cs:0.00548374247338566,0.688484219137437)
--(axis cs:0.00491463428145291,0.684403962394829)
--(axis cs:0.00449092438671418,0.680323705652222)
--(axis cs:0.00415037169662564,0.676243448909614)
--(axis cs:0.00383926387118105,0.672163192167007)
--(axis cs:0.00351725839197487,0.6680829354244)
--(axis cs:0.00316063130498714,0.664002678681792)
--(axis cs:0.00276339168982192,0.659922421939185)
--(axis cs:0.00233557517441842,0.655842165196577)
--(axis cs:0.00189862678210742,0.65176190845397)
--(axis cs:0.00147881748275446,0.647681651711362)
--(axis cs:0.00110045149524591,0.643601394968755)
--(axis cs:0.000780699656302879,0.639521138226148)
--(axis cs:0.00052719910894137,0.63544088148354)
--(axis cs:0.00033849120538979,0.631360624740933)
--cycle;

\path [draw=white!24.7058823529412!black, fill=color1, semithick]
(axis cs:1.00031399003482,0.516500340149614)
--(axis cs:0.99968600996518,0.516500340149614)
--(axis cs:0.99946984609306,0.521761405540531)
--(axis cs:0.999173176470887,0.527022470931448)
--(axis cs:0.998808899112633,0.532283536322364)
--(axis cs:0.998415072248775,0.537544601713281)
--(axis cs:0.998051965102017,0.542805667104198)
--(axis cs:0.997788361353047,0.548066732495115)
--(axis cs:0.997680605320236,0.553327797886032)
--(axis cs:0.997752905041611,0.558588863276949)
--(axis cs:0.997988131515498,0.563849928667866)
--(axis cs:0.998333538250928,0.569110994058783)
--(axis cs:0.998718176310912,0.5743720594497)
--(axis cs:0.99907319225325,0.579633124840616)
--(axis cs:0.999346097816529,0.584894190231533)
--(axis cs:0.999505030778788,0.59015525562245)
--(axis cs:0.999535345867037,0.595416321013367)
--(axis cs:0.999434605696913,0.600677386404284)
--(axis cs:0.999211348958935,0.605938451795201)
--(axis cs:0.998888817389176,0.611199517186118)
--(axis cs:0.998509878061854,0.616460582577035)
--(axis cs:0.99813678692956,0.621721647967952)
--(axis cs:0.99784112104806,0.626982713358869)
--(axis cs:0.997684615476623,0.632243778749785)
--(axis cs:0.997697582357888,0.637504844140702)
--(axis cs:0.997863958073699,0.642765909531619)
--(axis cs:0.998118830537866,0.648026974922536)
--(axis cs:0.998357618695471,0.653288040313453)
--(axis cs:0.998450804336964,0.65854910570437)
--(axis cs:0.998258080849242,0.663810171095287)
--(axis cs:0.997640423333855,0.669071236486204)
--(axis cs:0.996473223008095,0.674332301877121)
--(axis cs:0.994663404940249,0.679593367268038)
--(axis cs:0.992167787963099,0.684854432658954)
--(axis cs:0.989003701889927,0.690115498049871)
--(axis cs:0.985242599228071,0.695376563440788)
--(axis cs:0.980985252080495,0.700637628831705)
--(axis cs:0.976327855846536,0.705898694222622)
--(axis cs:0.971333155849531,0.711159759613539)
--(axis cs:0.966015529475895,0.716420825004456)
--(axis cs:0.960339170476139,0.721681890395373)
--(axis cs:0.954223834771325,0.72694295578629)
--(axis cs:0.947556349981255,0.732204021177207)
--(axis cs:0.940210865020157,0.737465086568123)
--(axis cs:0.932077006654258,0.74272615195904)
--(axis cs:0.923083827854774,0.747987217349957)
--(axis cs:0.913202106738153,0.753248282740874)
--(axis cs:0.902419773088053,0.758509348131791)
--(axis cs:0.890709364126213,0.763770413522708)
--(axis cs:0.878021297969246,0.769031478913625)
--(axis cs:0.864325077425201,0.774292544304542)
--(axis cs:0.849686435940948,0.779553609695459)
--(axis cs:0.83433555441773,0.784814675086375)
--(axis cs:0.818674130300428,0.790075740477292)
--(axis cs:0.803194828063164,0.795336805868209)
--(axis cs:0.788333002687889,0.800597871259126)
--(axis cs:0.774313105253226,0.805858936650043)
--(axis cs:0.761065574296186,0.81112000204096)
--(axis cs:0.748260336340691,0.816381067431877)
--(axis cs:0.735440261253171,0.821642132822794)
--(axis cs:0.722179374373466,0.82690319821371)
--(axis cs:0.708181885227035,0.832164263604627)
--(axis cs:0.693295336431297,0.837425328995544)
--(axis cs:0.677495562019418,0.842686394386461)
--(axis cs:0.660938260945877,0.847947459777378)
--(axis cs:0.644113420579701,0.853208525168295)
--(axis cs:0.628016169238649,0.858469590559212)
--(axis cs:0.614162320878665,0.863730655950129)
--(axis cs:0.604320411393093,0.868991721341046)
--(axis cs:0.6,0.874252786731963)
--(axis cs:0.601916230779718,0.87951385212288)
--(axis cs:0.609705786589725,0.884774917513796)
--(axis cs:0.622047954714989,0.890035982904713)
--(axis cs:0.637124625719118,0.89529704829563)
--(axis cs:0.653181308883049,0.900558113686547)
--(axis cs:0.668931645506563,0.905819179077464)
--(axis cs:0.683673057977743,0.911080244468381)
--(axis cs:0.697153737790905,0.916341309859298)
--(axis cs:0.709343938902908,0.921602375250215)
--(axis cs:0.720269331008186,0.926863440641132)
--(axis cs:0.729985050542673,0.932124506032048)
--(axis cs:0.738667738821483,0.937385571422965)
--(axis cs:0.746734059397554,0.942646636813882)
--(axis cs:0.754882177614975,0.947907702204799)
--(axis cs:0.763992995944512,0.953168767595716)
--(axis cs:0.774899994419713,0.958429832986633)
--(axis cs:0.788111807493164,0.96369089837755)
--(axis cs:0.803616264152558,0.968951963768467)
--(axis cs:0.820877423308559,0.974213029159384)
--(axis cs:0.839050503034652,0.979474094550301)
--(axis cs:0.857319358798091,0.984735159941217)
--(axis cs:0.875179662300091,0.989996225332134)
--(axis cs:0.892512251960864,0.995257290723051)
--(axis cs:0.909415898474221,1.00051835611397)
--(axis cs:0.925919367542751,1.00577942150488)
--(axis cs:0.941764426886293,1.0110404868958)
--(axis cs:0.956394253266979,1.01630155228672)
--(axis cs:0.969141340878761,1.02156261767764)
--(axis cs:0.979488649612433,1.02682368306855)
--(axis cs:0.987254332812076,1.03208474845947)
--(axis cs:0.992621018371184,1.03734581385039)
--(axis cs:1.00737898162882,1.03734581385039)
--(axis cs:1.00737898162882,1.03734581385039)
--(axis cs:1.01274566718792,1.03208474845947)
--(axis cs:1.02051135038757,1.02682368306855)
--(axis cs:1.03085865912124,1.02156261767764)
--(axis cs:1.04360574673302,1.01630155228672)
--(axis cs:1.05823557311371,1.0110404868958)
--(axis cs:1.07408063245725,1.00577942150488)
--(axis cs:1.09058410152578,1.00051835611397)
--(axis cs:1.10748774803914,0.995257290723051)
--(axis cs:1.12482033769991,0.989996225332134)
--(axis cs:1.14268064120191,0.984735159941217)
--(axis cs:1.16094949696535,0.979474094550301)
--(axis cs:1.17912257669144,0.974213029159384)
--(axis cs:1.19638373584744,0.968951963768467)
--(axis cs:1.21188819250684,0.96369089837755)
--(axis cs:1.22510000558029,0.958429832986633)
--(axis cs:1.23600700405549,0.953168767595716)
--(axis cs:1.24511782238502,0.947907702204799)
--(axis cs:1.25326594060245,0.942646636813882)
--(axis cs:1.26133226117852,0.937385571422965)
--(axis cs:1.27001494945733,0.932124506032048)
--(axis cs:1.27973066899181,0.926863440641132)
--(axis cs:1.29065606109709,0.921602375250215)
--(axis cs:1.3028462622091,0.916341309859298)
--(axis cs:1.31632694202226,0.911080244468381)
--(axis cs:1.33106835449344,0.905819179077464)
--(axis cs:1.34681869111695,0.900558113686547)
--(axis cs:1.36287537428088,0.89529704829563)
--(axis cs:1.37795204528501,0.890035982904713)
--(axis cs:1.39029421341027,0.884774917513796)
--(axis cs:1.39808376922028,0.87951385212288)
--(axis cs:1.4,0.874252786731963)
--(axis cs:1.39567958860691,0.868991721341046)
--(axis cs:1.38583767912133,0.863730655950129)
--(axis cs:1.37198383076135,0.858469590559212)
--(axis cs:1.3558865794203,0.853208525168295)
--(axis cs:1.33906173905412,0.847947459777378)
--(axis cs:1.32250443798058,0.842686394386461)
--(axis cs:1.3067046635687,0.837425328995544)
--(axis cs:1.29181811477297,0.832164263604627)
--(axis cs:1.27782062562653,0.82690319821371)
--(axis cs:1.26455973874683,0.821642132822794)
--(axis cs:1.25173966365931,0.816381067431877)
--(axis cs:1.23893442570381,0.81112000204096)
--(axis cs:1.22568689474677,0.805858936650043)
--(axis cs:1.21166699731211,0.800597871259126)
--(axis cs:1.19680517193684,0.795336805868209)
--(axis cs:1.18132586969957,0.790075740477292)
--(axis cs:1.16566444558227,0.784814675086375)
--(axis cs:1.15031356405905,0.779553609695459)
--(axis cs:1.1356749225748,0.774292544304542)
--(axis cs:1.12197870203075,0.769031478913625)
--(axis cs:1.10929063587379,0.763770413522708)
--(axis cs:1.09758022691195,0.758509348131791)
--(axis cs:1.08679789326185,0.753248282740874)
--(axis cs:1.07691617214523,0.747987217349957)
--(axis cs:1.06792299334574,0.74272615195904)
--(axis cs:1.05978913497984,0.737465086568123)
--(axis cs:1.05244365001874,0.732204021177207)
--(axis cs:1.04577616522868,0.72694295578629)
--(axis cs:1.03966082952386,0.721681890395373)
--(axis cs:1.03398447052411,0.716420825004456)
--(axis cs:1.02866684415047,0.711159759613539)
--(axis cs:1.02367214415346,0.705898694222622)
--(axis cs:1.01901474791951,0.700637628831705)
--(axis cs:1.01475740077193,0.695376563440788)
--(axis cs:1.01099629811007,0.690115498049871)
--(axis cs:1.0078322120369,0.684854432658954)
--(axis cs:1.00533659505975,0.679593367268038)
--(axis cs:1.0035267769919,0.674332301877121)
--(axis cs:1.00235957666614,0.669071236486204)
--(axis cs:1.00174191915076,0.663810171095287)
--(axis cs:1.00154919566304,0.65854910570437)
--(axis cs:1.00164238130453,0.653288040313453)
--(axis cs:1.00188116946213,0.648026974922536)
--(axis cs:1.0021360419263,0.642765909531619)
--(axis cs:1.00230241764211,0.637504844140702)
--(axis cs:1.00231538452338,0.632243778749785)
--(axis cs:1.00215887895194,0.626982713358869)
--(axis cs:1.00186321307044,0.621721647967952)
--(axis cs:1.00149012193815,0.616460582577035)
--(axis cs:1.00111118261082,0.611199517186118)
--(axis cs:1.00078865104106,0.605938451795201)
--(axis cs:1.00056539430309,0.600677386404284)
--(axis cs:1.00046465413296,0.595416321013367)
--(axis cs:1.00049496922121,0.59015525562245)
--(axis cs:1.00065390218347,0.584894190231533)
--(axis cs:1.00092680774675,0.579633124840616)
--(axis cs:1.00128182368909,0.5743720594497)
--(axis cs:1.00166646174907,0.569110994058783)
--(axis cs:1.0020118684845,0.563849928667866)
--(axis cs:1.00224709495839,0.558588863276949)
--(axis cs:1.00231939467976,0.553327797886032)
--(axis cs:1.00221163864695,0.548066732495115)
--(axis cs:1.00194803489798,0.542805667104198)
--(axis cs:1.00158492775122,0.537544601713281)
--(axis cs:1.00119110088737,0.532283536322364)
--(axis cs:1.00082682352911,0.527022470931448)
--(axis cs:1.00053015390694,0.521761405540531)
--(axis cs:1.00031399003482,0.516500340149614)
--cycle;

\addplot [semithick, white!24.7058823529412!black, dash pattern=on 2.25pt off 2.25pt]
table {%
-0.298035941673338 0.86746988
0.298035941673338 0.86746988
};
\addplot [semithick, white!24.7058823529412!black, dash pattern=on 4.5pt off 4.5pt]
table {%
-0.389169707660473 0.913385827
0.389169707660473 0.913385827
};
\addplot [semithick, white!24.7058823529412!black, dash pattern=on 2.25pt off 2.25pt]
table {%
-0.340183670654431 0.956521739
0.340183670654431 0.956521739
};
\addplot [semithick, white!24.7058823529412!black, dash pattern=on 2.25pt off 2.25pt]
table {%
0.711100066374765 0.830769231
1.28889993362524 0.830769231
};
\addplot [semithick, white!24.7058823529412!black, dash pattern=on 4.5pt off 4.5pt]
table {%
0.604 0.876404494
1.396 0.876404494
};
\addplot [semithick, white!24.7058823529412!black, dash pattern=on 2.25pt off 2.25pt]
table {%
0.712250499513879 0.923076923
1.28774950048612 0.923076923
};
\end{axis}

\end{tikzpicture}

\end{minipage}
\begin{minipage}[t]{0.48\textwidth}
\centering
\begin{tikzpicture}[scale=0.75,transform shape]

\definecolor{color0}{rgb}{1,0.647058823529412,0}

\begin{axis}[
legend cell align={left},
legend style={fill opacity=0.8, draw opacity=1, text opacity=1, draw=white!80!black},
tick align=outside,
tick pos=left,
title={Histogram},
x grid style={white!69.0196078431373!black},
xmin=0.5315384617, xmax=1.0223076923,
xtick style={color=black},
y grid style={white!69.0196078431373!black},
ylabel={Frequency},
ymin=0, ymax=174.3,
ytick style={color=black}
]
\draw[draw=none,fill=blue,fill opacity=0.5] (axis cs:0.666666667,0) rectangle (axis cs:0.7000000003,2);
\addlegendimage{ybar,ybar legend,draw=none,fill=blue,fill opacity=0.5};
\addlegendentry{ESFL}

\draw[draw=none,fill=blue,fill opacity=0.5] (axis cs:0.7000000003,0) rectangle (axis cs:0.7333333336,5);
\draw[draw=none,fill=blue,fill opacity=0.5] (axis cs:0.7333333336,0) rectangle (axis cs:0.7666666669,12);
\draw[draw=none,fill=blue,fill opacity=0.5] (axis cs:0.7666666669,0) rectangle (axis cs:0.8000000002,22);
\draw[draw=none,fill=blue,fill opacity=0.5] (axis cs:0.8000000002,0) rectangle (axis cs:0.8333333335,46);
\draw[draw=none,fill=blue,fill opacity=0.5] (axis cs:0.8333333335,0) rectangle (axis cs:0.8666666668,63);
\draw[draw=none,fill=blue,fill opacity=0.5] (axis cs:0.8666666668,0) rectangle (axis cs:0.9000000001,117);
\draw[draw=none,fill=blue,fill opacity=0.5] (axis cs:0.9000000001,0) rectangle (axis cs:0.9333333334,127);
\draw[draw=none,fill=blue,fill opacity=0.5] (axis cs:0.9333333334,0) rectangle (axis cs:0.9666666667,110);
\draw[draw=none,fill=blue,fill opacity=0.5] (axis cs:0.9666666667,0) rectangle (axis cs:1,115);
\draw[draw=none,fill=color0,fill opacity=0.5] (axis cs:0.553846154,0) rectangle (axis cs:0.5984615386,1);
\addlegendimage{ybar,ybar legend,draw=none,fill=color0,fill opacity=0.5};
\addlegendentry{English}

\draw[draw=none,fill=color0,fill opacity=0.5] (axis cs:0.5984615386,0) rectangle (axis cs:0.6430769232,1);
\draw[draw=none,fill=color0,fill opacity=0.5] (axis cs:0.6430769232,0) rectangle (axis cs:0.6876923078,0);
\draw[draw=none,fill=color0,fill opacity=0.5] (axis cs:0.6876923078,0) rectangle (axis cs:0.7323076924,12);
\draw[draw=none,fill=color0,fill opacity=0.5] (axis cs:0.7323076924,0) rectangle (axis cs:0.776923077,34);
\draw[draw=none,fill=color0,fill opacity=0.5] (axis cs:0.776923077,0) rectangle (axis cs:0.8215384616,86);
\draw[draw=none,fill=color0,fill opacity=0.5] (axis cs:0.8215384616,0) rectangle (axis cs:0.8661538462,126);
\draw[draw=none,fill=color0,fill opacity=0.5] (axis cs:0.8661538462,0) rectangle (axis cs:0.9107692308,166);
\draw[draw=none,fill=color0,fill opacity=0.5] (axis cs:0.9107692308,0) rectangle (axis cs:0.9553846154,111);
\draw[draw=none,fill=color0,fill opacity=0.5] (axis cs:0.9553846154,0) rectangle (axis cs:1,82);
\end{axis}

\end{tikzpicture}

\end{minipage}
\caption{Experimental result on the semantic derivation transparency on ESFL and English data.}
\label{fig: visualization} 
\end{figure}
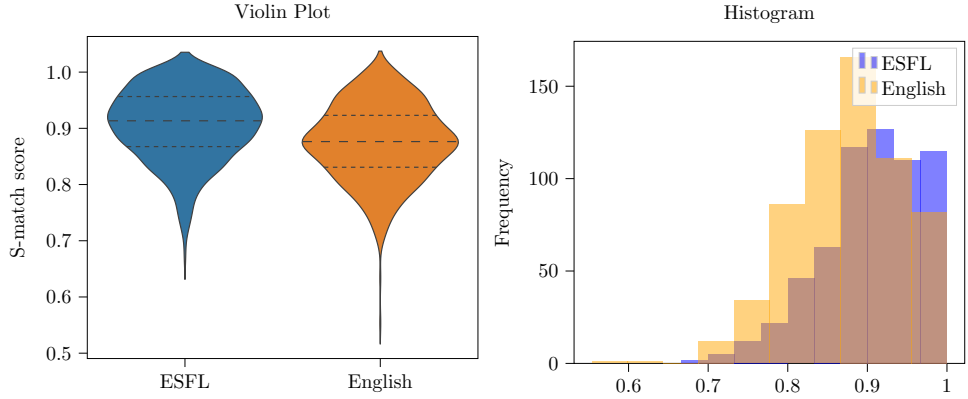

Figures \ref{fig: visualization} and Table \ref{tab:result} summarize the experimental results for both datasets. The ESFL dataset consistently achieves higher S-match scores than the English dataset, with a mean of 0.906 compared to 0.875, and exhibits lower variability, as indicated by a smaller standard deviation (0.064 vs. 0.068). The statistical analysis reported in Table \ref{tab:zttest} further corroborates these findings, demonstrating that ESFL not only attains higher accuracy but also shows greater stability in syntax–semantics mappings.

\begin{table}[htbp]
    \caption{Descriptive statistics of S-match scores for ESFL and English datasets.}
    \label{tab:result}
    \begin{tabular}{cccccc}\toprule
         & Mean & Median & SD & Max & Min\\\midrule
       ESFL  & 0.906 & 0.913 & 0.064 & 1& 0.668\\
       English &0.875 &0.876 & 0.068 & 1&0.554\\\bottomrule
    \end{tabular}
\end{table}

\begin{table}[htbp]
   \caption{Statistical tests comparing ESFL and English S-match scores.}
    \label{tab:zttest}
    \begin{tabular}{lll}\toprule
\multicolumn{1}{c}{\multirow{2}{*}{t-test}} & T-statistics & p-value   \\
\multicolumn{1}{c}{}                        & 14.343       & 8.772e-41 \\\midrule
\multirow{2}{*}{z-test}                     & z-score      & p-value   \\
                                            & 8.428      & 1.761e-17\\\bottomrule
\end{tabular}
\end{table}

From a linguistic perspective, this finding aligns with the hypothesis that ESFL, as a learner variety, tends toward more transparent form–meaning mappings, possibly as a cognitive adaptation that reduces processing complexity. Such transparency may facilitate both language comprehension and production in a second-language context, and it offers empirical support for theoretical accounts (e.g., the Linguistic Niche Hypothesis) that link communicative environments with structural regularity.

\section{Conclusion}
ESFL, used by multilingual speakers whose first language is not English, often departs from standard English through non-canonical lexical and grammatical patterns, creating challenges for systematic semantic derivation. At first glance, these deviations appear to threaten the principle of compositionality --- the idea that meaning arises from the meanings of parts and their syntactic combination \citep{partee1984}. Under this view, ESFL semantics might seem fundamentally unstable, given its variable lexical choices and non-canonical syntax. Yet multilingual speakers and listeners routinely comprehend ESFL with little difficulty, indicating that robust compositional mechanisms remain active beneath surface-level irregularities.

In this paper, we reconcile this apparent paradox by integrating constructivist theories with explicit modeling. Our results show that, despite syntactic variability, ESFL exhibits systematic and reliable mappings between form and meaning that support compositional interpretation. These findings suggest that ESFL speakers do not abandon grammatical organization when using a second or foreign language; rather, they adapt and reorganize linguistic resources to maintain communicative efficiency across typologically diverse systems. To support this claim empirically, we introduce ESFL SemBank, a gold-standard syntactico-semantic resource containing 1,643 manually curated ESFL sentences with explicitly aligned syntactic and semantic representations.

Given the global scale of multilingualism and the widespread use of ESFL, we argue that both our framework and ESFL SemBank have broad theoretical and practical significance. First, they provide a principled answer to a long-standing question in linguistics: whether ESFL constitutes a systematic linguistic system or merely a collection of deviations from native norms \citep{adjemian1976nature}. Our findings strongly support the former view, positioning ESFL as a structured outcome of multilingual competence rather than linguistic deficiency. Second, by offering a computationally grounded model of multilingual meaning construction, our work opens new avenues for research in applied linguistics, language acquisition, and NLP for learner language. As an initial demonstration, we evaluate the Linguistic Niche Hypothesis, illustrating its potential as a scalable testbed for future empirical investigations into multilingual language use and adaptation.

\backmatter

\bibliography{sn-bibliography}

\end{document}